\RequirePackage{fix-cm}
\documentclass[smallextended, referee]{svjour3}       
\smartqed  
\usepackage{amsfonts}
\usepackage[utf8]{inputenc}
\usepackage[english]{babel}
\usepackage{multirow}
\usepackage{multicol}
\usepackage{amssymb}
\usepackage{url} 
\usepackage{natbib}
\usepackage{algorithm}
\usepackage{algorithmicx}
\usepackage{dsfont}
\usepackage{algpseudocode}
\usepackage{epsfig}
\usepackage{epstopdf}
\usepackage{subfig}
\usepackage{booktabs} 
\usepackage{amsmath}
\usepackage{array}
\usepackage{tabularx}
\usepackage{ltablex}
\usepackage{enumitem}
\usepackage{varwidth}
\usepackage{longtable}
\usepackage[toc,page]{appendix}
\usepackage{multirow}
\newcolumntype{M}[1]{>{\centering\arraybackslash}m{#1}}
\usepackage[usenames, dvipsnames]{xcolor}
\definecolor{mylinkcolor}{HTML}{0000FA}
\usepackage{hyperref}
\hypersetup{
	colorlinks   = true, 
	urlcolor     = mylinkcolor, 
	linkcolor    = mylinkcolor, 
	citecolor   = mylinkcolor 
}
\newcolumntype{L}[1]{>{\raggedright\let\newline\\\arraybackslash\hspace{0pt}}m{#1}}
\newcolumntype{C}[1]{>{\centering\let\newline\\\arraybackslash\hspace{0pt}}m{#1}}

\usepackage{forloop}
\newcounter{ct}

\begin{document}

\title{Hyperbox based machine learning algorithms : A comprehensive survey 
}

\titlerunning{Hyperbox based machine learning algorithms : A comprehensive survey}        

\author{Thanh Tung Khuat \and Dymitr Ruta \and Bogdan Gabrys}


 \institute{Thanh Tung Khuat \at
 	            Advanced Analytics Institute, Faculty of Engineering and Information Technology, University of Technology Sydney \\
 		        \email{thanhtung.khuat@student.uts.edu.au}
 		   \and 
 		   Dymitr Ruta \at EBTIC, Khalifa University, UAE \\ \email{dymitr.ruta@kustar.ac.ae}
           \and
           Bogdan Gabrys \at
 	           Advanced Analytics Institute, Faculty of Engineering and Information Technology, University of Technology Sydney \\
 	            \email{bogdan.gabrys@uts.edu.au}
}

\date{Received: date / Accepted: date}

\maketitle

\begin{abstract}
With the rapid development of digital information, the data volume generated by humans and machines is growing exponentially. Along with this trend, machine learning algorithms have been formed and evolved continuously to discover new information and knowledge from different data sources. Learning algorithms using hyperboxes as fundamental representational and building blocks are a branch of machine learning methods. These algorithms have enormous potential for high scalability and online adaptation of predictors built using hyperbox data representations to the dynamically changing environments and streaming data. This paper aims to give a comprehensive survey of literature on hyperbox-based machine learning models. In general, according to the architecture and characteristic features of the resulting models, the existing hyperbox-based learning algorithms may be grouped into three major categories: fuzzy min-max neural networks, hyperbox-based hybrid models, and other algorithms based on hyperbox representations. Within each of these groups, this paper shows a brief description of the structure of models, associated learning algorithms, and an analysis of their advantages and drawbacks. Main applications of these hyperbox-based models to the real-world problems are also described in this paper. Finally, we discuss some open problems and identify potential future research directions in this field.

\end{abstract}

\section{Introduction}
According to \citet{Gross10}, learning is defined as a process of acquiring or changing existing knowledge, skills, behaviors, or preferences and synthesizing various kinds of information through experience, study, or being taught. Machine learning is a field of research concerned with the formulation and development of algorithms which provide the machine with the capability of learning and evolving their behaviors based on data coming from a variety of sources such as sensors or databases. \citet{Mitchell97} gave a formal definition of machine learning algorithms, which are software programs being able to do some tasks $ T $ by learning from experience $ E $ and their performance assessed by $ P $. Therefore, when designing a new machine learning algorithm, one needs to think of what data to collect $ (E) $, what decisions the algorithm must generate $ (T) $, and what metrics are used to assess its performance $ (P) $.

In the era of big data, a question of how we consume different data sources and transform them into valuable, actionable knowledge has become critically important. Over the last few decades, many data mining methods have been studied and expanded aiming to invent an effective way for discovering meaningful knowledge and information from raw data. These techniques have contributed to mining diverse patterns hidden in data repositories \citep{Zakaryazad16}. By comprehending the knowledge underlying data, one can make important decisions more accurately in numerous fields ranging from business, finance, medicine to manufacturing sectors. There have been a large number of studies conducted on the subjects of data mining, data analytics, as well as predictive modeling over the last 50 years with remarkable enhancements of the computing equipment and the algorithms \citep{Gabrys05}. Within all these studies, many machine learning algorithms have been developed with an emphasis on pattern clustering and classification.

Most prevalent classes of machine learning techniques for pattern classification are various types of artificial neural networks (ANNs) \citep{Mukhopadhyay07} because of their ability to learn from different types of input data, immunity to noise, generalization capability, and relatively high accuracy \citep{Kim11, Mohammed17b}. It has also been observed that the traditional machine learning models are trained and their parameters tuned on a given set of training data, and then the best model is deployed for dealing with specific problems without performing any updates afterward until the maintenance phase \citep{Fontama15}. With the exponential growth of digital content and information leading to the rise in data volume, such conventional batch learning or offline learning techniques face many limitations because they adapt poorly to the rapid changes of data and suffer from costly re-training when adaptation is needed. It is desirable to develop robust and scalable machine learning algorithms with the aim of robust adaptation to evolving data and changing environments. Learning algorithms need to provide the models with the ability to capture the new features of data in order to reduce the loss of predictive performance. One of the main issues with respect to the artificial neural networks and conventional classifiers using both incremental and batch learning is catastrophic forgetting also known as stability-plasticity dilemma \citep{Mccloskey89}, which relates to the inability of the classifier to retain previously learnt patterns when new patterns are absorbed by that classifier \citep{Polikar01}. Hence, classifiers frequently forget learned information while learning new information \citep{Grossberg13}. To tackle this issue, a classifier has to be able to remain plastic enough to absorb new information and simultaneously be stable enough to maintain previously acquired knowledge while learning new information \citep{Grossberg13}. Resolving the stability-plasticity dilemma problem is especially essential when using online-learning for classifiers \citep{Mccloskey89, Yang04}.

In addition to online learning, the effective machine learning model should possess several properties as follows \citep{Simpson92}:

\begin{itemize}
    \item Non-linear separability: This is ability to construct non-linear class boundaries \citep{Sonule17}.
    \item Overlapping region: The model is capable of formulating non-linear decision boundary to minimize the misclassification error for all overlapping classes.
    \item Soft and hard decisions: The algorithm should offer both soft and hard decisions. The hard decision allocates a sample to a single class, while the soft decision outputs the degree-of-fit of that pattern to a given class.
    \item Training time: The learning algorithm to train the model should be fast and have the ability to learn arbitrarily complex class decision boundaries.
    \item Verification and validation: This property imposes that each machine learning model should have the mechanisms to verify and validate its performance.
    \item Adjusting parameters: The model should have as few parameters that need to be adjusted during the training process as possible.
\end{itemize}

Motivated by all the above mentioned issues and desirable properties of the models, machine learning algorithms based on hyperboxes have been introduced to tackle supervised and unsupervised learning tasks. Simpson \citep{Simpson92} suggested deploying hyperbox fuzzy sets to generate and store information as hidden units in the form of neural network architecture. He introduced two kinds of hyperbox-based fuzzy min-max neural networks, one supervised learning technique for sample classification \citep{Simpson92} and one model for data clustering \citep{Simpson93}. The fuzzy min-max neural network (FMNN) possesses all useful characteristics mentioned above. Due to the benefits of FMNN, a great deal of its improved variants have been proposed such as general fuzzy min-max neural network (GFMN) \citep{Gabrys00}, weighted fuzzy min-max neural network \citep{Kim04}, an adaptive resolution fuzzy min-max neural network \citep{Rizzi02}, an inclusion/exclusion fuzzy hyperbox classification network \citep{Bargiela04}, a fuzzy min-max neural network classifier with compensatory neurons \citep{Nandedkar04} \citep{Nandedkar07b}, a data-core-based fuzzy min-max neural network \citep{Zhang11}, a multi-level fuzzy min-max neural network \citep{Davtalab14}. Each fuzzy min-max neural network includes many hyperboxes, each one covers an area determined by its minimum and maximum coordinates in the n-dimensional sample space. Each hyperbox is associated with a fuzzy membership function calculating the goodness-of-fit of an input sample to a certain class. From the original version proposed by \citet{Simpson92}, learning algorithms of the fuzzy min-max neural networks have been significantly enhanced with the emergence of algorithms combining supervised and unsupervised learning such as in the GFMN \citep{Gabrys00} and the general reflex fuzzy min-max neural network (GRFMN) \citep{Nandedkar09}; algorithms dealing with missing data and operating on observable subspaces without missing values imputation \citep{Gabrys02c, Castillo12}; combination of multiple hyperbox classifiers at a model level taking advantage of ensemble performance while reducing impact of user-defined parameters \citep{Gabrys02b}.

The traditional neural networks are considered as black boxes due to the fact that they are not able to explain their predicted results. When it comes to data analysis, one of the salient properties is the ability to extract explanatory rules for inference from data samples \citep{Cheng11}. Therefore, machine learning models should offer a useful explanatory mechanism of their outcomes to the user.  One of such models is the decision tree \citep{Seera15}. Fuzzy min-max neural networks can generate explanation based on the rules deduced from the hyperbox min–max values, but it cannot form a compact rule set interpretable for end-users because the number of hyperboxes can be large. Therefore,  instead of extracting rules directly from the individual hyperbox level, decision trees have been adopted to obtain rules at the global level. As a result, many researchers have introduced hybrid models in combination of hyperbox-based machine learning algorithms with decision trees or other rule extractors to increase the ability to explain the results for single models such as an enhanced FMNN with an ant colony optimization based rule extractor \citep{Sonule17}, a hybrid model of FMNN and the classification and regression tree \citep{Seera12} \citep{Seera14}, a fuzzy min-max based clustering tree \citep{Seera16}, a fuzzy min-max decision tree \citep{Mirzamomen16}, and combining multiple decision trees using the GFMM neural networks \citep{Eastwood11}. 

Apart from combination of the fuzzy min-max neural networks and other classification techniques, several researchers have introduced other methods to construct base hyperboxes and evolve them using optimization algorithms such as a differential evolution \citep{Reyes-Galaviz15} and an ant colony optimization \citep{Ramos09}.

This paper seeks to classify and clarify the properties of machine learning models based on the hyperbox representations, learning algorithms, as well as their enhancements. In other words, this work aims to provide a comprehensive survey of literature on hyperbox-based machine learning algorithms. The core ideas and key description of typical algorithms, their expansions, as well as their real-world applications are presented in detail. We expect to clarify issues as follows:

\begin{enumerate}
    \item The method of categorizing the machine learning models based on hyperbox representations. 
    \item What was the development trajectory of the original fuzzy min-max neural networks in the last two decades.
    \item What methods have been deployed to generate hybrid models between different types of FMNN and other classification or clustering techniques.
    \item Apart from network structures, what other methods are used to design and evolve hyperbox-based models.
    \item Identifying research gaps in the current hyperbox-based machine learning algorithms and propose new future research directions.
\end{enumerate}

To the best of our knowledge, this is the first comprehensive survey on hyperbox-based machine learning algorithms. Regarding the fuzzy min-max neural networks, which is part of our study, there have been several previously published surveys. \citet{Jambhulkar14} compared the multi-level fuzzy min-max neural networks with the original FMNN and its four different variants. However, that survey mentioned only six types of fuzzy min-max neural networks, and it did not yet analyze the limitations of existing types of networks. To overcome these drawbacks, \citet{Jain15} analyzed more details from some additional variants of the original FMNN, and they concluded that multi-level fuzzy min-max neural networks are the best one among the discussed methods. However, the survey was only limited to seven types of fuzzy min-max networks, and the authors used the classification accuracy of training samples as a comparison criterion for fuzzy min-max classifiers. The high accuracy on the training sample does not guarantee the good performance of the constructed classifiers because it may reflect overfitting and the loss of generality. In another study, \citet{Kulkarni16} reviewed different types of fuzzy neural networks for classification and clustering. They categorized the networks to three groups consisting the ones for classification, clustering, and hybrid models for both classification and clustering. However, they not only focused on fuzzy min-max networks but also other types of fuzzy neural networks such as hypersphere and hyperline ones. These three surveys have not yet clarified the still existing limitations and applications of the fuzzy min-max neural networks to real-world problems. Recently, there has been a survey on the fuzzy min-max neural networks for pattern classification until 2017 introduced by \citet{Sayaydeh18}. Authors classified the types of fuzzy min-max neural networks into two groups, i.e., ones with and without contraction process. They summarized the use of different types of fuzzy min-max neural networks in tackling real-world applications. Nonetheless, the paper did not present in depth the reasons for the proposals of variants of the original fuzzy min-max classifier and their improvements compared to previously proposed versions. The research direction part only mentioned a small aspect regarding the potential of the family of fuzzy min-max neural networks. Our study is not restricted to the fuzzy min-max neural networks but expands to general hyperbox-based machine learning algorithms, the combination of hyperbox fuzzy sets and tree-based algorithms, ensembles of multiple hyperbox-based models, the use of hyperbox fuzzy sets to deal with missing data, and learning algorithms based on hyperboxes without forming the neural network struture. 

The remainder of this paper is structured as follows. Section \ref{sec2} shows the summary of searching results, the taxonomy and description of main hyperbox-based machine learning algorithms and their applications in the real world. In section \ref{sec3}, we describe the background knowledge of machine learning models using the hyperbox representations. The overview of the architecture and main content of types of fuzzy min-max neural network and its improved versions are detailed in section \ref{sec4}. Section \ref{sec5} gives descriptions of hybrid models based on hyperboxes, while section \ref{sec6} provides other hyperbox-based machine learning techniques. The applications of hyperbox-based learning models for practical problems are shown in section \ref{sec7}. Section \ref{sec8} discusses main characteristics of hyperbox-based machine learning algorithms presented in this paper as well as potential research directions. Concluding remarks are described in section \ref{sec9}. 

\section{Summary and taxonomy of hyperbox-based machine learning algorithms} \label{sec2}
This survey targets to determine the studies related to machine learning algorithms using the hyperbox representations and their applications in the real world. We sought for research articles, conference papers or book chapters in five popular databases including \citet{ScienceDirect18}, \citet{IEEEXplore18}, \citet{SpringerLink18}, \citet{ACM18}, \citet{IOSPress18}. Readers interested in the methods of filtering and selecting the publications can refer to Appendices \ref{appendix1} and \ref{appendix2} for more details.

\subsection{Searching results}
\begin{figure}
    \centering
    \includegraphics[width=0.8\textwidth]{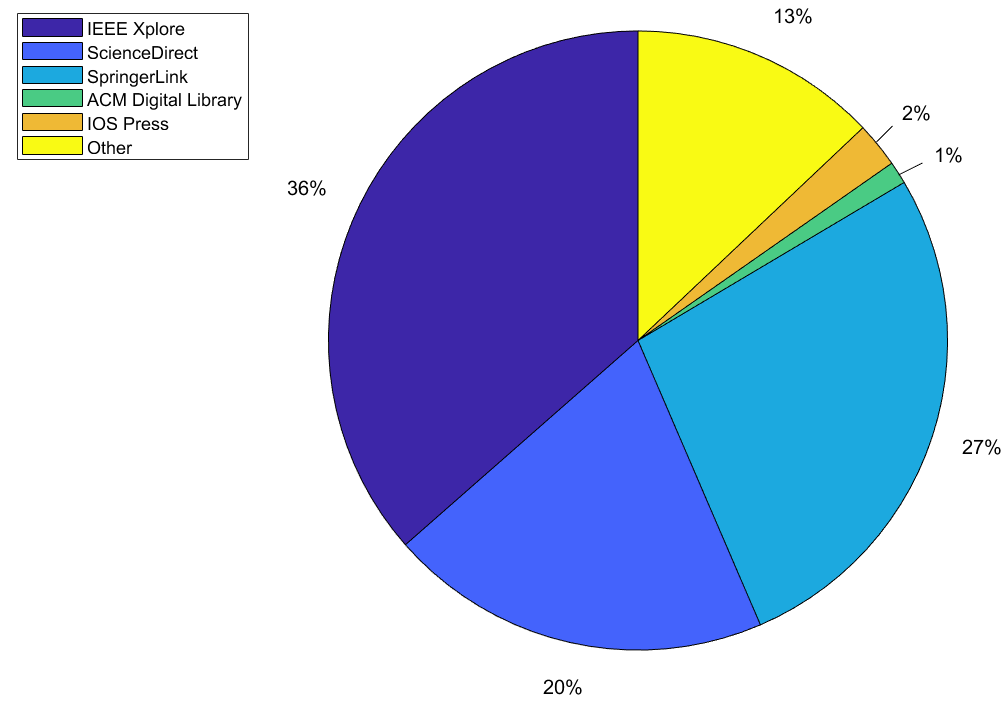}
    \caption{Proportion of publications from different data sources used in this paper}
    \label{fig1_n}
\end{figure}

The searching results over the available databases are shown in Table \ref{table1}. The ratio of publications from different research repositories covered in this work are illustrated in Fig. \ref{fig1_n}. For each search, we collected some essential information as follows: authors, year of publication, main purpose of forming the new proposal, brief description of the approach for problem solving, and potential applications of the proposed or existing methods.

\begin{table}[!ht]
    \caption{Literature search result from different data sources}
    \centering
    \begin{tabular}{l|c|c|c}
        \hline
        \textbf{Source} & \textbf{Total results} & \textbf{Screening} & \textbf{Final selection} \\
        \hline
        ScienceDirect & 524 & 45 & 17 \\
        IEEE Xplore	& 330 & 54 & 31 \\
        SpringerLink & 710 & 51 & 23 \\
        ACM Digital Library	& 6 & 2 & 1 \\
        IOS Press & 55 & 3 & 2 \\
        Other journals and conferences & - & - & 11 \\
        \hline
        \textbf{Total} & \textbf{1625} & \textbf{155} & \textbf{85} \\
        \hline
    \end{tabular}
    \label{table1}
\end{table}

\begin{figure}
    \centering
    \includegraphics[width=1\textwidth]{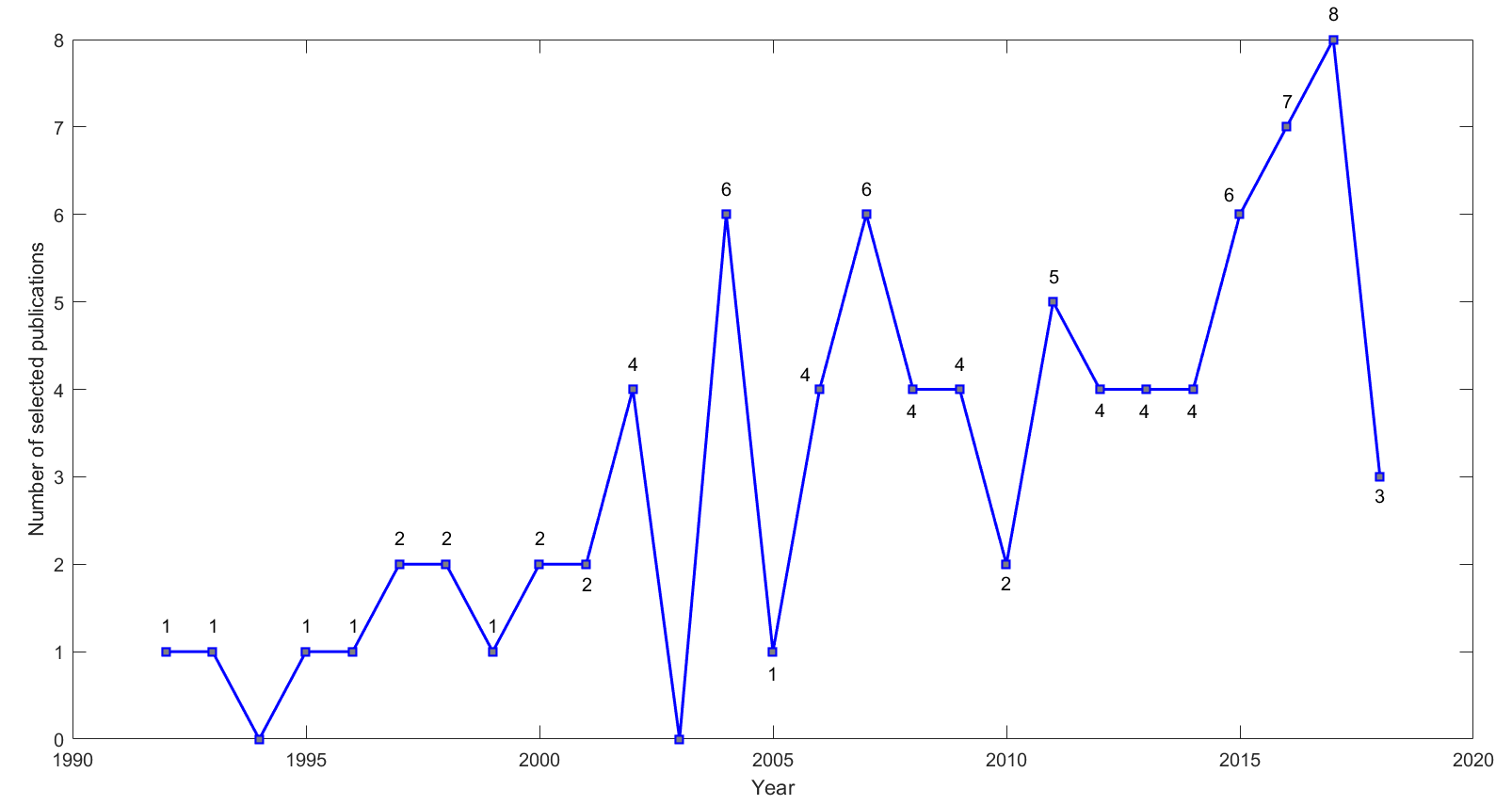}
    \caption{Selected publications per year}
    \label{fig2_n}
\end{figure}

In general, most relevant studies are stored in three key sources, i.e., IEEE Xplore, SpringerLink, and ScienceDirect. Fig. \ref{fig2_n} represents the number of selected papers for this review over years from 1992 to 2018. It can be clearly seen that after year 2012, the number of studies in this field has been increasing.

\subsection{Summary of main hyperbox-based machine learning algorithms}

\begin{figure}
    \centering
    \includegraphics[width=1\textwidth]{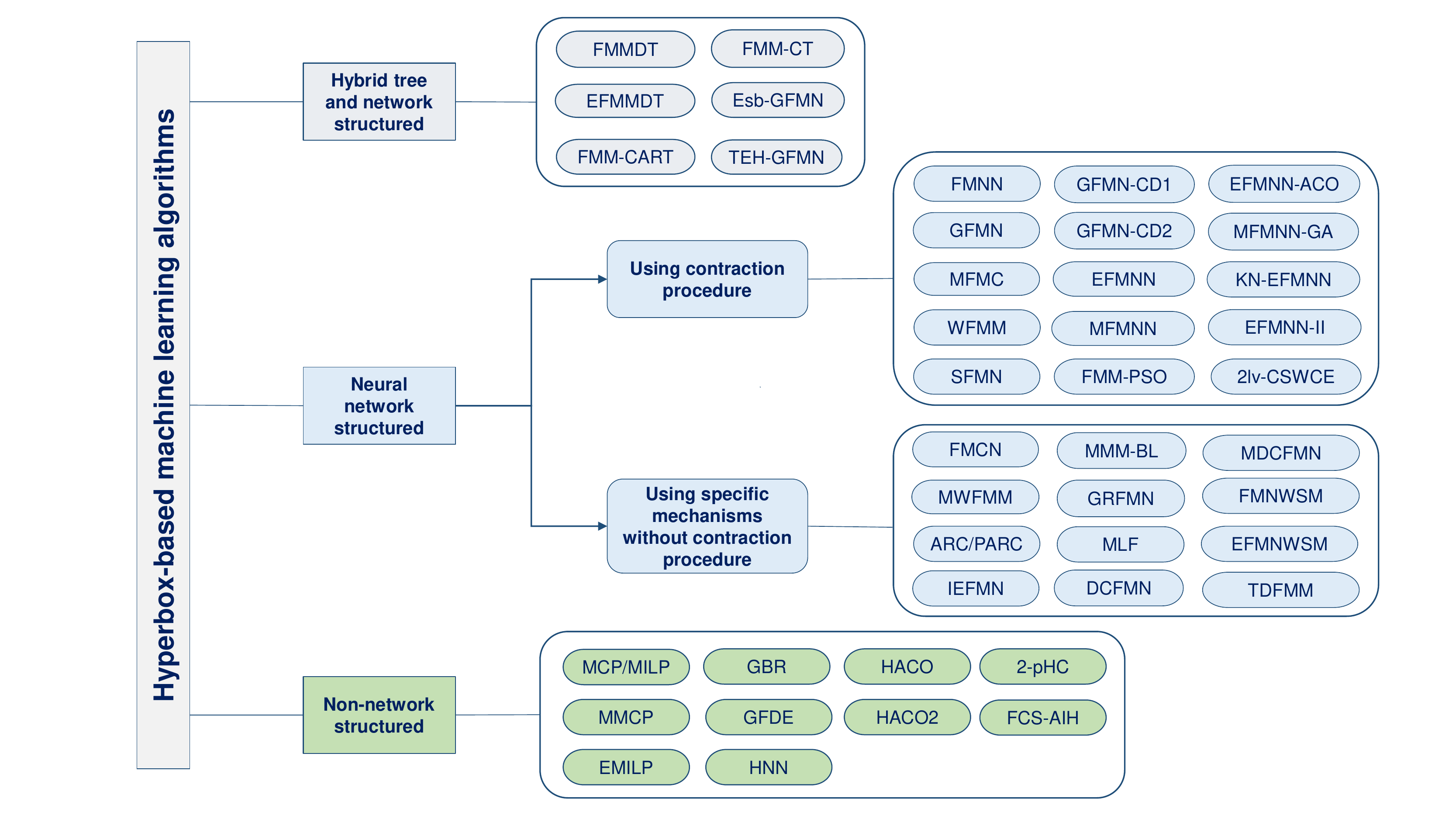}
    \caption{Taxonomy of hyperbox-based machine learning models}
    \label{fig4_n}
\end{figure}

Based on the contents of selected publications, we classify the hyperbox-based machine learning algorithms into three groups. The first group comprises studies that construct a neural network architecture from hyperbox fuzzy sets. Learning algorithms are designed to adjust the placement of hyperboxes to cover the training samples in the input space. In the process of expanding hyperboxes to include new training patterns, hyperboxes belonging to different classes are likely to overlap with each other. There are two methods to deal with this overlapping problems. While several studies use the contraction procedure to avoid overlapping regions, other researchers introduce specialized hyperboxes to handle overlapping areas. Therefore, fuzzy min-max neural networks can be divided into two sub-groups depending on the employed mechanism for handling overlapping hyperboxes. The second group of hyperbox-based machine learning models consists of ones that integrate the strong points of various fuzzy min-max neural networks and tree-based classification techniques or the combination of base models to build the ensembles. The last group includes the models without being connected by network structures. Hyperboxes in these machine learning models are constructed and evolved using different approaches such as mathematical formulas, or optimization algorithms. The illustration of our classification method for models based on the hyperbox representation is shown in Fig. \ref{fig4_n}. It is easily seen that most hyperbox-based machine learning algorithms are organized in the form of neural networks.

Several main types of machine learning algorithms based on hyperboxes have been summarized in Table \ref{table2}.

{
\scriptsize
\begin{longtable}[ht]{|p{0.05\textwidth}|p{0.20\textwidth}|p{0.1\textwidth}|p{0.18\textwidth}|p{0.4\textwidth}|}
\caption{Summary of hyperboxed-based machine learning algorithms} \label{table2} \\
\hline 
	\textbf{ID} & \textbf{Model} & \textbf{Year} & \textbf{Type} & \textbf{Characteristics}\\
	\endhead
        \hline
        T2.1 & Fuzzy min-max classification neural network \citep{Simpson92} & 1992 & classification & supervised learning; a hybrid neuro-fuzzy system built using hyperbox fuzzy sets for sample classification; three-step training process - hyperbox expansion, hyperbox overlap test, and hyperbox contraction; claimed ability to learn online and avoid re-training \\
        \hline
        T2.2 & Fuzzy min-max clustering neural network \citep{Simpson93} & 1993 & clustering & unsupervised learning; a hybrid neuro-fuzzy system built using hyperbox fuzzy sets for data clustering; four-step training process - initialization, hyperbox expansion, hyperbox overlap test, and hyperbox contraction; ability to learn online \\
        \hline
        T2.3 & Hyperbox fuzzy classifier (HFC) \citep{Abe95} & 1995 & classification & using activation and inhibition hyperboxes to extract fuzzy rules directly from numerical inputs; building a membership function between a sample and a rule \\
        \hline
        T2.4 & Stochastic fuzzy min-max neural network (SFMN) \citep{Likas96} \citep{Likas01}  & 1996; 2001 & classification; reinforcement & introducing a concept of random hyperboxes to deal with the discrete output space; a stochastic learning automaton is assigned to each hyperbox to manage the randomness degree in the process of action selection; reinforcement learning \\
        \hline
        T2.5 & Adaptive resolution min-max neural network classifier (ARC) \citep{Rizzi98} & 1998 & classification & consisting of the adaptive resolution classifier and pruning adaptive resolution classifier; definition of the hybrid and the pure hyperboxes; not depend on sample presentation order, position and hyperbox size as in FMNN \\
        \hline
        T2.6 & Modified fuzzy min-max neural network with a new batch learning algorithm (MMM-BL) \citep{Meneganti98} & 1998 & classification & using the same structure as FMNN; changing the learning algorithm by the hyperbox cutting methods; learning process includes five operations; introducing two kinds of hyperboxes: non-partitionable and partitionable; proposing a new membership function \\
        \hline
        T2.7 & Two-level classification system with testing in dynamically changing environment (2lv-CSWCE) \citep{Gabrys99} & 1999 & classification & the first level uses a GFMM classifier to select one of the $ n $ components in the second level; the second level comprises $ n $ neural networks, in which each network is trained on a part of the training set; the outcomes of second-level neural networks are classification results \\
        \hline
        T2.8 & General fuzzy min-max neural network (GFMN) \citep{Gabrys00} \citep{Gabrys02a} & 2000; 2002 & classification and clustering & dealing with both labeled, partly labeled, and unlabeled data; inputs are in form of hyperboxes; support both online and batch training; dealing with missing values \citep{Gabrys02c} \\
        \hline
        T2.9 & Top down fuzzy min-max (TDFMM) neural network \citep{Tagliaferri01} & 2001 & classification & simplifying steps in the learning algorithm of the MMM-BL; reducing the complexity; defining two types of partition for the hyperbox cutting process \\
        \hline
        T2.10 & Top down fuzzy min-max regressor (TDFMMR) \citep{Tagliaferri01} & 2001 & regression & building a regressor based on FDFMM; only one node in the output layer of the network; using clustering to find class labels of samples; apply to back-propagation algorithm to adjusting parameters of the membership function in the hidden neurons\\
        \hline
        T2.11 & Ensemble of neuro-fuzzy classifiers (Esb-GFMN) \citep{Gabrys02b} & 2002 & classification; clustering & combine hyperbox fuzzy sets of base classifiers; repeat 2-fold splitting of the training data; hyperbox fuzzy sets from different component classifiers are combined as inputs to the agglomerative training algorithm \\
        \hline
        T2.12 & Weighted fuzzy min-max (WFMM) neural network \citep{Kim04} \citep{Kim05} \citep{Kim06} & 2004-2006 & classification; feature extraction & a new membership function; assign a weight factor to each dimension within a hyperbox; new mechanisms for expansion, contraction, and weight updating procedures \\
        \hline
        T2.13 & Inclusion/Exclusion fuzzy min-max neural network (IEFMN) \citep{Bargiela04} & 2004 & classification & employing the inclusion hyperboxes to include the input samples of the same class, utilize exclusion hyperboxes to cover other overlapped input samples; two-step training process - hyperbox expansion and overlap test; inputs are in form of hyperboxes \\
        \hline
        T2.14 & Fuzzy min-max neural network with compensatory neuron (FMCN) \citep{Nandedkar04} \citep{Nandedkar07b} & 2004; 2007 & classification & solving hyperbox overlaps and containment problems by compensatory neurons; define new membership function and network structure for three types of neurons: classifying, overlap compensation, containment compensation neurons \\
        \hline
        T2.15 & Two-phase hyperbox classifier (2-pHC) \citep{Bortolan07} & 2007 & classification & in the first phase, ``seed hyperboxes" are constructed by means of the fuzzy C‐means algorithm; in the second phase, these hyperboxes are expanded by genetic algorithms \\
        \hline
        T2.16 & Modified fuzzy min-max neural network for two-stage pattern classification (MFMNN) \citep{Quteishat08b} & 2008 & classification & the first stage performs the training and hyperbox pruning process, while the second stage deploys the rule extraction; use both membership function and Euclidean distance in the prediction phase \\
        \hline
        T2.17 & Hyperbox classifier with ant colony optimization - type 2 (HACO2) \citep{Ramos08} & 2008 & classification & the ant colony algorithm is used to evolve the geometry of hyperboxes after hyperboxes are grouped into clusters\\
        \hline
        T2.18 & Hyperbox based clustering with ant colony optimization (HACO) \citep{Ramos09} & 2009 & clustering & using ant colony optimization to scatter the hyperboxes on the feature space; group generated hyperboxes into corresponding clusters\\
        \hline
        T2.19 & General reflex fuzzy min-max neural network (GRFMN) \citep{Nandedkar09} & 2009 & classification; clustering & conditions of the input data are the same as the GFMM networks; a reflex mechanism is deployed to resolve the hyperbox overlap and containment issues; define new activation functions of classifying, overlap compensation, and containment compensation neurons \\
        \hline
        T2.20 & Mathematical programming-based hyperbox classifier (MCP/MILP) \citep{Xu09} & 2009 & classification & construct for each class a number of hyperboxes; initially, a box is constructed for each class, and then an repetitive algorithm is deployed to form multiple boxes for each class; special constraints are employed to prevent overlapping of boxes representing different classes; the optimal position and dimension of each hyperbox is identified using a mixed integer linear programming method \\
        \hline
         T2.21 & Modified fuzzy min-max neural network with genetic algorithms (MFMNN-GA) \citep{Quteishat10} & 2010 & classification & performing training and hyperbox pruning in the first stage and rule extraction using genetic algorithms in the second stage; employ both membership function and Euclidean distance in the prediction phase  \\
        \hline
         T2.22 & Data-core-based fuzzy min-max neural network (DCFMN) \citep{Zhang11} & 2011 & classification & unlike FMCN, only overlapping neurons are used to solve the overlap and containment issues; allow expanded hyperboxes to overlap repeatedly with the previous hyperboxes; define new membership function for classifying neurons based on the characteristics of data and the impact of noise \\
        \hline
        T2.23 & Tree ensemble hyperboxes via general fuzzy min-max neural network (TEH-GFMN) \citep{Eastwood11} & 2011 & classification & using an ensemble of decision trees to robustly label a set of hyperboxes; each hyperbox corresponds to an overlap of leaf nodes; the classifier is built on labelled hyperbox patterns \\
        \hline
        T2.24 & Hyperbox neural network algorithm (HNN) \citep{Palmer-Brown11} & 2011 & classification & each class is associated with only one hyperbox and a single neuron; using hyperboxes to classify easy samples; neurons are trained to classify samples within the overlapping regions among hyperboxes\\
        \hline
        T2.25 & Modified data-core-based fuzzy min–max neural network with new learning mechanism (MDCFMN) \citep{Ma12} & 2012 & classification & proposing a new training algorithm, remove the contraction process without adding any neurons; using center of data in each hyperbox to find the suitable hyperbox in case of many hyperboxes with the same membership value \\
        \hline
        T2.26 & General fuzzy min–max neural networks for categorical data (GFMN-CD1) \citep{Castillo12} & 2012 & classification & each input neuron representing categorical data is linked to each hyperbox in the second layer by two connections; define a new membership function for both numerical and categorical data; modify expansion condition for categorical data; dealing with missing data \\
        \hline
        T2.27 & Offline and online fuzzy min–max neural network and classification and regression trees (FMM-CART) \citep{Seera12} \citep{Seera14} & 2012; 2014 & classification; regression & employing the FMNN for pattern classification and CART for extracting rules; providing online and offline learning \\
        \hline
        T2.28 & Multi-level fuzzy min-max neural network (MLF) \citep{Davtalab14} & 2014 & classification & implementing a multi-level tree structure to form a homogeneous cascading classifier; resolving the overlapped area issue by a multi-level network structure; the recognition rate for training pattern is 100\% potential causing over-fitting \\
        \hline
        T2.29 & Fuzzy min-max neural network with symmetric margin (FMNWSM) \citep{Forghani15} & 2015 & classification & only hyperbox creation and expansion procedures are performed in the training phase; being simpler and faster than other types of FMNN; propose new membership function \\
        \hline
        T2.30 & Enhanced fuzzy min-max neural network (EFMNN) \citep{Mohammed15} & 2015 & classification & keeping the structure of the original FMNN; changing the learning algorithm by adding more cases for the hyperbox overlap test and contraction process \\
        \hline
        T2.31 & Modified fuzzy min-max neural network for clustering (MFMMC) \citep{Seera15} & 2015 & clustering & associated each hyperbox fuzzy set with a centroid construction procedure; validating the hyperbox updating process by ensuring that the data centroid is still located within that hyperbox; the cophenetic correlation coefficient and centroid are employed to analyze the cluster validity \\
        \hline
        T2.32 & Enhanced general fuzzy min–max neural networks for categorical data (GFMN-CD2) \citep{Shinde16} & 2016 & classification & adding a set of binary strings representing discrete attributes to each hyperbox; each neuron for categorical input is connected to each hyperbox in the second layer by only one connection; define new membership function for both categorical and numerical data; define new expansion condition for categorical attributes; modify overlap test and contraction procedures \\
        \hline
        T2.33 & Fuzzy min-max clustering neural network with the clustering tree (FMM-CT) \citep{Seera16} & 2016 & clustering & combining online learning and rule extraction; integrating a new data centroid into each hyperbox; hyperbox centroids along with their confidence factors are used as input data samples to build the clustering tree \\
        \hline
        T2.34 & Fuzzy min-max neural network with genetic algorithms (FMM-GA) \citep{Azad16} & 2016 & classification & using the original version of the FMNN; genetic algorithms are used to optimize the size of hypeboxes generated by the FMNN \\
        \hline
        T2.35 & Fuzzy min–max neural network based decision tree (FMMDT) \citep{Mirzamomen16} \citep{Mirzamomen17} & 2016; 2017 & classification & each decision node of the tree holds a contraction-less fuzzy min-max neural network for batch learning or a concept adapting contraction less fuzzy min–max neural network \citep{Mirzamomen17} for online learning; not using the contraction process; using neural network to form the splitting condition for each decision node \\
        \hline
        T2.36 & Enhanced fuzzy min-max neural network with K-nearest hyperbox expansion rule (KN-EFMNN) \citep{Mohammed17a} & 2017 & classification & maintaining the structure of the original FMNN; adding more cases for the hyperbox overlap test and contraction operation; adopting K-nearest neighbour principle to select the winning hyperbox for the expansion process \\
        \hline
        T2.37 & Enhanced fuzzy min-max neural network with K-nearest hyperbox expansion rule and pruning (EFMNN-II) \citep{Mohammed17b} & 2017 & classification & having the same features as the KN-EFMNN; deploying a pruning process to reduce the network complexity \\
        \hline
        T2.38 & Enhanced fuzzy min–max neural network with ant colony optimization (EFMNN-ACO) \citep{Sonule17} & 2017 & classification & adding more cases for the hypberbox overlap test and contraction; using outputs of the network to construct a graph of AntMinerPlus algorithm for rule extractor; extracting the rule-list and prune rules by paths on the graph \\
        \hline
        T2.39 & Modified fuzzy min–max neural network for data clustering (MFMC) \citep{Liu17} & 2017 & clustering & retaining the architecture of the original fuzzy min-max clustering neural network; introducing a new hyperbox selection rule, a reservation rule, a hyperbox entropy measure for contraction process; using a parameter to avoid boundary overlapping \\
        \hline
        T2.40 & Fuzzy min-max neural network with the particle swarm optimization (FMM-PSO) \citep{Azad17} & 2017 & classification & using the original version of the FMNN; the particle swarm optimization algorithm is deployed to optimize the size of hypeboxes produced by the FMNN\\
        \hline
        T2.41 & Fuzzy Min–Max neural network with an ensemble of clustering trees (FMM-ECT) \citep{Seera18} & 2018 & clustering & proposing new rules to check the hyperbox centroid; using a new measure for the impurity of node; using an ensemble model of clustering trees; supporting online learning and rule extraction \\
        \hline
\end{longtable}
}

\subsection{Summary of applications}

\begin{figure}[!ht]
    \centering
    \includegraphics[width=0.7\textwidth, height=0.8\textheight]{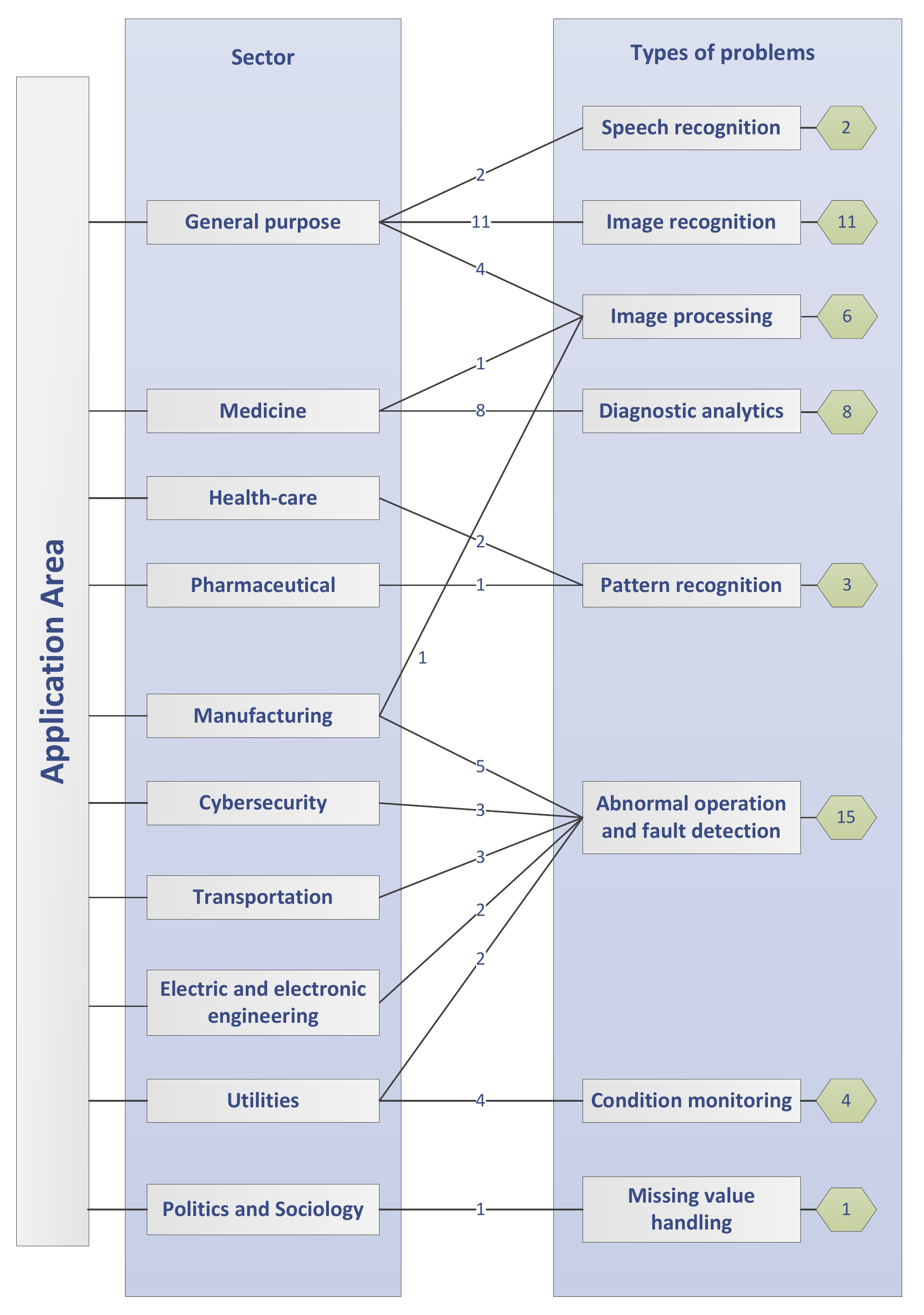}
    \caption{Taxonomy of applications and the number of selected publications}
    \label{fig_app}
\end{figure}

Hyperbox-based learning algorithms have been primarily applied to pattern classification problems \citep{Abe95}, \citep{Gabrys99}, \citep{Ma12}, \citep{Mohammed15}, \citep{Nandedkar06a}, \citep{Nandedkar07a}, \citep{Quteishat08b}, \citep{Quteishat10}, \citep{Rizzi02}, \citep{Simpson92}, \citep{Xu09}, \citep{Yang15}, and data clustering \citep{Simpson93}, \citep{Reyes-Galaviz15}, \citep{Liu17}, \citep{Seera16}. In addition to these two types of typical problems, hyperbox-based machine learning techniques have been used to deal with a great deal of applications in the real world. These applications can be categorized into two main groups. The first one includes general purpose applications, which are common application problems applicable to many specific case studies. For instance, image recognition, which is a general problem, contains different sub-problems such as face detection, character recognition, or signature recognition. Face detection is likely to be used to construct the access control systems in buildings or medication adherence monitoring systems in healthcare. The second group comprises the domain-driven applications such as transportation, cybersecurity, and medicine. Fig. \ref{fig_app} shows the taxonomy of real-world applications of hyperbox-based machine learning models and the number of selected publications of each class of applications.

Table \ref{table3} summarizes the main studies where the hyperbox-based machine learning models have been applied to solve the general purpose and domain-driven applications problems in practice.

{
	\scriptsize
	\begin{longtable}{|p{0.13\textwidth}|p{0.13\textwidth}|p{0.13\textwidth}|p{0.13\textwidth}|p{0.05\textwidth}|p{0.07\textwidth}|p{0.23\textwidth}|}
		\caption{Summary of main application areas} \label{table3} \\
		\hline 
		\textbf{Sector} & \textbf{Type of problems} & \textbf{Research subject} & \textbf{Author} & \textbf{Year} & \textbf{Method} & \textbf{Characteristics}\\
		\hline
		\endhead
		\multirow{2}{*}{\parbox{1\linewidth}{\vspace{0.2cm} General purpose (subsection \ref{generalpurpose})}} & speech recognition (subsection \ref{speechrecog}) & spoken Marathi India digits & \citet{Doye02} & 2002 & T2.8 & modifying the GFMM with a new transfer function of output layer; reaching 90.2\% average recognition accuracy in the speaker dependent mode \\
		\cline{3-7}
		\multirow{4}{*}{\parbox{1\linewidth}{\vspace{7cm} General purpose (subsection \ref{generalpurpose})}} & speech recognition (subsection \ref{speechrecog}) & text independent speaker & \citet{Jawarkar11} & 2011 & T2.1 & dataset comprises speech utterances with digits, words, and sentences in the Marathi language of fifty speakers; one minute duration speech utterance is deployed to train for each speaker; the maximum accuracy is higher than 96\% \\
		\cline{2-7}
		& \multirow{3}{*}{\parbox{1\linewidth}{\vspace{5cm} image processing (subsection \ref{imageprocessing})}} & color image segmentation & \citet{Deshmukh06} & 2006 & T2.2 & performing adaptive multilevel color image segmentation in HSV color space; seeking for clusters of pixels and their labels; can be used for object extraction from noisy environments\\
		\cline{3-7}
		& & color image or video frames segmentation & \citet{Nandedkar09} & 2009 & T2.19 & handling a group of pixels instead of individual color pixels, so reduce computational expense; different parts in an image are used to train the model; online training \\
		\cline{3-7}
		&  & shadow removal in color images & \citet{Nandedkar13} & 2013 & T2.19 & detecting and eliminating shadows from color images interactively; deploying GRFMN as a shadow classifier; the hyperboxes represent min, max, and mean values of the pixels from grids of shadow and non-shadow regions \\
		\cline{2-7}
		\multirow{4}{*}{\parbox{1\linewidth}{\vspace{5cm} General purpose (subsection \ref{generalpurpose})}} & \multirow{10}{*}{\parbox{1\linewidth}{\vspace{5cm} image recognition (subsection \ref{imagerecog})}} & handwritten Chinese character & \citet{Chiu97} & 1997 & T2.1 & hyperboxes were used because of the high shape variation of handwritten Chinese characters, fuzzy borders of the character classes, and the difficulty in representative training pattern in advance; ring-data features were extracted from characters and used to train the FMNN; the learning speed is rapid \\
		\cline{3-7}
		& & optical character & \citet{Nandedkar04b} & 2004 & T2.1 & system is invariable to translation, rotation, and scale; recognizing 26 uppercase letters of the alphabet from rotation, translation, and scale invariant features \\
		\cline{3-7}
		& & Bengali and Marathi digits & \citet{Nandedkar06b} & 2006 & T2.14 & invariance with rotation, translation, and scale invariant (RTSI) of digits; 300 samples with 10 classes were used for testing the model  \\
		\cline{3-7}
		\multirow{3}{*}{\parbox{1\linewidth}{\vspace{6cm} General purpose (subsection \ref{generalpurpose})}} & \multirow{4}{*}{\parbox{1\linewidth}{\vspace{6cm} image recognition (subsection \ref{imagerecog})}} & face detection & \citet{Kim06} & 2006 & T2.12 & constructing a real-time face detection system; using weight factor to consider the relevance of each feature for computing the hyperbox membership values; using weighted FMM for the skin-color filter; skin-color and other features of the face are deployed to train the classifier \\
		\cline{3-7}
		& & \multirow{2}{*}{\parbox{1\linewidth}{\vspace{2.5cm} shape}} & \citet{Nandedkar06b} & 2006 & T2.14 & building a rotation, translation, and scale invariant (RTSI) object recognition system; RTSI features of images are used to train the model; capable of learning from both labeled and unlabeled data; the unlabeled hyperboxes are considered as floating neurons and are removed from the process of classification \\
		\cline{4-7}
		& & & \citet{Nandedkar09} & 2009 & T2.19 & using data granules as inputs for object recognition; applying the GRFMN with granules constructed from minimum and maximum values of RTSI features in images \\
		\cline{3-7}
		\multirow{3}{*}{\parbox{1\linewidth}{\vspace{5cm} General purpose (subsection \ref{generalpurpose})}} & \multirow{3}{*}{\parbox{1\linewidth}{\vspace{5cm} image recognition (subsection \ref{imagerecog})}} & image retrieval & \citet{Kshirsagar16} & 2016 & T2.19 & color and texture features are extracted from the images to train the GRFMN in unsupervised mode aiming to create different clusters of images in the form of hyperboxes; clusters are labeled and saved in database; the model is used to assess the similarity of each image in the database to a query image \\
		\cline{3-7}
		& & offline single signature & \citet{Chaudhari09} & 2009 & T2.1 & the inputs are digital image from optical scanner; offline operation; decision boundaries are fuzzy; incorporate new and enhance existing classes without retraining; accuracy is approximately 53\% for single signature samples per class \\
		\cline{3-7}
		& & online signature & \citet{Chaudhari10} & 2010 & T2.14 & real time classification; employing invariant Krawtchouk moment method for preprocessing \\
		\cline{3-7}
		\multirow{2}{*}{\parbox{1\linewidth}{\vspace{5cm} General purpose (subsection \ref{generalpurpose})}} & \multirow{2}{*}{\parbox{1\linewidth}{\vspace{5cm} image recognition (subsection \ref{imagerecog})}} & hand gesture & \citet{Kim13b} & 2013 & T2.12 & the system can handle a three-dimensional image; convolutional neural network is utilized to extract features from the motion history data of hand gestures; using the WFMM to find the relevance factor between sample classes and features; the weight factor of the WFMM can represent the relationship of feature range and its distribution; IF-THEN rules are extracted from the classifier  \\
		\cline{3-7}
		& & sign language & \citet{Kim13} & 2013 & T2.12 & three kinds of features from sign language instances are extracted; defining an enhanced membership function of hyperboxes with the frequency factor of the features; extracting rules based on the relevance factors between feature values and sample classes\\
		\hline
		\multirow{3}{*}{\parbox{1\linewidth}{\vspace{6cm} Medicine (subsection \ref{medicine})}} & \multirow{3}{*}{\parbox{1\linewidth}{\vspace{6cm} diagnostic analytics}} & angiographic disease status & \citet{Kim04} & 2004 & T2.12 & selecting features to support for medical diagnosis data; each feature is associated with a weight factor; a relevance factor for each feature to a class from the trained hyperbox network is computed to select relevant features; doing experiments on a five-class dataset with 297 patterns \\
		\cline{3-7}
		& & arrhythmia & \citet{Bortolan07} & 2007 & T2.15 & analyzing and classifying electrocardiography (ECG) data; 26 features from ECG and premature ventricular contraction signals are used \\
		\cline{3-7}
		& & tumors & \citet{Juan07} & 2007 & T2.1 & clustering gene expression data into the groups by KNN; choosing the top-ranked genes from each cluster using an enhanced fuzzy min-max neural network; performing experiments on the dataset of the small, round blue-cell tumors with 6567 genes of 88 samples \\
		\cline{3-7}
		\multirow{4}{*}{\parbox{1\linewidth}{\vspace{6cm} Medicine (subsection \ref{medicine})}} & \multirow{4}{*}{\parbox{1\linewidth}{\vspace{6cm} diagnostic analytics}} & acute coronary syndrome & \citet{Quteishat08a} & 2008 & T2.1 & using different variants of the FMNN; testing the model on a set of 118 practical medical records in Penang Hospital, Malaysia; 16 features related to the acute coronary syndrome are extracted \\
		\cline{3-7}
		& & stroke & \citet{Quteishat10} & 2010 & T2.21 & predicting the Rankine scale related to the stroke in the medical records of patients; datasets contain 661 patient records; 18 typical features are extracted \\
		\cline{3-7}
		& & lung cancer & \citet{Zhai14} & 2014 & T2.13 & detecting lung nodules in X-ray images by the FMCN and K-means clustering algorithm; the lung volume segmentation from X-ray image for detecting the nodule Candidates; 11 features, including intensity and geometry features, of these candidates are extracted and used to train the FMCN \\
		& & lung cancer & \citet{Deshmukh16} & 2016 & T2.1 & classifying pathological lung cancers;  using Lung cancer data from UCI machine learning repository for experiment \\
		\cline{3-7}
		\cline{3-7}
		Medicine (subsection \ref{medicine}) & diagnostic analytics & liver disease & \citet{Tran18} & 2018 & T2.2 & semi-supervised clustering; training data include 4156 samples of patients from Gang Thep Hospital and Thai Nguyen National Hospital; all input samples are unlabeled \\
		\hline
		\multirow{2}{*}{\parbox{1\linewidth}{\vspace{3cm} Health-care (subsection \ref{healthcare})}} & \multirow{2}{*}{\parbox{1\linewidth}{\vspace{3cm} pattern recognition}} & medical risk profiles recognition & \citet{Ramos09} & 2009 & T2.18 & conducting experiments on a data set including 185 patterns with 21 features of oral mucosa and a questionnaire for 42 physical characteristics and habits among the local residents\\
		\cline{3-7}
		& & fall behavior recognition of patients & \citet{Jahanjoo17} & 2017 & T2.28 & detecting fall from wearable acceleration sensor data; dataset contains two motion categories, i.e., fall and daily living activities of six males and five females with various heights weights, and ages; 43 features are extracted \\
		\hline
		Pharmaceutical (subsection \ref{drug}) & pattern recognition & drug discovery & \citet{Tardu16} & 2016 & T2.20 & filtering the compounds in the initial libraries unsuitable for drug candidates \\
		\hline
		Manufacturing (subsection \ref{manufacturing}) & abnormal operation and fault detection & cooling system & \citet{Meneganti98} & 1998 & T2.6 & finding anomalies in the cooling system of a blast furnace; input data were measured by sensors in the automation system including three failure classes and 21 features;  \\
		\cline{3-7}
		\multirow{3}{*}{\parbox{1\linewidth}{\vspace{5cm} Manufacturing (subsection \ref{manufacturing})}} & \multirow{3}{*}{\parbox{1\linewidth}{\vspace{5cm} abnormal operation and fault detection}} & production line & \citet{Meneganti98} & 1998 & T2.6 & defect diagnosis for the production line of a hardware plant; product images are processed to extract 27 features; classifying defects to one of seven classes  \\
		\cline{3-7}
		& & heat transfer in the condensers of circulating water systems & \citet{Chen04} & 2004 & T2.1 & integrating the FMNN with symbolic rule extraction based on real sensor data; performing experiments on a set of 2439 real sensor data samples with 12 features collected from the circulating water system of a power generation plant in Penang, Malaysia \\
		\cline{3-7}
		& & heat transfer and blockage conditions in the condenser tubes of circulating water systems & \citet{Quteishat07, Quteishat08b} & 2007, 2008 & T2.16 & modifying the fuzzy min–max neural network; using rule extraction for binary classification of heat transfer conditions; doing experiments on a data set of 2439 samples with 12 features from sensor data of the condenser tubes in a circulating water system of a power generation plant \\
		\hline
		Cybersecurity (subsection \ref{cybersecurity}) & abnormal operation and fault detection & intrusion detection & \citet{Azad16} & 2016 & T2.34 & integrating the genetic algorithm to the FMNN to construct a intrusion detection system; optimizing hyperboxes by the genetic algorithm \\
		\cline{4-7}
		\multirow{2}{*}{\parbox{1\linewidth}{\vspace{2cm} Cybersecurity (subsection \ref{cybersecurity})}} & \multirow{2}{*}{\parbox{1\linewidth}{\vspace{2cm} abnormal operation and fault detection}} & intrusion detection & \citet{Azad17} & 2017 & T2.40 & combining the particle swarm optimization (PSO) with the FMNN to build the intrusion system; hyperboxes are optimized by the PSO \\
		\cline{3-7}
		& & attack intention recognition & \citet{Ahmed18} & 2018 & T2.1 & real time operation system; computing the similarity degrees of evidence in every observed attack using the FMNN; hyperboxes represent the possible attack evidence \\
		\hline
		\multirow{2}{*}{\parbox{1\linewidth}{\vspace{2cm} Transportation (subsection \ref{transportation})}} & \multirow{2}{*}{\parbox{1\linewidth}{\vspace{2cm} abnormal operation and fault detection}} & autonomous vehicle navigation & \citet{Likas96} & 1996 & T2.4 & using the FMNN as an action selection network with five suitable driving commands of the autonomous vehicle; input data is the current status of the vehicle determined by eight sensors \\
		\cline{3-7}
		& & the training system of robot navigation & \citet{Duan07} & 2007 & T2.1 & integrating the FMNN and reinforcement learning; hyperboxes represent the segmentation regions within the state space of reinforcement learning; FMNN is used to refine the ability of detecting faults in the training process of the obstacle avoidance behavior for the robot\\
		\cline{3-7}
		Transportation (subsection \ref{transportation}) & abnormal operation and fault detection & fault identification in rail vehicle systems & \citet{Lv15} & 2015 & T2.1 & fault diagnosis in rail vehicle suspension components; building the fault simulation platform of suspension system; using 63 useful feature values for each component fault; classification for four types of faults \\
		\hline
		\multirow{2}{*}{\parbox{1\linewidth}{\vspace{0.2cm} Electric and electronic engineering (subsection \ref{engineering})}} & \multirow{2}{*}{\parbox{1\linewidth}{\vspace{0.2cm} abnormal operation and fault detection}} & \multirow{2}{*}{\parbox{1\linewidth}{\vspace{0.2cm} induction motors}} & \citet{Seera12} & 2012 & T2.27 & features of fault conditions in induction motors are extracted; classification of fault conditions; offline operation \\
		\cline{4-7}
		& & & \citet{Seera14}  & 2014 & T2.27 & online motor fault detection; extracting fault conditions from vibration signals\\
		\hline
		Utilities (subsection \ref{powersystem}) & abnormal operation and fault detection  & water distribution systems & \citet{Gabrys99, Gabrys00} & 1999; 2000 & T2.8 & the input samples can be either fuzzy or crisp, unlabeled or labeled; require the ability of constant learning without retraining because of unpredicted size, states, and anomalies of the system; training data of 24-h period of operation were generated by computer simulations including normal operating states and 10 levels of leakages \\
		\cline{2-7}
		\multirow{4}{*}{\parbox{1\linewidth}{\vspace{5cm} Utilities (subsection \ref{powersystem})}} & \multirow{4}{*}{\parbox{1\linewidth}{\vspace{5cm} condition monitoring}} & the running status of oil pipeline system & \citet{Zhang11} & 2011 & T2.22 & classification of running statuses including four types; performing 200 simulations with water rather than oil in a real pipeline system \\
		\cline{3-7}
		&  & \multirow{2}{*}{\parbox{1\linewidth}{\vspace{3cm} power quality}}  & \citet{Seera15} & 2015 & T2.31 & modifying the FMM clustering neural network to find the root reason of power supply disruptions and to predict the power quality for the operation of medical equipment; performing the experiments on real data sets of 1601 samples with 12 features from a hospital in the state of Pahang, Malaysia \\
		\cline{4-7}
		& & & \citet{Seera16} & 2016 & T2.33 & the system is able to explain when encountering anomalies; identifying anomalies in power quality data \\
		\cline{3-7}
		&  & oil pipeline internal status & \citet{Liu17} & 2017 & T2.39 & analyzing inspection data of the oil pipeline; analyzing the leakage data to find abnormal phenomena in the pipeline \\
		\hline
		Politics and sociology (subsection \ref{politics}) & missing value handling & imputation of missing values in surveys of voting intention polls & \citet{Castillo12} & 2012 & T2.26 & imputing the missing voting intention values from the values of the category with the highest membership degree; considering eleven categories for the voting intention variables; input features are 16 numerical and ordered and non-ordered categorical variables;  \\
		\hline
	\end{longtable}
}

\section{Overview of fuzzy min-max machine learning models} \label{sec3}
\begin{figure}[!ht]
	\centering
	\includegraphics[width=0.6\textwidth]{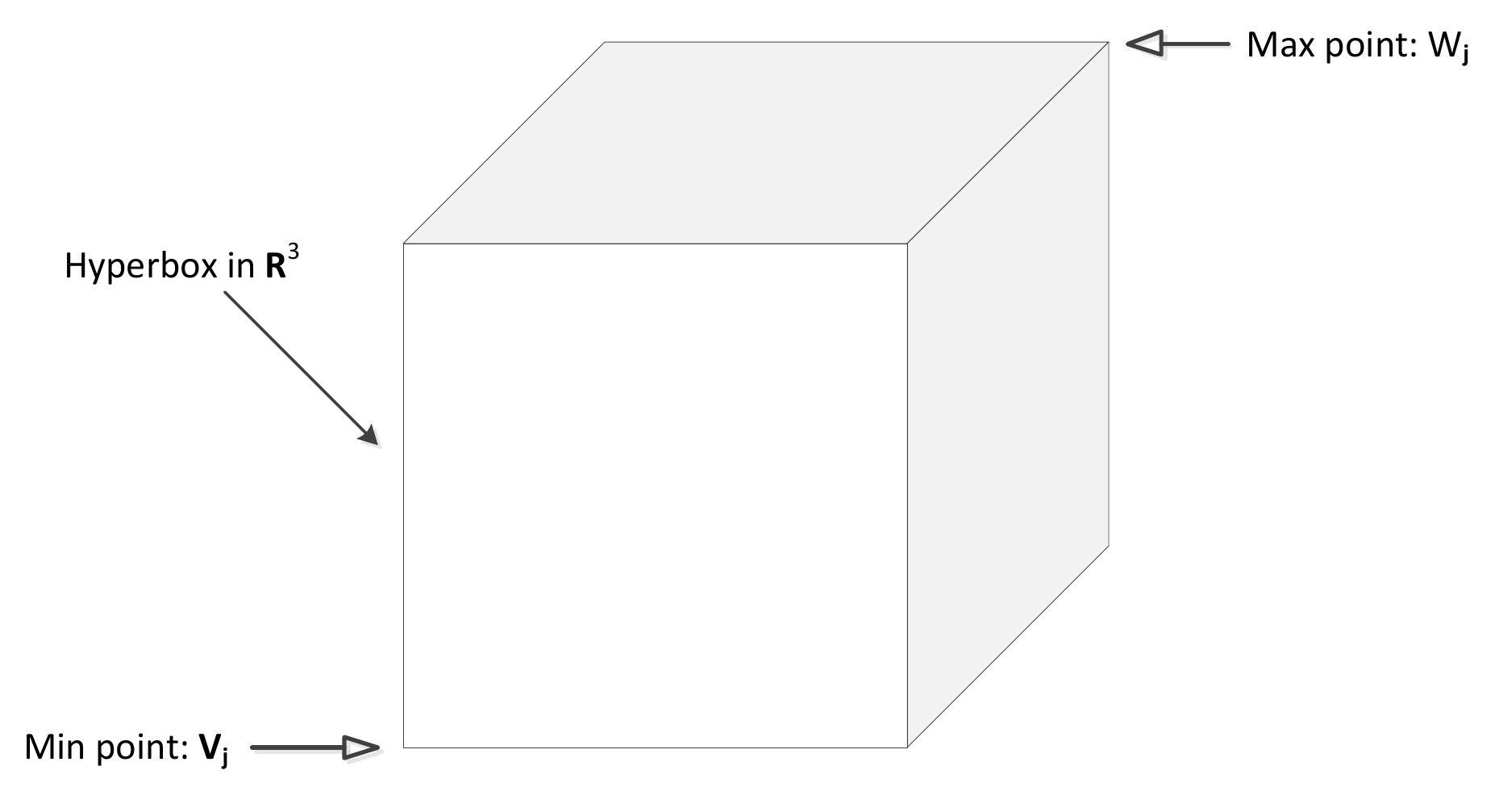}
	\caption{3-D Hyperbox}
	\label{fig1}
\end{figure}

Fuzzy min-max (FMM) machine learning models comprise hyperboxes and corresponding membership functions which are utilized to generate fuzzy subsets of the n-dimensional sample space \citep{Simpson92}. Each hyperbox occupies a region in the feature space and is defined by pairs of minimum and maximum points. Fig. \ref{fig1} represents an example of a three-dimensional hyperbox along with its min-point ($ V_j $) and max-point ($ W_j $). Based on new incoming data samples, FMM model generates a number of hyperboxes incrementally to establish new classes/clusters or tune the existing hyperboxes to cover new samples. It is possible to produce hyperboxes covering an arbitrary value range in each dimension, but the range from 0 to 1 is widely used for each dimension to make the computations simpler. Therefore, each hyperbox is usually determined by a set of minimum and maximum vertices in the n-dimensional unit cube ($ I^n $). 

Formally, each hyperbox fuzzy set is given by an ordered set

\begin{equation}
\label{eq1}
B_j = \{X_h, V_j, W_j, b_j(X_h, V_j, W_j)\}
\end{equation}
for all $ h = 1, 2,..., N $ where $ N $ is the quantity of input data samples, $ B_j $ is the $ j^{th} $ hyperbox fuzzy set, $ X_h = (x_{h1}, x_{h2},\ldots, x_{hn}) \in I^n $ is the $ h^{th} $ input data instance, $ V_j = (v_{j1}, v_{j2},..., v_{jn}) $ and $ W_j = (w_{j1}, w_{j2},..., w_{jn}) $ are minimum and maximum vertices of hyperbox $ B_j $ respectively, and the membership function of $ B_j $ is represented by $ 0 \leq b_j(X_h, V_j, W_j) \leq 1 $.

The membership function is a crucial component in the fuzzy min-max classification and clustering techniques. It is utilized to measure the degree to which the $ h^{th} $ input pattern, $ X_h $, belongs to the $ j^{th} $ hyperbox defined by the minimum point $ V_j $ and the maximum point $ W_j $.  When the sample is completely contained within the hyperbox, the degree of membership of $ X_h $ is one, and it decreases when $ X_h $ moves away from the hyperbox $ B_j $.

By the use of fuzzy min-max model, the first step to classify or cluster an input sample is to compute its membership function for each class/cluster $ c_k $ by accumulating the membership functions of all the hyperboxes representing this class/cluster as follows:

\begin{equation}
\label{eq2}
c_k = \bigcup \limits_{j \in K}{b_j}
\end{equation}
where $ K $ is the set of hyperboxes associated with class/cluster $ c_k $. It is noted that fuzzy union operator in this equation is the maximum of its membership functions as shown in Eq. \ref{eq4}. The next step is to classify the pattern to a corresponding class/cluster with the highest degree of membership \citep{Castillo12}.

\section{Fuzzy min-max neural network architectures}\label{sec4}
Fuzzy min-max neural network (FMNN) is a special type of a hybrid neuro-fuzzy system built using hyperbox fuzzy sets, and it is very competitive with other machine learning methods in terms of accuracy of the classification or clustering results and online adaptation ability \citep{Joshi97}. Learning process in the fuzzy min-max neural network is realized by properly constructing and tuning hyperboxes in the sample space. From the FMNN proposed by \citet{Simpson92}, a great deal of studies have introduced different methods to improve the performance of this type of neural network. As mentioned above, the enhancements have focused on two main directions. The first direction aims to improve the training algorithms by modifying the expansion and contraction processes to reach better performance of classification. The second approach does not require the contraction step, but it targets constructing novel architectures of neural networks with special neurons responsible for overlapping regions among hyperboxes covering different class labels. In the following section, the original fuzzy min-max neural network will be first presented, and then its improved versions are briefly described.

\subsection{Original fuzzy min-max neural networks} \label{sec4.1}
\subsubsection{The original FMM classification network} \label{fmnn}

To better address the stability-plasticity dilemma, the family of adaptive resonance theory (ART) neural networks with the incremental learning ability was proposed by \citet{Carpenter91, Carpenter92}. This type of neural network adapts the learned prototype only if the input pattern is similar enough to the prototype. An input sample which deviates too much in comparison with all existing prototypes is considered a new one, and the ART network will generate a new category with the input sample being the prototype. Inspired by the learning paradigm of the ART networks and to overcome their observed limitations, \citet{Simpson92} proposed the FMM classification neural network, which is a classification technique capable of generating nonlinear boundaries for splitting the input variables space into classes with any size and shapes \citep{Gabrys02a}. The training phase can be carried out with only one pass over the training samples, and it might be employed for pattern classification tasks \citep{Simpson92, Davtalab14}. 

\begin{figure}[!ht]
	\centering
	\includegraphics[width=0.7\textwidth]{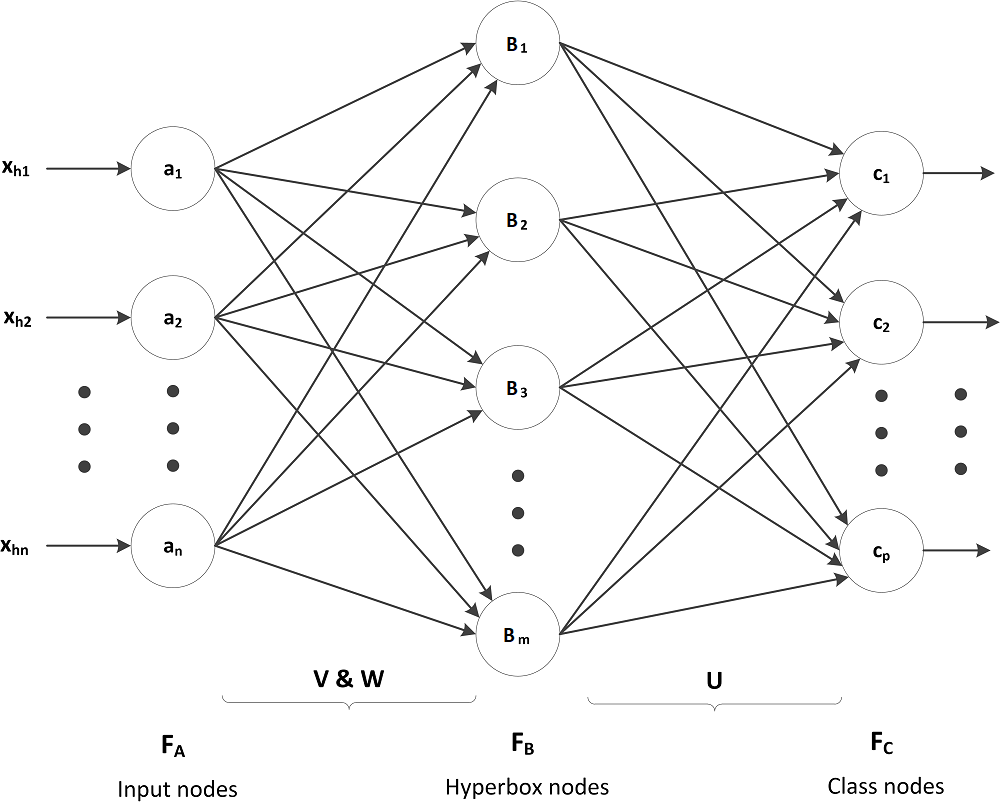}
	\caption{Three-layer FMM network}
	\label{fig2}
\end{figure}

Fig. \ref{fig2} describes a three-layer feed-forward neural network implementing a fuzzy min–max neural classifier. The FMM structure encompasses three layers in which the first layer ($ F_A $) is the input layer, the intermediate one ($ F_B $) shows hyperbox nodes using their membership function as transition function, and each node in the last layer ($ F_C $) represents a class. The number of $ F_C $ nodes is equal to the number of output classes. The output of each node in the layer $ F_C $ renders the degree to which the input value $ X_h $ fits within an output class. Each node $ B_j $ in the layer $ F_B $ shows a hyperbox fuzzy set and is connected with each node of the layer $ F_A $ via two weights ($ v_{ij} $ and $ w_{ij} $) that are the minimum and maximum points of the $ B_j $ hyperbox and $ i $ is the index of nodes in the layer $ F_A $. The weight connections between nodes in the second-layer and third-layer nodes are binary values and stored in a matrix $ U $, as shown in Eq. \ref{eq3}.

\begin{equation}
\label{eq3}
u_{jk} = \begin{cases}
1, & \mbox{if } b_j \in c_k \\
0, & \mbox{if } b_j \notin c_k
\end{cases}
\end{equation} 
where $ b_j $ is the $ j^{th} $ intermediate layer node and $ c_k $ is the $ k^{th} $ output layer node. The transfer function for each node in the output layer is the fuzzy union of the suitable hyperbox fuzzy set values as defined in Eq. \ref{eq4} for each node in the layer $ F_C $.

\begin{equation}
\label{eq4}
c_k = \max \limits_{j = 1}^{m}{b_j \cdot u_{jk}}
\end{equation}

There are two ways to generate the classifier output for each input sample $ X_h $. The outputs of the class nodes $ F_C $ can be utilized directly if a soft decision is required. In contrast, for a hard decision, a winner-takes-all principle \citep{Kohonen89} is used to select the node in the layer $ F_C $ with the greatest $ c_k $ value as the predicted class for the pattern $ X_h $.

All hyperboxes are produced and tuned in a learning process. When a new training sample $ X_h $ is provided, FMM employs the membership function to seek the closest hyperbox matching this sample. The membership function is shown in Eq. \ref{eq5} \citep{Simpson92}:

\begin{equation}
\label{eq5}
b_j(X_h) = \cfrac{1}{2n} \cdot \sum \limits_{i = 1}^{n}{[\max(0, 1 - \max(0, \gamma \cdot \min(1, x_{hi} - w_{ji}))) + \max(0, 1 - \max(0, \gamma \cdot \min(1, v_{ji} - x_{hi})))]}
\end{equation}
where $ b_j $ is the membership function for the $ j^{th} $ hyperbox, $ X_h = (x_{h1}, x_{h2},..., x_{hn}) \in I^n $ is the $ h^{th} $ input sample, $ V_j = (v_{j1}, v_{j2},..., v_{jn}) $ and $ W_j = (w_{j1}, w_{j2},..., w_{jn}) $ are min and max points for $ B_j $ respectively, and $ \gamma $ is a sensitivity parameter regulating how fast the membership function values decrease as the distance between $ X_h $ and $ B_j $ increases.

If there is a hyperbox $ B_j $ with the membership function being equal to 1, meaning that the sample falls within it, then no further step is required and training phase goes on with the next sample. Otherwise, the three-step process, i.e., hyperbox expansion, hyperbox overlap test, and hyperbox contraction, is performed as follows:

\textit{a. Hyperbox expansion}

In this step, we have to identify a hyperbox with the highest degree of fit using Eq. \ref{eq5} in order to perform an expansion process. For an ordered pair $ \{X_h, c_k\} $ in the training set, where $ X_h = (x_{h1}, x_{h2},..., x_{hn}) \in I^n $ is the input sample and $ c_k \in \{1,..., K\} $ is the index of one of $ K $ classes, it is expected to find a hyperbox $ B_j $ representing the same class as $ c_k $ with the highest membership value and after potential expansion, the size of the hyperbox not exceeding the user-defined parameter $ \theta $ ($ 0 \leq \theta \leq 1 $) so that it can include a new input sample $ X_h $. The constraint as shown in Eq. \ref{eq6} must be satisfied for $ B_j $:

\begin{equation}
\label{eq6}
n \cdot \theta \geq \sum \limits_{i = 1}^{n}{\max(w_{ji}, x_{hi}) - \min(v_{ji}, x_{hi})}
\end{equation}

If the extension criterion is met for hyperbox $ B_j $, the minimum point of $ B_j $ is tuned using Eq. \ref{eq7} and its maximum point is adjusted using Eq. \ref{eq8}.

\begin{equation}
\label{eq7}
v_{ji}^{new} = \min(v_{ji}^{old}, x_{hi}), \forall i = 1,...,n
\end{equation}

\begin{equation}
\label{eq8}
w_{ji}^{new} = \max(w_{ji}^{old}, x_{hi}), \forall i = 1,...,n
\end{equation}

If the constraint in Eq. \ref{eq6} is not satisfied for all existing hyperboxes, a new hyperbox is generated with min and max points equal to the corresponding input sample so that input pattern is encoded into the network. This growth method enables new input samples with new classes to be supplemented without retraining \citep{Gabrys99}. 

\textit{b. Hyperbox overlap test}

This step is to determine whether the expansion generated any overlapping areas between the expanded hyperbox and the existing ones of other classes. Assume that the hyperbox $ B_j $ was expanded in the previous step and the hyperbox $ B_k $ describing another class is being checked for possible overlapping. The overlap test is based on four cases for each dimension $ i $ of two hyperboxes as follows ($ \delta = 1 $ initially):

\textit{Case 1:} $ v_{ji} < v_{ki} < w_{ji} < w_{ki}, \qquad \delta^{new} = \min(w_{ji} - v_{ki}, \delta^{old}) $

\textit{Case 2:} $ v_{ki} < v_{ji} < w_{ki} < w_{ji}, \qquad \delta^{new} = \min(w_{ki} - v_{ji}, \delta^{old}) $

\textit{Case 3:} $ v_{ji} < v_{ki} < w_{ki} < w_{ji}, \qquad \delta^{new} = \min(\min(w_{ki} - v_{ji}, w_{ji} - v_{ki}), \delta^{old}) $

\textit{Case 4:} $ v_{ki} < v_{ji} < w_{ji} < w_{ki}, \qquad \delta^{new} = \min(\min(w_{ji} - v_{ki}, w_{ki} - v_{ji}), \delta^{old}) $

When $ \delta^{old} - \delta^{new} > 0 $, then $ \Delta = i $ and $ \delta^{old} = \delta^{new} $, it means that there is an overlap for the $ \Delta^{th} $ dimension and the overlap checking proceeds with the next dimension. If not, the testing process stops, and the next contraction step is flagged as `not essential' by setting $ \Delta = -1 $.

\textit{c. Hyperbox contraction}
  
If overlap between hyperboxes of different classes does exist, a hyperbox contraction step occurs to eliminate the overlapping areas by minimally contracting each of overlapped hyperboxes \citep{Simpson92}. If $ \Delta > 0 $, the $ \Delta^{th} $ dimensions of the two hyperboxes are adjusted. During the contraction stage, the hyperbox size is kept as large as possible by contracting only one of the $ n $ dimensions in each overlapped hyperbox aiming to provide more robust pattern classification. The same four cases are examined to determine the adjustment to be performed:

\textit{Case 1:} $ v_{j\Delta} < v_{k\Delta} < w_{j\Delta} < w_{k\Delta}, \qquad w_{j\Delta}^{new} = v_{k\Delta}^{new} = \cfrac{w_{j\Delta}^{old} + v_{k\Delta}^{old}}{2} $

\textit{Case 2:} $ v_{k\Delta} < v_{j\Delta} < w_{k\Delta} < w_{j\Delta}, \qquad w_{k\Delta}^{new} = v_{j\Delta}^{new} = \cfrac{w_{k\Delta}^{old} + v_{j\Delta}^{old}}{2} $

\textit{Case 3a:} $ v_{j\Delta} < v_{k\Delta} < w_{k\Delta} < w_{j\Delta} $ and $ (w_{k\Delta} - v_{j\Delta}) < (w_{j\Delta} - v_{k\Delta}), \qquad v_{j\Delta}^{new} = w_{k\Delta}^{old} $

\textit{Case 3b:} $ v_{j\Delta} < v_{k\Delta} < w_{k\Delta} < w_{j\Delta} $ and $ (w_{k\Delta} - v_{j\Delta}) > (w_{j\Delta} - v_{k\Delta}), \qquad w_{j\Delta}^{new} = v_{k\Delta}^{old} $

\textit{Case 4a:} $ v_{k\Delta} < v_{j\Delta} < w_{j\Delta} < w_{k\Delta} $ and $ (w_{k\Delta} - v_{j\Delta}) < (w_{j\Delta} - v_{k\Delta}), \qquad w_{k\Delta}^{new} = v_{j\Delta}^{old} $

\textit{Case 4b:} $ v_{k\Delta} < v_{j\Delta} < w_{j\Delta} < w_{k\Delta} $ and $ (w_{k\Delta} - v_{j\Delta}) > (w_{j\Delta} - v_{k\Delta}), \qquad v_{k\Delta}^{new} = w_{j\Delta}^{old} $

\subsubsection{The original FMM clustering network}
The original FMM clustering neural network, which is a method of partitional clustering, was first introduced by \citet{Simpson93}. Similarly to the FMM classification network, learning in the fuzzy min-max clustering neural network includes generating and adjusting hyperboxes in the sample space as they come. Once the fuzzy min-max clustering neural network is trained, it is used to classify a sample presented to the network by calculating the membership degree of that sample in each of the current hyperbox fuzzy sets. Given an input sample $ X_h $, min point $ V_j $ and max point $ W_j $ for the $ j^{th} $ hyperbox, its membership function is defined in Eq. \ref{eq9}.

\begin{equation}
\label{eq9}
b_j(X_h, V_j, W_j) = \cfrac{1}{n} \cdot \sum \limits_{i = 1}^{n}{[1 - f(x_{hi} - w_{ji}, \gamma) - f(v_{ji} - x_{hi}, \gamma)]}
\end{equation}
where $ f(\xi, \gamma) $ is a two-parameter ramp threshold function:

\begin{equation}
\label{eq10}
f(\xi, \gamma) = \begin{cases}
1, & \mbox{if } \xi \cdot \gamma > 1 \\
\xi \cdot \gamma, & \mbox{if } 0 \leq \xi \cdot \gamma \leq 1 \\
0, & \mbox{if } \xi \cdot \gamma < 0
\end{cases}
\end{equation}
where $ \gamma $ is the sensitivity parameter controlling the speed with which the membership value of an input sample decreases with its distance from the hyperbox core. When the value of $ \gamma $ is large, the fuzzy set becomes more crisp, and when it is small, the fuzzy set becomes less crisp.

Given a training set $ \mathbb{T} = \{X_h | h = 1, 2,..., m \} $, where $ X_h = (x_{h1}, x_{h2},..., x_{hn}) \in I^n $ is the input sample, the learning process starts by choosing a sample from $ \mathbb{T} $ and seeking the closest hyperbox to that sample being able to expand for containing the pattern. If there is no hyperbox satisfying the expansion criterion, a new hyperbox is created and added to the network. This growth method enables existing clusters to be refined incrementally, and it also allows new clusters to be formed without retraining. Generally, the learning process of the FMM clustering network comprises four steps, i.e., hyperbox initialization, expansion, overlap test, and contraction.

\textit{a. Hyperbox initialization }

As for the training algorithm of the fuzzy min-max clustering neural network, there are two cluster sets utilized, i.e., the committed set $ \mathbb{C} $ and the uncommitted set $ \mathbb{U} $. The committed set of clusters are those with their min-max coordinates tuned, while the uncommitted set of clusters are those waiting to be committed. Initially, the network is set up with an empty committed clusters set and an arbitrary quantity of clusters in the uncommitted set \citep{Simpson93}. The hyperboxes $ B_j $ that lie in the uncommitted set $ \mathbb{U} $ have min points $ V_j $ initialized to $ \overrightarrow{1} $  and max points $ W_j $ initialized to $ \overrightarrow{0} $, where $ \overrightarrow{1} $ is the n-dimensional vector of all ones and $ \overrightarrow{0} $ denotes the n-dimensional vector of all zeros. 
The goal of this initialization process is to make sure that the first hyperboxes taken out of the uncommitted set $ \mathbb{U} $ will perform the expansion to result in a single point identical to the input pattern: $ V_j = W_j = X_h $.

\textit{b. Hyperbox expansion}

Given a sample $ X_h \in \mathbb{T} $, search for a hyperbox $ B_j \in \mathbb{C} $ with the highest degree-of-fit and satisfying the expansion criteria. The degree of membership $ b_j(X_h, V_j, W_j) $ is computed using Eq. \ref{eq9}. The expansion constraint is determined in the same way as the FMM classification network using Eq. \ref{eq6}. If this criterion is satisfied, the min and max points are tuned using Eqs. \ref{eq7} and \ref{eq8} respectively.  

If none of the hyperboxes $ B_j \in \mathbb{C} $ satisfy the expansion criterion then a hyperbox in $ \mathbb{U} $ is selected and Eqs. \ref{eq7} and \ref{eq8} applied. 

\textit{c. Hyperbox overlap test}

After an expansion operation, an overlap is likely to exist between the expanded hyperbox and the other existing hyperboxes. Assuming that the hyperbox $ B_j \in \mathbb{C} $ is extended in the previous step, the hyperbox overlap test is executed between the hyperbox $ B_j $ and all remaining ones $ B_k \in \mathbb{C} $. There are four cases used to test whether the hyperbox $ B_j $ and the hyperbox $ B_k $ may form an overlapping region for each of the $ n $ dimensions:

\textit{Case 1:} $ v_{ji} < v_{ki} < w_{ji} < w_{ki} $

\textit{Case 2:} $ v_{ki} < v_{ji} < w_{ki} < w_{ji} $

\textit{Case 3:} $ v_{ji} < v_{ki} \leq w_{ki} < w_{ji} $

\textit{Case 4:} $ v_{ki} < v_{ji} \leq w_{ji} < w_{ki} $

The hyperbox contraction is used to eliminate the overlap regions among hyperboxes if one of the four above cases is satisfied.
 
\textit{d. Hyperbox contraction} 

If there is no overlap, this step is not necessary, else the contraction process is executed for each pair of overlapping hyperboxes $ B_j $ and $ B_k $. Based on the four cases previously presented, the contraction rules are shown as follows:

\textit{Case 1:} If $ v_{ji} < v_{ki} < w_{ji} < w_{ki}, \qquad v_{ki}^{new} = w_{ji}^{new} = \cfrac{v_{ki}^{old} + w_{ji}^{old}}{2} $

\textit{Case 2:} If $ v_{ki} < v_{ji} < w_{ki} < w_{ji}, \qquad v_{ji}^{new} = w_{ki}^{new} = \cfrac{v_{ji}^{old} + w_{ki}^{old}}{2} $

\textit{Case 3:} $ v_{ji} < v_{ki} \leq w_{ki} < w_{ji} $, if $ w_{ki} - v_{ji} < w_{ji} - v_{ki} $, then $ v_{ji}^{new} = w_{ki}^{old} $ else $ w_{ji}^{new} = v_{ki}^{old} $

\textit{Case 4:} $ v_{ki} < v_{ji} \leq w_{ji} < w_{ki} $, if $ w_{ji} - v_{ki} < w_{ki} - v_{ji} $, then $ v_{ki}^{new} = w_{ji}^{old} $ else $ w_{ki}^{new} = v_{ji}^{old} $

The overlap test and the contraction operations in the FMM clustering network are significantly different from those in the FMM classification network. The overlap test and the corresponding contraction steps of the FMNN for classification problem are built to ensure that a dimension with overlap is selected to remove overlapping regions among classes. Hence, only one dimension with the smallest potential overlap is handled. On the contrary, in FMM clustering network, the clusters are tuned to remove overlap in every dimension so that the obtained clusters are more compact \citep{Simpson93}.

Four steps above are iterated for each sample in the training data set until cluster stability is attained. Cluster stability is the case when all min and max vertices of hyperboxes are not changed during consecutive presentations of samples in the same order. By using the hyperbox generation, expansion, and contraction processes along with the maximum hyperbox size parameter ($\theta$), the algorithm can adjust the existing clusters or add new clusters to the model as data points come in without specifying the number of clusters apriori.
 
\subsubsection{Analysis of the original fuzzy min-max neural networks}
In this type of neural network, the maximum size of hyperboxes $ (\theta) $ is the most essential factor determining the number of generated hyperboxes. In general, the larger the value of $ \theta $, the fewer hyperboxes produced, and the network may show higher generality, but the overlapping regions increase, and the capability of capturing nonlinear boundaries between classes decreases. This also lowers the predictive accuracy of the trained network. A smaller $ \theta $ leads to a larger number of generated hyperboxes and potential overfitting, thus reducing the generalization ability \citep{Gabrys00, Davtalab14}. Hence, there is a trade-off between the generality and predictive accuracy of these networks.

Generally, the learning process of the FMNN is based on an online adaptation of the hyperboxes \citep{Simpson92, Gabrys00}. In other words, online learning provides FMM with the ability to create new classes and adjust the existing classes without, generally, influencing information already captured in the network. In principle, this capability allows the FMNN to add new classes and adjust the existing ones without the demand for retraining \cite{Gabrys99}. Online adaptation is the main feature in a neural network learning to address the stability-plasticity dilemma \citep{Grossberg80}.

Nevertheless, the traditional fuzzy min–max neural network has not yet dealt with several issues as follows:

\begin{itemize}
	\item \textit{Expansion problem.} The winner expandable hyperbox in the conventional fuzzy min–max learning algorithms is randomly identified in the case of existing several winner hyperboxes
	\item \textit{Hyperbox boundary problem.} Though an overlap eliminating process has been proposed, two newly contracted hyperboxes are still possible to be overlapped on the boundary edge due to the nature of the contraction formulas
	\item \textit{Problem of contraction process.} The contraction approach inadvertently removes from the two overlapping hyperboxes some unambiguous part of the sample space, while simultaneously retaining some contentious part of the sample space in each hyperbox \citep{Bargiela04}. This weakness should be analyzed in detail in subsection \ref{vwctr}.
	\item \textit{Data representation order problem.} The training step conducts a process of dynamic hyperbox creation, expansion, and contraction in the sample space when an individual training instance is presented. Therefore, the predictive accuracy depends on the presentation order of the training samples \citep{Meneganti98}, and the approach is sensitive to outliers and noise \citep{Gabrys02a}. As a result, noisy data from real world applications might cause serious stability issue when deploying the model in practice
	\item \textit{Maximum hyperbox size parameter sensitivity problem.} As analyzed above, the classification performance and generalization ability of the fuzzy min-max neural networks depends on the user-defined maximum hyperbox size threshold. This problem can be partly resolved by using an adaptive maximum hyperbox size mechanism as presented in \citet{Gabrys00}.
	\item \textit{Membership value problem.} The membership function used in the original fuzzy min-max neural networks assigns a relatively high membership degree to an input pattern being quite far from the cluster centroid as shown in \citep{Gabrys00}. It is necessary to build a membership function that monotonically decreases with the increase in the distance from the input pattern to the cluster prototype.
\end{itemize}

In conclusion, the learning algorithm introduced by Simpson builds the connections starting by the first example and then adds new hyperboxes by a process of expansion/contraction. The algorithm faces two main problems: the difficulty to determine the threshold value and the dependence of classification performance on the presentation order of the input samples \citep{Meneganti98}.

\subsection{Enhanced variants of the fuzzy min-max neural networks}
Due to the existence of considerable drawbacks in the original fuzzy min-max neural networks, many researchers have made efforts to improve this type of neural network. Enhanced versions have focused on two directions. While several studies have aimed to overcome the existing restrictions in the training algorithm of the original FMNN, other studies have changed the network architecture of the original model and proposed new structures with special neurons to handle the overlapping regions between hyperboxes belonging to different class labels. 

\subsubsection{Modified variants using hyperbox expansion and contraction procedures} \label{mfmnn-excontr}
First significant extension of the original FMNN is a general fuzzy min-max neural network designed by \citet{Gabrys00}. The GFMN is built on the basis of expansion and contraction concepts, and it can deal with both labeled and unlabeled data in a single algorithm. The architecture of GFMN almost resembles the original fuzzy min-max neural network topology except for two main alterations \citep{Gabrys00}. The first change is that the number of nodes in the input layer has been extended from $ n $ to $ 2n $. Primarily, it allows an input to be a hyperbox rather than a point in the n-dimensional space. The second modification is that the output layer has been added an extra node to represent all the unlabeled hyperboxes from the intermediate layer. This helps GFMN to deal with both supervised learning and unsupervised learning. Other changes in the GFMN in comparison with the original FMNN comprises the format of input patterns, a new fuzzy hyperbox membership, and adaptive modification of the maximum hyperbox size. Input samples of the GFMN are able to be in form of fuzzy hyperboxes or crisp points. The labeled and unlabeled input samples might be handled simultaneously. This feature enables the algorithm to be adopted for clustering, classification, or a hybrid of clustering and classification \citep{Gabrys00}. The allowable maximum size of hyperboxs, $ \theta $, can be modified gradually during the training process after each presentation of the training instance as follows: $ \theta^{new} = \varphi * \theta^{old} $, where $ \varphi $ is the coefficient responsible for the decreasing pace of $ \theta $ \citep{Gabrys00}. Two different algorithm type have been proposed to train the GFMM models \citep{Gabrys04}: an incremental learning \citep{Gabrys00} and an agglomerative learning \citep{Gabrys02a}. The incremental (online) learning is a dynamic hyperbox expansion and contraction procedure where hyperboxes are generated and tuned in the sample space after every presentation of a training instance \citep{Gabrys02a}. A general idea is the production of quite large hyperboxes in the early phases of training process and the reduction of the maximum allowable size of the hyperboxes in subsequent training executions aiming to accurately capture complicated nonlinear boundaries among various classes. Nevertheless, this learning strategy, at all incremental learning approaches, results in the input-output mapping depending on the order of presentation of the training input samples, and the algorithm is sensitive to outliers and noise \citep{Gabrys02a}. The overlapping hyperboxes are also another undesired result derived from the dynamic feature of the algorithm. The agglomerative learning introduced in \citep{Gabrys02a} is an alternative and a complimentary method to the incremental learning for an off-line training process done on a finite training sets. While the incremental learning is more appropriate for on-line adaptation and is able to tackle large training data sets, agglomerative learning represents robust behaviour in presence of outliers and noise as well as insensitivity to the order of training samples presentation \citep{Gabrys04}. In contrast to the incremental learning algorithm, the data clustering operation utilizing the agglomerative learning may be considered as a bottom-up technique where one begins with all individual samples as the set of initial hyperboxes, and larger representations of the original data groups are formed by aggregating smaller clusters (hyperboxes) \citep{Gabrys04}.

With the aim of generating good classifiers, \citet{Gabrys04} analyzed five different algorithm-independent model generation schemes from the GFMM neural network using the agglomerative learning algorithm. These methods include construction of the predictive models using full training data set, employing a k-fold and multiple 2-fold cross-validation along with different pruning procedures, and an ensemble of various GFMM classifiers at the decision or model level. He claimed that the method of generating the GFMM model using base learning algorithms without any hyperbox pruning steps is swift and able to adapt to the changing environment. However, it is likely to be overfitted and exhibit a poor generalization performance. In the case that the classifiers can be built in the off-line modes, the techniques based on the combination of multiple cross-validation and pruning procedures or the ensemble of base classifiers tend to yield a better classification accuracy.

An interesting property regarding the use of hyperboxes as inputs for the general fuzzy min-max neural network is the capability of handling missing values in the data. One of the most common methods of tackling missing values is to replace them with estimated values such as the mean value computed from all patterns. Nevertheless, this makes data set no longer a good representation of the problem and may result in bad solutions \citep{Berthold98}. \citet{Gabrys02c} introduced a method of dealing with missing data within the classification algorithm automatically by employing the general fuzzy min–max neural network model. The GFMN represents the missing features as real valued intervals being able to get the whole range of possible values. In other words, if the value of the $ i^{th} $ feature is missing, the lower bound of hyperbox on the $ i^{th} $ dimension is assigned to one and its upper bound receive the value of zero. This operation makes the hyperbox membership associated with the missing feature to be one, and so it does not lead to a decrease in the overall membership value. This substitution also makes sure that the neural network structure would not be modified when handling missing dimensions. The only changes of the training algorithm relate to the way of performing the overlap test and the usage of assumption that the missing features are possible to get all values \citep{Gabrys02c}. The overlap test is conducted after each hyperboxes updating operation for only hyperboxes in which their value of maximum point is larger than or equal to their value of minimum point on every dimension. The ultimate model trained on incomplete data may contain a set of hyperboxes for which some dimensions are missing or not established. Empirical results illustrated the effectiveness and performance of the GFMN in dealing with missing data.

In another study, \citet{Kim04, Kim05} proposed a weighted fuzzy min–max (WFMM) neural network for the sample classification and feature extraction problems. They introduced a new membership function and expansion scheme by considering a weight factor for each dimension of a hyperbox. This improvement makes the WFMM less sensitive to the unusual or noisy features in a data set in comparison with the FMNN \citep{Kim04}. Therefore, the WFMM may handle better data sets containing highly uneven distribution of features or noisy features \citep{Zhang11}. The use of the weight factor for each dimension of the hyperbox aims for membership function being capable of taking into consideration both the happening of samples and the importance level of each input feature \citep{Kim06}. The architecture and learning algorithm of the WFMM are similar to the FMNN except for some changes in the hyperbox membership function and expansion mechanism. In the WFMM, each dimension in the original membership function shown in Eq. \ref{eq5} is multiplied with a weight factor to take the relevance of each feature into consideration. These weights are initialized to 1.0 when a new hyperbox is generated, and their values are updated based on new data coming. To reduce the influence of unusual or noisy samples, \citet{Kim04} altered the expansion procedure such that the membership values gradually rise for these patterns. In the later research, \citet{Kim06} introduced an enhanced version of WFMM neural network with changes in the connection weight of features, membership function, and a new contraction method including the weight updating mechanism. Like FMNN, the learning algorithm of the modified WFMM neural network also includes three operations: hyperbox expansion, overlap test, and hyperbox contraction.

With the aim of handling a number of drawbacks of the original FMM neural network, \citet{Mohammed15} introduced an enhanced fuzzy min–max neural network (EFMNN). They retained the structure of original FMNN and only modified the training algorithm. Authors analyzed three main shortcomings that may influence the FMNN's performance. The first drawback is that hyperbox expansion condition can lead to the increasing of the overlapping region between different classes as several dimensions surpass the expansion coefficient but sum of all dimensions still falls in the allowable limit. The second shortcoming is that the original FMNN only use four cases to detect the overlapping area between two hyperboxes belonging to different classes, so some other overlapping cases might not be uncovered by these tests as shown in \citep{Mohammed15}. The last disadvantage of the original FMNN is that the subsequent hyperbox contraction procedure based on four cases is not strong enough to handle all overlapping regions in practice. To deal with the first limitation, authors employed a new constraint, in which each dimension of the $ j^{th} $ hyperbox is examined individually to identify if it is larger than the expansion parameter $ \theta $. This mechanism is the same as that proposed in the GFMM neural network \citep{Gabrys00}. Regarding the second shortcoming, \citet{Mohammed15} introduced nine cases for overlap test and corresponding contraction operations instead of four cases in original FMNN.

In the later study, \citet{Mohammed17a} claimed that the expansion rule limitation has still not been resolved thoroughly in the EFMNN. It is obvious that the expansion process in the original FMNN as well as the EFMNN concentrates on choosing the hyperbox with the largest value of membership function to become the sole winner in a set of hyperboxes. This mechanism can result in the production of numerous small hyperboxes located in the vicinity of the winning hyperbox when the winner cannot meet the expansion criterion; therefore increasing the network complexity. Based on this analysis, \citet{Mohammed17a} proposed a novel approach, known as the K-nearest hyperbox expansion rule (KN-EFMNN), to decrease the network complexity by reducing the number of small hyperboxes within the surrounding area of the winning hyperbox during the training phase. First of all, the hyperbox with the highest membership value is chosen. All its dimensions are then examined against the expansion conditions. If there is any expansion criterion violation of the winning hyperbox, the next nearest hyperbox is selected to execute the same examining process. If all K-nearest hyperboxes cannot satisfy constraints, a new hyperbox is generated to include the input pattern. A set of K-nearest hyperboxes with the same class label is selected for finding the ultimate winning hyperbox for the hyperbox expansion procedure, which leads to reducing of the number of small hyperboxes \citep{Mohammed17a}. Like other classifiers, the EFMNN is sensitive to noise in the data sets. If noise exists in the training data, it will cause the production of numerous `noisy' hyperboxes; thus degrading the performance of the model. Besides the K-nearest hyperbox selection rule, \citet{Mohammed17b} introduced a useful strategy to deal with noise, i.e., pruning. This strategy aims to determine and delete hyperboxes giving low accuracy, which regularly are produced because of outliers or noise. Hence, the available data patterns should be split into training, prediction, and test sets. In the learning process, the training set is utilized to formulate hyperboxes, while the prediction one is then used to prune the trained network architecture \citep{Mohammed17b}. Although the EFMN and its improved versions outperform the original FMNN, the use of the contraction step for the training process may lead to classification error as analyzed in subsection \ref{vwctr}.

Though hyperbox fuzzy sets can explain their predictive results based on rules extracted directly from the min-max values, the interpretability of classification is usually not friendly for users. It is due to the fact that the rule sets become extremely complex in case of a large number of hyperboxes and high dimensions. Therefore, it is desired to construct the rule extraction methods from hyperboxes to form a compact rule set, which is capable of accounting for the predictive results. As a result, \citet{Quteishat08b, Quteishat10} introduced a two-stage pattern classification system using a modified FMNN in the first stage and a rule extraction procedure in the second stage. \citet{Sonule17} described an enhanced FMNN model with an ant colony optimization based rule extractor for decision making by a list of rules. In the two-stage sample classification and rule extraction system using different FMNN models, a data set is divided into three sections, which are training set, prediction set, and test set. Training set is applied for training the FMNN, prediction set is responsible for hyperbox pruning and rule extraction, while test set is employed to assess the performance of classification systems. The first stage of the two-stage classifier comprises FMM training and pruning, while the second stage is a process of rule extraction. The learning process in this system is the same as original FMNN. After the network is trained, a pruning operation is implemented to lower the number of generated hyperboxes.

The pruning technique used by \citet{Quteishat08b} relied on a confidence factor similar to Carpenter's work \citep{Carpenter95}. Prior to performing the pruning procedure, the confidence factor ($ CF_i $) of each hyperbox $ B_j $ is computed according to its usage frequency and predictive accuracy over the prediction set. The confidence factor has likelihood of identifying good hyperboxes being ones which are usually utilized and generally give accurately predicted results or ones that are rarely employed but highly accurate \citep{Quteishat08b}. This confidence factor is then associated to the fuzzy if–then rule extracted from the corresponding hyperboxes with the aim of indicating its certainty level to predicted results. After confidence factor of each hyperbox is calculated, hyperboxes with their values of confidence factor smaller than a user-defined threshold are removed, and the remaining hyperboxes are utilized in stage 2 of the classification model. \citet{Quteishat08b} conducted the rule extraction process once the first stage is finished.

To enhance the prediction performance, however, in the later work, \citet{Quteishat10} conducted several preprocessing steps before extracting rules from the modified FMNN, i.e., open hyperbox generation and genetic algorithm (GA) rule selection. This method is abbreviated MFMNN-GA. An open hyperbox is a hyperbox with at least one dimension undefined by its minimum and maximum points, whereas a closed hyperbox is the one with all its minimum and maximum vertices defined. This mechanism is similar to the method proposed by \citet{Gabrys02c} for handling missing values in the input dataset. The non-declared dimension is considered as the ``don't care" dimension and fully contain the specific ``don't care" feature of the input space \citep{Quteishat10}. In the open hyperbox generation, all possible combinations of open hyperboxes for each hyperbox are checked. After that, all hyperboxes (closed and open) are put through an evolution process by using the GA \citep{Quteishat10}. The GA is in charge of evolving and choosing a set of hyperboxes that are able to generate a good predicted result with a small number of features. This operation contributes to reducing the complexity of fuzzy rules generated from hyperboxes.  In the prediction stage of the original FMNN, the sample is assigned to the class represented by the hyperbox having the largest membership value. However, there exists the case that many hyperboxes have high membership values for a new input sample. Hence, the winner-takes-all technique in identifying the winner hyperbox could result in incorrect prediction results. To cope with this issue, \citet{Quteishat08b} proposed a combined method of the membership function and Euclidean distance to predict the output of classification system. Hyperboxes with membership values larger than a user-defined threshold are selected and stored to a pool. After that, the Euclidean distances between the input sample and the centroid of the chosen hyperboxes are identified, and the hyperbox with the shortest distance value is chosen as the winner. This modification enhanced the classification results in the case that the constructed network contains only the relatively small number of hyperboxes \citep{Quteishat08b}. However, these proposed classifiers have lost single pass-through online adaptation power of FMNN \citep{Forghani15}.

In another study, \citet{Sonule17} proposed an enhanced fuzzy min–max neural network model with a rule extractor based on ant colony optimization (EFMNN-ACO) for classifying the data samples and decision making by rule-list. The output of the EFMNN \citep{Mohammed15} is used as input of a rule extractor based on AntMinerPlus algorithm to build a graph. Hyperboxes generated by the neural network are pruned and the rule extraction process occurs after pruning of the path selection graph. After training the enhanced fuzzy min-max neural network, its outputs are used to construct a graph of AntMinerPlus algorithm. This algorithm forms the graph by using min-points, max-points and unit matrix of the classified hyperboxes. Then the algorithm will perform a process of extracting the rule-list and pruning rules by finding the paths on the graph. The experimental result of \citet{Sonule17} showed that the quality of rules is consistent and the number of obtained rules is reduced due to the optimization algorithm. One of the strong points of this method is that it can optimize simultaneously a list of rules instead of separate rules. However, training time of the system is longer than other classifiers since there are more cases considered and the construction of graph in the AntMinerPlus algorithm takes a long time.

As mentioned in section \ref{fmnn}, the original fuzzy min-max neural network encounters several problems in the expansion and contractions steps as well as the overlapped boundary after doing contraction. Hence, \citet{Liu17} proposed a modified fuzzy min–max neural network for data clustering (MFMC) to deal with those issues. The authors still maintained the architecture of the original FMNN for clustering, but they introduced several changes for training process such as a hyperbox selection rule, a reservation rule and a hyperbox entropy measure for contraction process, and a parameter $ \delta $ to avoid boundary overlapping. When a new training instance is presented, the membership values of all hyperboxes are computed. If no existing hyperboxes contain the input sample, a hyperbox with the largest membership degree is chosen as the winner hyperbox and the expansion process is conducted. Nevertheless, there may be numerous winner hyperboxes in the case of hyperboxes with the same membership values. In this situation, the original fuzzy min-max clustering neural network will select the final winner hyperbox randomly. In contrast, the MFMC calculates a centroid for each winner hyperbox to cope with this issue \citep{Liu17}. After all centroids of winner hyperboxes are calculated, the distances between input sample and these centroids are taken into account. Finally, the hyperbox with minimum distance value is the ultimate winner hyperbox to continue with expansion criteria checking. If there exists an overlapping region, the contraction procedure is executed. Different from the original fuzzy min–max learning algorithms which only separate the overlapping area into half, \citet{Liu17} introduced the method to reserve the hyperboxes with higher performance when performing the contraction operation. The hyperboxes which include more data with smaller size are regularly considered to be the ones with higher performance. The reservation ability maintains the good structure of the whole learning algorithm, avoid disturbing the size of hyperboxes because of noise, and refine the robustness for the algorithm \citep{Liu17}. The authors introduced a formula called the hyperbox entropy $(HE)$ to assess the performance of each overlapping hyperbox. The hyperbox with higher value of $ HE $ is considered to be a better hyperbox. The contraction process is changed by integrating the value of hyperbox entropy. 

Although \citet{Liu17} made some changes to handle drawbacks of the original fuzzy min-max clustering neural network, there are still some existing unsolved problems. The method using only four test cases for the overlapping checking as in the original FMNN is not adequate to verify all situations happening in the practice as described in \citep{Mohammed15}. The hyperbox selection rule only solves the case of many winner hyperboxes. If the ultimate winner hyperbox cannot be expanded, it still creates a new hyperbox to contain the input training sample. This issues can lead to the generation of numerous hyperboxes with the small size, which causes the network complexity problem. In addition, Liu's proposal still implements the hyperbox contraction operation, which is more likely to result in potential errors as explained in next subsection \ref{vwctr}. Moreover, the performance of model depends highly on the value of the parameter of maximum hyperbox size $ \theta $.

\citet{Azad16} argued that the series of expansion and contraction steps lead to the change in the sizes of hyperboxes and affect the performance of the predictive model. Hence, authors proposed to use the genetic algorithms to optimize the min-max values of hyperboxes generated by the original FMNN. Hyperboxes are selected to perform the processes of crossover and mutation to produce better offspring. The new offspring are verified for possible overlaps and the contraction process is conducted if there is any existing overlapping region among hyperboxes representing different classes. Finally, the best individuals in the evolving process will replace the current hyperboxes. In later work, \citet{Azad17} used the particle swarm optimization instead of genetic algorithms, and they obtained a better performance.

To improve the performance of the original fuzzy min-max neural network for clustering problems, \citet{Seera15} proposed to integrate the centroid information of data samples into each hyperbox to construct a modified fuzzy min-max neural network for data clustering (MFMMC). With the use of centroid, the operations of hyperboxes in the learning process are changed. In practice, the centroid of the hyperbox is likely to be outside the hyperbox structure when performing the contraction process, so a rule to check the hyperbox centroid was proposed. This rule forces the centroid to be inside the hyperbox. If the centroid is outside the hyperbox after expanding the current hyperbox to cover the new pattern and then doing the contraction step, the structures of two contracted hyperboxes are brought back to those before the process of expansion and contraction is performed. After that, a new hyperbox is created to contain the input instance. When the clusters are formed, the centroid information and the cophenetic correlation coefficient metric \citep{Mirzaei10} are used to evaluate the quality of created clusters.

An attribute of the fuzzy min–max neural networks mentioned above is that all the input variables for training and classifying processes are continuous numerical values \citep{Castillo12}. When the categorical variables are presented, they can be substituted by numerical values and treated as continuous values. However, there is no meaningful correspondence between the continuous values created by this method and the original categorical ones \citep{Brouwer02}. Hence, it is expected to form a new method for handling categorical variables. \citet{Castillo12} proposed a new method to extend the GFMN inputs for dealing with discrete variables (GFMN-CD1) by formulating the distance between the categories of categorical variables. \citet{Shinde16} also introduced another method to refine the GFMN for categorical input data (GFMN-CD2). In \citeauthor{Castillo12}'s proposal, each categorical variable $ a_i $ is one point of the p-dimensional space $ \mathbf{R}^p $ represented by a vector $ a_i = (a_{i1}, a_{i2},..., a_{ip}) $ such that $ a_{i1} + a_{i2} + ... + a_{ip} = 1 $. The distance between two categorical variables can be computed according to two ways, i.e., Euclidean distance or logarithmic distance. Similar to the hyperbox in the numerical dimensions, each hyperbox fuzzy set in the $ i^{th} $ categorical dimension is determined by two categories $ e_{ji} $ and $ f_{ji} $ with a full membership \citep{Castillo12}. The authors defined a new membership function for both numerical and categorical variables. They also extended the architecture of the GFMN \citep{Gabrys00} to include both numerical and categorical inputs. Compared to GFMN, besides $ 2n $ numerical variable nodes, there are $ r $ additional nodes representing the categorical inputs, and each categorical node is connected to each hyperbox in the second layer by two connection weights - one for each category $ e_{ji}, f_{ji} $ defining boundaries for the $ i^{th} $ categorical variable of the hyperbox $ B_j $. The training process for this improved version also consists of four steps similar to the GFMN. In the architecture proposed by \citet{Castillo12}, each input neuron representing categorical data is linked to each hyperbox in the second layer by two connections corresponding to two categories defining the hyperbox. \citet{Shinde16} proposed a simpler architecture by adding a set of binary strings, where each binary string represents a given discrete attribute, to each hyperbox fuzzy set of the network. Each neuron for categorical input is connected to each hyperbox in the intermediate layer by only one connection. If the categorical attribute includes $ m $ values, then it will be represented by a m-bit binary string, and the bit position corresponding to the current value of that discrete variable is 1 while remaining bit positions are 0s. The authors introduced new membership functions with the logical bitwise `and' and `or' operators conducted on two binary strings for categorical variables. The learning algorithm is similar to the GFMN with only small changes to accommodate discrete data.

\subsubsection{Variants with novel training mechanism without using the hyperbox contraction procedure} \label{vwctr}
Two main shortcomings of the learning algorithm developed by Simpson are the sensitivity to the input data presentation order and the difficulty in finding the maximum hyperbox size. Hence, \citet{Meneganti98} introduced a novel learning algorithm while keeping unchanged the structure of the fuzzy min-max neural network. The algorithm begins with building the minimum size hyperboxes covering all patterns of the same class, in which each class is represented by only one hyperbox. Then, a five-step process including partition, decomposition, recomposition, removal, and expansion is used to minimize the number of hyperboxes while maximizing their dimensions. Authors also proposed a new method to split intersecting hyperboxes, which represent different classes, based on the similarity. A new membership function using Gaussian function was introduced to evaluate the similarity level of each pattern against each hyperbox. This learning algorithm makes the classification performance independent from the sample presentation, and the algorithm does not employ any threshold. In the later study, \cite{Tagliaferri01} claimed a top-down fuzzy min-max (TDFMM) classifier by simplifying the complexity of the hyperbox splitting approach proposed by \citet{Meneganti98}. Based on the TDFMM algorithm, they developed a top-down fuzzy min-max regressor to deal with the regression problem. First of all, clustering methods are deployed to assign the labels to input patterns. Next, the TDFMM algorithm is utilized to build a hyperbox. The authors introduced a new membership function to compute the output of the network from activation functions of hidden hyperboxes. The parameters of the membership function are adjusted using the back-propagation algorithm.

For the design of classification systems, in addition to generalization ability and noise robustness, a high degree of automation is also one of the most critical properties of data driven modeling tools \citep{Rizzi00}. As a result, constructive learning algorithms are indispensable to enhance the degree of automation, make the system work in a self-governing way, and automatically establish structural parameters during training. Among neuro-fuzzy machine learning algorithms, the fuzzy min-max networks proposed by \citet{Simpson92} have the potential to be trained in a constructive way \citep{Rizzi02}. However, the original Simpson's training algorithm for the FMM neural network depends excessively on pattern presentation order and on position as well as size of the hyperboxes generated during training. These parameters impose the same condition on covering resolution in the whole input space. This results in reducing the generalization ability of the neural model. Aiming to tackle these inconveniences, two new learning algorithms were introduced by \citet{Rizzi98}, i.e., the adaptive resolution classifier (ARC) and pruning ARC (PARC) algorithms. In the improved version of these algorithms, \citet{Rizzi00, Rizzi02} suggested a feasible enhancement of the training process by utilizing a new cutting strategy in order to handle recursively hybrid hyperboxes, namely R-ARC/R-PARC. This procedure cuts a hyperbox into two halves by a hyperplane, perpendicularly to one coordinate axis and corresponding to an appropriate point. The main aim of these operations is to yield hyperboxes covering a set of samples belonging to only one class. These hyperboxes would be considered as \textit{pure}, otherwise hyperboxes are said \textit{hybrid}. According to the principles of learning theory, pure hyperboxes of the same class are fused if no overlaps occur with both pure hyperboxes and hybrid hyperboxes of different classes \citep{Rizzi00}. After cutting a hybrid hyperbox, a new coverage $ L $ is formed, a fusion operation is carried out. Then, the net generation procedure is used to find an optimal fuzzy min-max neural network. The computational expense of the fusion operation is expensive with the complexity being $ O(q^3) $ \citep{Rizzi02} ($ q $ is the number of pure hyperboxes in the coverage $ L $), so the effectiveness of ARC algorithm is dependent mostly on how many times a fusion operation is carried out and on the number of hyperboxes before the fusion process begins. Hence, a method to improve the ARC algorithm is reduction of the total number of nets created during training. \citet{Rizzi98} introduced a pruning version of ARC with two subsequent actions: doing an ARC operation without the fusion procedure and doing a pruning procedure. The pruning activity terminates if the removal of some pure hyperboxes lead to the incomplete actual coverage \citep{Rizzi02}. The pruning operation is to execute an automatic optimization of the network architecture and to attain noise robustness. The key idea of further enhancement of ARC/PARC algorithm is to isolate recursively the non-overlapping regions of the training set. Therefore, \citet{Rizzi02} proposed a recursive ARC algorithm.

\begin{figure}[!ht]
	\centering
	\includegraphics[width=0.45\textwidth]{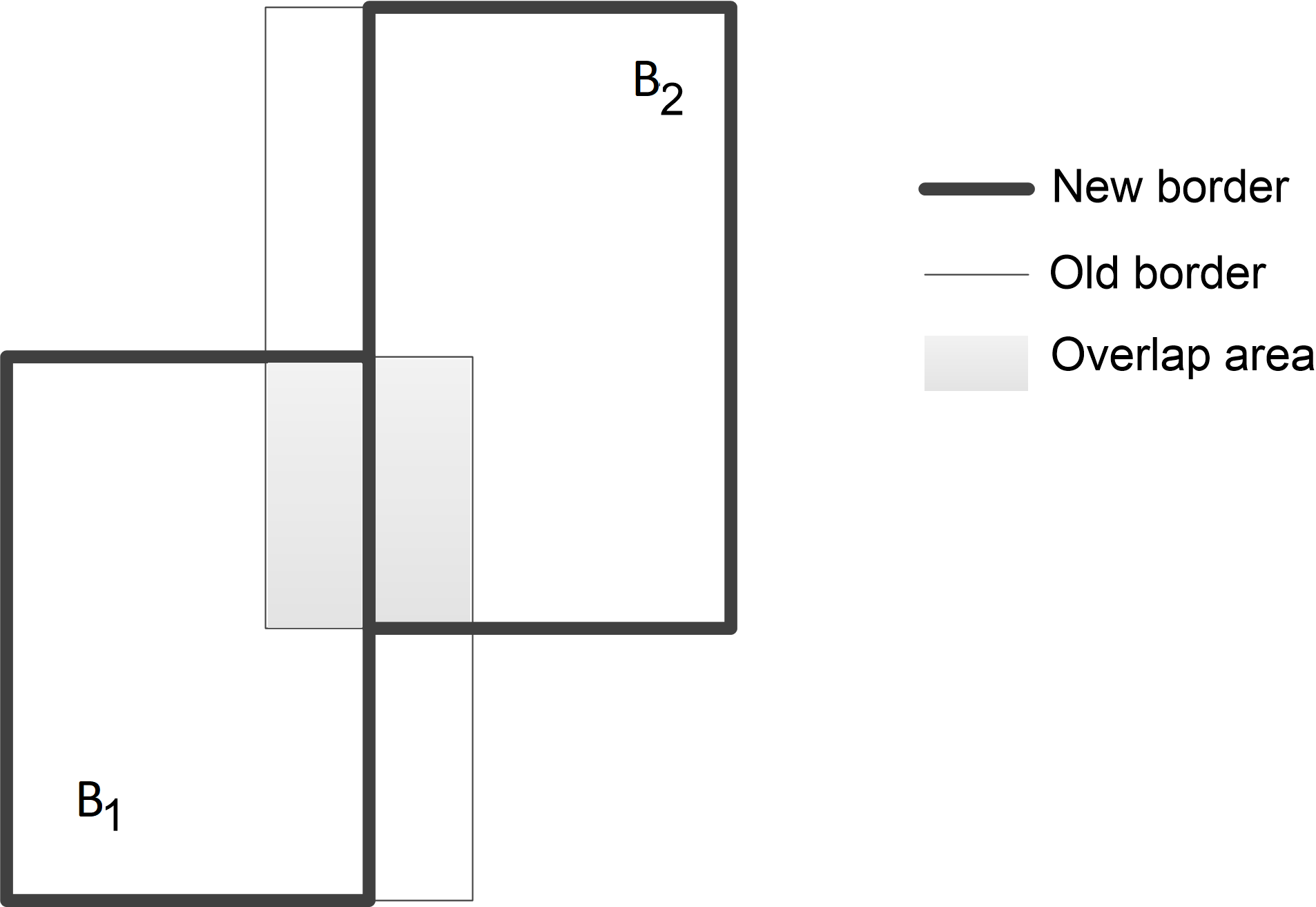}
	\caption{Contraction of hyperboxes $ B_1 $ and $ B_2 $ with elimination along one coordinate}
	\label{fig6}
\end{figure}

The contraction manner in Simpson's FMM \citep{Simpson92} and general FMM networks \citep{Gabrys00} has a drawback in which it removes from the two overlapping hyperboxes several zones of the sample space that was unambiguous while simultaneously maintaining some contentious part of the pattern space in each hyperbox. For instance, as shown in Fig. \ref{fig6}, some part of the original hyperbox $ B_1 $ now are contained in $ B_2 $ after the contraction procedure and likewise the original hyperbox $ B_2 $ has an overlapping region with contracted hyperbox $ B_1 $. Hence, the overlap removal algorithm yields errors during training process. Another issue derived from the contraction process is that it unnecessarily removes parts of the original hyperboxes \citep{Davtalab14}. The removal of these parts of hyperboxes implies that the contribution of the data located in these regions to the network training process is nullified. This problem is more likely to cause an irreversible loss if the neural network is trained using only one pass through the training data \citep{Bargiela04}. In some cases, additional hyperboxes can be generated to include the deleted portions of the original hyperboxes, for example, the use of adaptive hyperbox size with multiple data presentations in the GFMM neural network \citep{Gabrys00}. This leads to the increase in the number of hyperboxes and the reduction in the interpretability of classification results. To tackle these problems, \citet{Bargiela04} introduced an inclusion/exclusion fuzzy hyperbox classification network (IEFCN), where the overlapping regions of the sample space are explicitly represented as exclusion hyperboxes. 

Unlike the classical FMM neural network, the inclusion/exclusion fuzzy hyperbox classification model employs two kinds of hyperboxes, which are the inclusion and exclusion ones, and does not use contraction to eliminate overlaps. The inclusion hyperboxes cover the input samples of the same class, whilst the exclusion hyperboxes contain other overlapped input samples. By the use of combination of inclusion and exclusion hyperboxes, the learning phase from three steps, which are expansion, overlap test, and contraction in classical FMM, is reduced to two steps (expansion and overlap test) \citep{Davtalab14}. Similar to GFMN, the input layer of inclusion/exclution network also has the form of hyperboxes in the n-dimensional sample space. In the middle layer, there is a set $ e $ of exclusion nodes constructed adaptively. The minimum and maximum points of the exclusion hyperbox are determined when there is a positive value returned from the overlap test of two hyperboxes showing different classes. If the new exclusion hyperbox accommodates any of the exclusion hyperboxes created in the previous step, the included hyperboxes are removed from the set $ e $ \citep{Bargiela04}. The output layer has $ p + 1 $ nodes ($ p $ is the number of classes) with node $ c_{p + 1} $ representing the exclusion hyperbox class. 

The efficiency of the IEFMN is influenced when the number of exclusion hyperboxes is equivalent to that of inclusion hyperboxes, where percentage of data samples categorized as void class can become unacceptably high \citep{Nandedkar07b}. Therefore, \citet{Nandedkar04, Nandedkar07b} proposed a fuzzy min-max neural network with compensatory neurons (FMCNs), which are generated dynamically during learning process, to tackle hyperbox overlaps and containment problems. The FMCN is also known as the Reflex Fuzzy Min Max Neural Network (RFMN) since the concept of compensatory neurons (CNs) originates from the reflex system of human brain. The use of compensatory neurons contributes to removing the contraction process for the labeled hyperboxes and controlling membership in the overlapped region. Another drawback of the FMNN is that their performance depends mostly on the maximum allowable hyperbox size $ \theta $. To deal with this limitation, \citet{Rizzi98, Rizzi02} introduced a recursive training algorithm using hyperbox cut concept to avoid $ \theta $ initialization and sample order dependency. Nevertheless, these results have been gained by performing a recursion procedure, so a single pass-through online adaptation ability is lost. Meantime, the FMCN still maintains a single pass-through and online learning capability \citep{Nandedkar07b}. The number of nodes in the input layer of the FMCN is equivalent to the number of dimensions of input vector $ X_h $. The intermediate layer neurons and output layer neurons are separated into two parts: 1) classifying neuron (CLN) segment and 2) reflex section. The classifying part is used for computing membership values for various classes.

The Reflex section comprises two subsections, Overlap Compensation Neuron (OCN) part and Containment Compensation Neuron (CCN) part \citep{Nandedkar06b}. This section is active whenever a pattern is inside the class overlapping region. Each node in the intermediate layer presents a n-dimensional hyperbox, which is dynamically generated during the learning process. Each hyperbox node in the set of CLN nodes is formed when the training instance presents a class not been met so far or existing hyperboxes of that class are not able to enlarge further to include it \citep{Nandedkar07b}. Hyperboxes located in the second layer of OCN and CCN segments are built whenever the overlap or containment issue occurs in the network respectively. Each OCN-type node generates two outputs corresponding to two overlapping classes. OCN is active only if the pattern falls inside the overlapping area. An overlap compensation neuron renders a hyperbox with size equal to the overlapping area between two hyperboxes belonging to different classes. The containment compensation neuron is trained to handle the case that a hyperbox of one class is accommodated fully or partly in a hyperbox of another class. The CCN also renders a hyperbox with size equal to the overlapping area of two hyperboxes. This neuron is activated only if the pattern belongs to the containment zone as well. Each CCN-type node has just one ouput and its output is linked to the class of the hyperbox covering another hyperbox. The FMCN eliminates the usage of contraction procedure, thus it avoids errors resulting from contraction operation. FMCN may maintain the knowledge of the already trained samples more effectively in comparison with FMNN and GFMN due to the fact that already produced hyperboxes are not contracted. The accuracy of FMCN is better in single pass through the data \citep{Nandedkar07b} as it is able to approximate the complicated structure of data more properly thanks to the efficient capability of tackling hyperbox overlap and containment. Another benefit of FMCN is that it can evade the dependency of systems on the learning parameter compared to FMNN and GFMN. Furthermore, the FMCN is also robust to the noise. 

\begin{figure}
	\centering
	\includegraphics[width=0.4\textwidth]{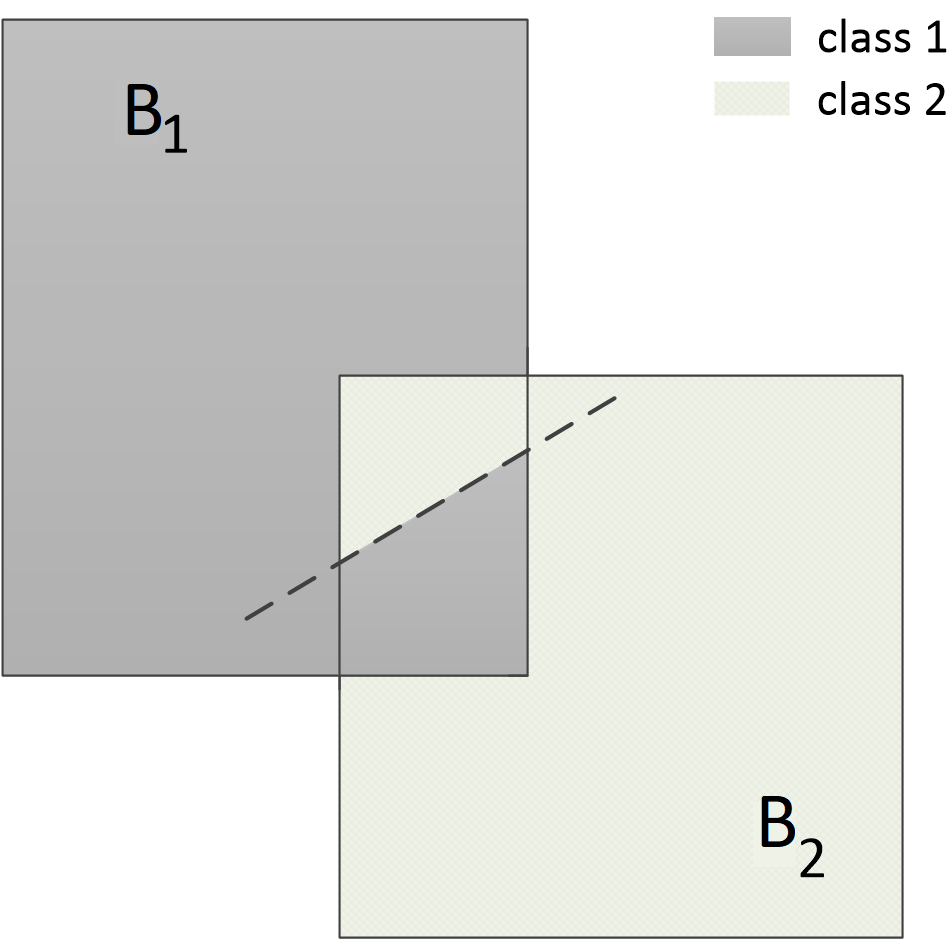}
	\caption{The way of classifying sample spaces by FMCN}
	\label{fig9}
\end{figure}

In spite of above strength points, the FMCN also faces the following drawbacks. It does not implement a suitable membership function for overlap compensatory neurons, thereby it is unable to precisely classify high percentage of patterns present in overlapping areas. As described in Fig. \ref{fig9}, patterns falling inside the overlapping region and being closer to hyperbox $ B_1 $ (the light color regions within the $ B_1 $) are assigned to the class label of $ B_2 $ and vice versa \citep{Davtalab12}. Furthermore, FMCN only deals with simple and containment overlaps, so there would be types of overlaps such that the algorithm cannot remove them, e.g. two hyperboxes crossing each other. Another case is that hyperboxes including just one point, when contained in any hyperbox representing another class, will not expand in next steps \citep{Davtalab14}. In the learning algorithm, if any overlap between expanded hyperbox and other hyperboxes representing other classes exists, a compensatory neuron is added to the network. Hence, duplicate nodes are more likely to be generated. To overcome these disadvantages, \citet{Davtalab12} invented a new fuzzy min-max model based on fuzzy min-max neural network with modified compensatory neurons. This classification model is also an online, single-pass and supervised learning method, but the authors made several changes in the structure of FMCN and training phase to reduce time and space complexity. The new structure reduces generating and storing useless hyperboxes during the training process, so it results in a faster classifier. The authors suggested using a new membership function of compensatory nodes to deal with overlapping areas and increase classification accuracy. Experimental results indicated that the new membership function tackled the first drawback of FMCN mentioned in above section.

To construct a system being capable of learning from the mixture of labeled and unlabeled data like the GFMN, \citet{Nandedkar06b} introduced a modified version of reflex fuzzy min-max neural network (RFMN) by adding floating neurons (RFMN-FN). Like the original RFMN, the RFMN with floating neurons also uses compensatory neurons to uphold the hyperbox dimensions and control the membership in the overlapping areas among hyperboxes representing different classes. Floating neurons are added to the network to keep the unlabeled hyperboxes and stop them from contributing to the classification. The RFMN-FN may be trained by two methods, which are supervised learning and semi-supervised learning \citep{Nandedkar06b}. In the mode of semi-supervised learning, after accomplishment of training, many hyperbox fuzzy sets are more likely to remain unlabeled because of the shortage of evidence for these sets. Neurons defining such hyperboxes are called ``floating neurons", and they are prevented from contributing to the output. These neurons may be labeled and influence the output of model if evidence for a suitable class is found out later on \citep{Nandedkar06b}. To handle labeled, unlabeled and partly labeled data sets, \citet{Nandedkar07a} proposed a general reflex fuzzy min-max neural network (GRFMN). However, the input pattern to this network is in numeric form.

In the later studies, \citet{Nandedkar09, Nandedkar08, Nandedkar06a} extended this network for both granular data and numeric data by representing input sample as a hyperbox $ X_h = [X_h^l, X_h^u] $. GFMN does not make use of contraction process, but instead, in GRFMN a reflex mechanism is used to tackle the hyperbox overlap and containment issues. The reflex mechanism comprises compensatory neurons added dynamically to the network during training procedure \citep{Nandedkar07b}. These neurons control the membership value in the overlapping areas. In terms of partly labeled data set, GFMN training algorithm labels an unlabeled hyperbox upon there is a single labeled pattern falling inside it, meantime the GFMN's learning algorithm eliminates the overlap of unlabeled hyperboxes with all other hyperboxes. In constrast, GRFMN enables the unlabeled hyperbox to overlap with labeled hyperboxes, and it only performs a contraction process if there is an overlap between two unlabeled hyperboxes \citep{Nandedkar07a}. GRFMN implements the same contraction method as GFMN \citep{Gabrys00} for the overlap amongst unlabeled hyperboxes, in which overlapped hyperboxes are contracted along a dimension with minimal overlap. It is noted that compensatory neurons are not produced for GRFMN in the pure clustering problem. The training algorithm of GRFMN is of incremental type, in which data patterns in the training set are subsequently provided to the algorithm. The network attempts to learn the labeled input samples by finding the most suitable hyperbox among the existing hyperboxes of the same class to cover those patterns. If no hyperbox is found, a new hyperbox is added to the network. When an unlabeled sample is presented to the learning process, the network should attempt to contain it in one of the existing labeled or unlabeled hyperboxes, otherwise produce a hyperbox without label \citep{Nandedkar07a}. If any overlap or containment occurs during hyperbox expansions, a corresponding compensatory neuron is added to the network. Contraction operation proposed by \citet{Gabrys00} is run if there is any overlap between unlabeled hyperboxes \citep{Nandedkar08}.

\citet{Zhang11} introduced a data core based fuzzy min–max neural network (DCFMN) with new structure and learning algorithm. Like FMCN \citep{Nandedkar07b}, DCFMN also removes the contraction procedure to reduce classification errors. While FMCN employs the overlapped compensatory neuron and the containment compensatory neuron to deal with the overlapping region among hyperboxes from different classes, only one type of neurons, i.e., overlapping neurons (OLNs), is implemented to handle the overlap and containment issue in DCFMN. As for FMCN, the extended hyperbox is not overlapping with any prior hyperbox of different class, so the number of hyperboxes in FMCN may be large and this results in a complex network and waste of time. To overcome this drawback, DCFMN allows the hyperboxes to be expanded to overlap repeatedly with the previous hyperboxes \citep{Zhang11}. Hence, the number of hyperboxes in DCFMN is lower than that in the FMCN and it uses less computation time. Furthermore, only three kinds of overlap were handled in FMCN, whereas the DCFMN tackles all kinds of overlap \citep{Zhang11}. In addition, the authors proposed a new membership formula of classification neurons formed based on the characteristics of data and the impact of the noise which makes DCFMN more robust as Eq. \ref{eq49}. The membership formula of OLNs built on the basis of the relative position of data within the hyperbox may remove the impact of different values of data and data normalization. A novel training and classifying algorithm is introduced to make the resulting classifier faster and more accurate. The learning algorithm for DCFMN can be divided into two parts: 1) creation and expansion of hyperboxes 2) overlap test, and construction of OLN if needed. The overlap test takes place after the process of initializing and expanding hyperboxes over all samples in the training data set. In spite of these advantages, DCFMN cannot accurately classify high proportion of patterns being located in the overlapping areas, and also is unable to classify all learning examples properly either \citep{Davtalab14}. Moreover, the architecture proposed by \citet{Zhang11} can only handle numeric data. This method can be extended as GRFMN to deal with granular data.

\begin{equation}
\label{eq49}
b_j(X_h) = \min \limits_{i = 1}^{n}\left(\min(f(x_{hi} - w_{ji} + \epsilon, c_{ji}), f(v_{ji} + \epsilon - x_{hi}, c_{ji})))\right)
\end{equation}
where $ \epsilon $ represents noise, $ c $ is a difference between the data core in the hyperbox and the geometric center of the respective hyperbox, and $ f $ is a ramp threshold function. $ \epsilon $ is in charge of restraining the impact of noise, and its value changes depending on various noise values. The ramp threshold function $ f(a, b) $ is given as Eq. \ref{eq51}.

\begin{equation}
\label{eq51}
f(a, b) = \begin{cases}
e^{-a^2 \cdot (1 + b) \cdot \lambda}, & \mbox{if } a > 0, b > 0 \\
e^{-a^2 \cdot (1 - b) / \lambda}, & \mbox{if } a > 0, b < 0 \\
1, & \mbox{if } a < 0
\end{cases}
\end{equation}
where $ \lambda $ is employed to control the descending pace of the membership function.

As mentioned above, FMCN and DCFMN deploy compensatory neurons to deal with the overlapping regions among hyperboxes of different classes. Nonetheless, these networks are unable to classify a high ratio of patterns positioned in overlapping areas accurately and also face several structural issues in their training algorithms leading to increased complexity but decreased efficiency \citep{Davtalab14}. To increase the classification accuracy in the boundary areas, \citet{Davtalab14} proposed a multi-level fuzzy min-max (MLF) neural network implementing a multi-level tree structure to construct a homogeneous cascading classifier. Hyperboxes with different sizes are yielded in various network levels to tackle the overlapping issue. Each node in the network is a subnet as well as an independent classifier that can classify patterns belonging to the defined area of sample space. The classifier in root level will classify most nonboundary areas of sample space, while each node in the $ i^{th} $ level of the network is in charge of classifying samples inhabiting in an overlapping region of $ (i-1)^{th} $ level. Overlap handling in the MLF is executed after establishment and adjustment of all hyperboxes, which leads to reduced space and time complexity. MLF is able to classify samples belonging to boundary regions with a high accuracy thanks to smaller hyperboxes in child nodes. Experimental results of \citet{Davtalab14} indicated that MLF is more proper than original FMNN, GFMN, DCFMN, IEFCN and FMCN, and the training process is faster than that of other types of FMM networks in most cases. 

Although FMCN and DCFMN do not use the contraction procedures, they supplement numerous new neurons to the fuzzy neural network, which turns the neural network out to be more complex than before. To cope with all issues, \citet{Ma12} introduced a modified data-core-based fuzzy min–max neural network (MDCFMN) with new learning mechanism. In that algorithm, the contraction process was removed without adding any new neuron to the network \citep{Ma12}. The learning algorithm for the MDCFMN only consists of one procedure: identify if hyperboxes need expand or not and compute the center of gravity of data in the same hyperbox. When the sample is presented to the model, if there is no existing hyperboxes being able to cover the sample, then we need to find expandable hyperboxes and expand them. The constraint to determine whether the hyperbox can be expanded or not is computed for each dimension in the same way as in the GFMN. If it is satisfied, the minimum and maximum vertices of hyperbox are tuned by using Eqs. \ref{eq7} and \ref{eq8}; otherwise, we need to create a new hyperbox to contain the new data sample. The gravity center of data in the same hyperbox is recomputed as Eq. \ref{eq79} after the new data have been added to the hyperboxes.

\begin{equation}
    \label{eq79}
    g'_i = \cfrac{g_i \cdot \eta + X_h}{\eta + 1}
\end{equation}
where $ g_i $ is the old gravity center; $ \eta $ is the number of samples in the hypebox, $ X_h $ is the input data.

It is seen that the modified DCFMN used a recursive procedure to determine the average value of data located in a hyperbox. The new learning algorithm does not need to perform the overlap test procedure with the aim of forming the special neurons for overlapping areas, the speed of pattern classification may be raised. Although the authors indicated that their proposal does not yield an special node for each overlapped area, they employed such nodes implicitly \citep{Forghani15}. While the contraction process is more likely to lead to classification errors, the formulation of a special node for each overlapped region in training stage may lead to the complexity in the architecture of neural networks and increase in time and space complexity. There is an interesting finding that the misclassification probability in the case of both training and test patterns being from identical probability distribution is minimized if the classifier has symmetric margin \citep{Vapnik00}. However, none of above methods is capable of classifying patterns positioned in overlapped regions with symmetric margin. Hence, \citet{Forghani15} proposed a fuzzy min–max neural network with symmetric margin (FMNWSM). In that approach, only hyperbox expansion procedure is carried out in training phase. It does not find overlapped regions, does not perform the contraction process to eliminate overlapped areas and does not yield any special hyperbox for overlapped regions. Due to using only expansion procedure, the training time of FMNWSM is lower than that of kinds of conventional FMNNs such as FMNN, GFMN, FMCN, and DCFMN. In real-world applications, data can be infected by noise, which results in formulating hyperboxes with wrong boundaries. To cope with the impact of noise on classification results, \citet{Forghani15} proposed an extended version of FMNWSM, called EFMNWSM, by constructing a new membership function as in Eq. \ref{eq81}.

\begin{equation}
    \label{eq81}
    b_j(X_h, V_j, W_j) = \min \limits_{i}[\epsilon s_{ji} - |x_{hi} - y_{ji}|]
\end{equation}
where $ y_{ji} = \frac{e_{ji} + g{ji}}{2} $, $ g_{ji} $ is the average value of the $ i^{th} $ dimension of training data contained in the $ j^{th} $ hyperbox, and $ \epsilon > 0 $ is an optional parameter. The use of parameter $ \epsilon $ and $ g_j $ in the membership function contributes to the change of size of the $ j^{th} $ hyperbox and the movement of center following its middle point and mean of data covered by it. Hence, these values may correct the boundary of the $ j^{th} $ hyperbox. The change of hyperbox membership function according to its data average is to established higher membership value to the data that are nearer to hyperbox data average \citep{Zhang11}.

In an attempt to reduce the ``black-box" property of WFMM, \citet{Kim13} introduced a rule extraction technique for sign language recognition. Authors used a combination of the weighted fuzzy min-max model \citep{Kim06} and new rule extraction technique for the pattern classification step. The training process of modified weighted fuzzy min-min (MWFMM) neural network comprises only two procedures: creation and expansion of hyperboxes. The connection weight $ w_{ji} $ is constantly updated during the learning phase. This value reflects relevance factors between the features and the hyperboxes. Hence, the network can deal with the hyperbox overlapping problem without performing the hyperbox contraction process \citep{Kim13}.

\section{Hyperbox based hybrid machine learning models} \label{sec5}
\citet{Mirzamomen16} proposed a fuzzy min-max decision tree (FMMDT), in which each internal node of the decision tree includes a contraction-less fuzzy min-max neural network (CLFMNN). The strong points of the decision tree and the fuzzy min-max neural networks are integrated into a hybrid model to mitigate the drawbacks of each algorithm. The shortcomings of the FMNN, i.e., the dependency on the maximum size of hyperboxes and the performance degradation  due to the contraction process, can be tackled based on the hierarchical structure of the decision tree. Unlike conventional decision trees where a single attribute is chosen as the split test, by using of the FMNN, each non-leaf node is split non-linearly based on multiple attributes, and thus it can capture the complicated structures in the data as well as forming decision trees with much smaller depth. The CLFMNN has several changes compared to the original FMNN as follows. Instead of using the contraction process, a new mechanism is designed to handle the input sample located in the overlapping regions. The CLFMNN uses a membership function as the data-core-base fuzzy min-max neural network given in Eq. \ref{eq49}. The CLFMNN embedded at the internal nodes of the decision tree has two primary goals. The first one is to employ multiple features for a split decision at a given node in order to lower the depth of the decision trees. The second purpose is to incorporate fuzzy logic in the decision tree aiming to provide soft decisions for splitting an internal node \citep{Mirzamomen16}. The FMMDT is grown according to the top-down method starting from the root node containing all training patterns. Samples in each node are then partitioned recursively until all or most of the samples are of the same class. Splitting a node in the FMMDT is performed by training the CLFMNN on the input sample of the node and using its output for splitting operation. The final model comprises several label-leaf nodes and only one functional leaf node, in which the authors proposed to employ a naive Bayes classifier to identify the class label for the pattern that reached it. During the learning process, the value of maximum allowable hyperbox size $ \theta $ in the FMMDT algorithm is assigned a large value in the root node, and it will be decreased in the subsequent levels adaptively. However, the FMMDT constructed according to this method is a batch decision tree learner for the static context \citep{Mirzamomen17}.

To learn from continuous attributes in data streams, which is more challenging compared to the static context, \citet{Mirzamomen17} proposed an evolving fuzzy min–max decision tree (EFMMDT) learning algorithm, where each decision node of the tree includes a concept adapting contraction less (CACL) fuzzy min–max neural network. The authors used trainable split checks relied on multiple attributes for each decision nodes by embedding the CACL fuzzy min–max neural networks with concept-drift processing mechanisms to it. Hence, the decision tree is able to be easily adapted to the new concept through employing a forgetting mechanism and training on new information \citep{Mirzamomen17}. The CACL fuzzy min–max neural network in each internal node is utilized to decide the branch that each sample should be put through based on its output. In addition, the CACL netwok provides the decision tree with the needed flexibility with the aim of adapting to concept drift. For completely adapting to all kinds of concept drift, the classifier needs a mechanism to re-size the previously learned hyperboxes or discard them if they are not suitable to the new situation. In CACL network, therefore, each hyperbox keeps a statistics record and this record would be updated whenever a training sample matches that hyperbox. After that, the hyperboxes would be modified or deleted based on its evolving statistics \citep{Mirzamomen17}. The statistics record constructed on the basis of the sliding window of recent samples consists of hit rate, which represents the number of times a hyperbox has included a new sample, and min-max points describing the minimum and maximum coordinates computed from the matched training samples. When the number of recent training samples exceeds a threshold value from the last update, the hyperboxes in the model are adapted according to their calculated statistics. It is seen that the statistics-based updating mechanism frequently shrinks the hyperboxes following the presentation of recent training patterns. Whenever a new training sample is used, the algorithm traverses the tree from the root to a leaf to separate the node using Hoeffding bound \citep{Hulten01} for electing the features. 

In another study related to combination structures of network and tree, \citet{Seera12} proposed a hybrid model comprising the fuzzy min–max neural network \citep{Simpson92} and the classification and regression tree (CART) \citep{Breiman84}, namely FMM–CART, for fault detection and diagnosis of induction motors. Although FMNN has online learning capabilities with one-pass training, it does not have the ability to explain its predictions and the capability of handling categorical data. Meanwhile, CART is able to explain its predictive results with rules as well as handling both numerical and categorical data, but it is less flexible with regard to learning from data patterns. The hybrid model can tackle the drawbacks of both models with the ability to learn from data samples and produce rules for explanation of its predictions simultaneously. The hyperboxes of FMNN form a crisp input data set to construct a tree based on the CART procedure. The operation of CART comprises three basic steps \citep{Seera12}. The first one is tree constructing. A tree is formulated by recursive separating of nodes and assigning each leaf node to a predicted class. The obtained tree contains decision nodes and leaf nodes. The second step is to prune the tree using the method of cost-complexity pruning \citep{Lewis00} with the aim of enhancing the classification performance. The third step is the optimal tree selection. The final optimized tree is deployed for performing classification tasks.

In the later study, \citet{Seera14} extended FMM-CART by proposing modifications of the CART algorithm and FMNN. In the learning algorithm of FMNN, a confidence factor, $ CF_j $, for each hyperbox is identified. The confidence factor contributes in determining hyperboxes used very often and being generally accurate in prediction, or hyperboxes seldom employed but highly accurate. The centroid of a hyperbox which is the center of the most prevalent data patterns included in the hyperbox is computed. If the centroid is outside the hyperbox after contraction process, the hyperbox will be reverted to the state before expansion and a new hyperbox is generated to cover the input pattern. To improve the accuracy of a classification tree, the authors provided each class of the decision tree with a confidence level, called class weight, using the values of the FMM hyperboxes' confidence factors \citep{Seera12}. In the CART model, Gini index is used to identify the splitting criteria. \citet{Seera14} introduced a new way of calculating this Gini index to improve the performance of the CART.

The fuzzy min–max clustering network network \citep{Simpson93} does not require providing the number of clusters in advance since it offers capability of online learning, and its number of clusters can be grown incrementally according to the data distribution. However, the FMM clustering neural network is unable to generate comprehensible rules to explain its clustering results. In contrast, the clustering tree (CT) is capable of explaining the underlying cluster structures. Therefore, \citet{Seera16} combined the modified fuzzy min-max neural network for clustering \citep{Seera15} with the clustering tree to create a new model, called FMM-CT model, for tackling data clustering problems with the strengths of online learning and rule extraction. In the traditional clustering tree, an error could appear if an input pattern is sent through an improper branch of the tree due to the noise in the input. To cope with this problem, the FMM-CT uses the modified FMM clustering neural network to cluster all input samples into $ c $ hyperboxes, in which each hyperbox centroid represents a cluster. All hyperbox centroids along with their confidence factors are deployed as input data samples to build the clustering tree. In the CT, the maximum deviance reduction method \citep{Krasteva15} is deployed to split internal nodes.

\citet{Seera16} also introduced a new measure for the impurity of node with deviance computed following the weight of the hyperbox and the ratio of the number of instances contained in that hyperbox. To further improve the clustering performance of FMM-CT, an ensemble model of clustering trees (ECTs) is introduced by \citet{Seera18}. The authors also proposed the use of additional three controid computational approaches, i.e., geometric, harmonic, and root mean square (RMS) measures, in addition to the arithmetic mean method as mentioned above. An ensemble model is able to deal with noise and outliers more effectively in comparison with individual clustering trees \citep{Fauber15}. It is capable of generating a good balance between good and weak models in the final ensemble model. The authors used the ECT model incorporating multiple CTs by implementing the bagging method with random feature selection, which may decrease the variance by sub-sampling the trees and minimize the error rate \citep{Seera18}. The majority voting mechanism is then employed to aggregate results of base clustering tree. To evaluate the validity and quality of constructed clusters, authors employed the Cophenetic Correlation Coefficient metric \citep{Mirzaei10} in combination with the centroid of each cluster.

Ensemble classifiers using bagging are able to significantly enhance the classification performance, but it makes the classifier system complex, in which computational time and a large memory are required for attaining the final classification outcomes. The classifier ensembles also face loss of transparency the decision making process since an ensemble of numerous individual classifiers would be much more difficult to comprehend by users. In order to reduce the final predictor complexity while preserving good classification performance, \citet{Gabrys02b} proposed the way of combining the hyperbox fuzzy sets of base models, i.e., GFMN models created during repeated 2-fold splitting of the training data, rather than their decisions. The combination of hyperboxes is straightforward and has already been implemented in the agglomerative learning algorithm \citep{Gabrys02a}. The hyperbox fuzzy sets from different component machine learning models are incoporated as inputs to the agglomerative training algorithm with the aim of removing the redundant hyperboxes and adding or refining hyperboxes covering the regions near the class borders or overlapping areas. Practically, hyperboxes from different individual models can be overlapping hyperboxes representing different classes. This fact may result in classification ambiguity when combining the models. Therefore, these undesired overlapping regions must be eliminated by using contraction process of GFMN learning algorithm for online learning. After resolving all overlapping hyperboxes, the agglomerative learning procedure is applied. To avoid overfitting for the final model, pruning operation has to be employed to delete from the model all the hyperboxes which misclassify more input instances in the validation set than they classify accurately \citep{Gabrys02b}. The experiment results of \citet{Gabrys02b} indicated that the combination at the model level gives much better performance than incorporation at the decision level or by utilizing individual component classifiers learned from only part of the training set. However, training time increases compared to combining at the decision level as an additional training cycle has to be conducted to combine the hyperboxes from individual models. One specially interesting feature of combining at the model level is that it is scalable for large data sets by employing mutually exclusive subsets of the original data set to train base models on parallel computers.

One of the main issues ensemble models encounter is the lack of a simple structure \citep{Nugent05}. This can make the classification systems become a `black box' model without capability of explaining their predicted results. It is desired to gain the advantage of incorporating classifiers whilst maintaining a classifier with a relatively simple architecture. Motivated by this goal, \citet{Eastwood11} proposed a way of building a model level combination scheme in which an ensemble of decision trees is employed to robustly label a set of hyperboxes covering the input instances. These hyperboxes are then used as inputs to form a single, relatively simple classifier within the general fuzzy min-max framework \citep{Gabrys00}. By constructing the model on robustly labelled hyperbox instances, instead of directly from the data, the classification accuracy of the final model can be significantly enhanced on most data sets. Although experimental results showed that performance of tree ensemble hyperboxes via GFMN generally does not reach that of a full ensemble technique, this approach is still a good choice in cases when simplicity of model is a desirable factor \citep{Eastwood11}. 

\section{Other hyperbox based machine learning models} \label{sec6}
\citet{Abe95} proposed an effective method to form a fuzzy classification system by extracting fuzzy rules directly from numerical input and output data through hyperbox representation. Fuzzy rules are determined by activation hyperboxes, which represent the existence region of data for a given class, and inhibition hyperboxes preventing the existence of data in corresponding activation hyperboxes. Two kinds of hyperboxes are constructed recursively. An activation hyperbox for each class is identified by the minimum and maximum values on each dimension of the corresponding set of data belonging to that class. If two activation hyperboxes representing two different classes, e.g. classes $ i $ and $ j $, are overlapped, the overlapping area forms an inhibition hyperbox. If data for classes $ i $ and $ j $ exist within the inhibition hyperbox, new activation hyperboxes for each respective class is generated. If two newly created activation hyperboxes are overlapped, an additional inhibition hyperbox is produced for the overlapping region. Corresponding fuzzy rules would be formed from recursively resolving overlaps between two classes. To classify a new sample $ X $, the authors proposed the way of building a membership function between a sample and a rule. If the sample $ X $ is contained by an existing activation hyperbox, the membership degree of $ X $ with the rule formed from that activation hyperbox is one. In contrast, the membership value reduces if $ X $ moves away from the activation hyperbox. The way of dealing with the overlap problem between hyperboxes proposed by \citeauthor{Abe95} has several strong points. First, the hyperboxes are not contracted as in the original fuzzy min-max neural network and its variants, thereby the potential errors caused by the contraction process as mentioned in previous section do not occur. Second, the hyperboxes are constructed around complex topological information of the data \citep{Reyes-Galaviz15}. However, the recursive method of forming activation and inhibition hyperboxes to resolve overlapping regions between two classes each time can lead to a complicated architecture for the classifier when tackling difficult classification tasks. Based on the hyperbox construction method of \citeauthor{Abe95}, \citet{Thawonmas97} proposed a feature selection approach using the analysis of class regions created by hyperboxes. The degree of overlap in the different class areas is defined as the exception ratio to make a measure for feature assessment. The feature with minimum exception ratio is remove permanently. The experimental results conducted on four benchmark data sets showed that their proposed technique could successfully eliminate irrelevant features.

To use the FMNN as an action selection network in the reinforcement learning problems, \citet{Likas96} introduced a stochastic fuzzy min-max neural network, where each hyperbox holds a stochastic automaton. In the original FMNN, each hyperbox is featured by its position and relevant class label, while, with regard to the stochastic fuzzy min-max network, the stochastic automaton whose probability vector defines the corresponding operation via random election process is adopted instead of the class label. The purpose of the learning process of the stochastic FMM network is to adjust the position and border of each hyperbox and the probability vector of each stochastic automaton \citep{Mohammed17a}. There are two kinds of hyperboxes being taken into account, i.e., deterministic hyperbox, which is accompanied by a specific action label, and random hyperbox, where the relevant action is chosen using uniform random selection. In that work, learning is an activity of supplementing random hyperboxes so that they are more likely to become deterministic ones. After a sufficient number of stages, it is desired that no random hyperboxes appear anymore. The use of random hyperboxes contributes to the capability of exploring the discrete output space to find out the best action. If such an action is sought, it would be allocated to the random hyperbox which becomes deterministic at that stage \citep{Likas01}. However, this method has several disadvantages. Firstly, all actions in a random hyperbox have equal probability of occurring, hence, we are unable to support a certain specific action. Secondly, there is an immediate transition from stochastic to deterministic when a rewarded action is chosen, and this may causes several problems as the rewarded action may not be the best one. Another drawback is that the modification of the action label of a given hyperbox is impossible \citep{Likas01}. Finally, the creation method of a new random hyperbox within the original hyperbox and the shrinking of both of them to prevent the appearance of overlapping regions result in the formulation of an immoderate quantity of small volume hyperboxes and cause the difficulty in the adaptation of learning algorithm. To cope with these issues, in the extended version, \citet{Likas01} proposed an improved stochastic fuzzy min-max network, where all hyperboxes are taken into account as random ones, and each of them holds a stochastic automaton. The key role of the automaton is to determine the degree of randomness in the process of choosing suitable actions. The online learning algorithm of the stochastic fuzzy min-max network consists of two main procedures. The first one is the tuning of the location and volume of each hyperbox by employing the expansion-contraction procedure of the original fuzzy min-max network. The second activity is the allocation of the action label to each suitable hyperbox following the parameter adjustment of the stochastic automaton.

\citet{Xu09} introduced a mathematical programming-based classifier modelling classification boundaries as hyperboxes. Training process in the proposed method tends to construct for each class a number of hyperboxes covering as many training instances as possible. The optimal position and dimension of each hyperbox is identified by a mixed integer linear programming model (MILP) aiming to minimize the total number of misclassifications. In this model, any two hyperboxes of two different classes are not allowed to overlap because hyperboxes cover the unique sample of corresponding classes. A binary variable, $ Y_{jki} $, is introduced to model non-overlapping conditions mathematically. To enhance the accuracy of the model, an iterative algorithm is performed after building a hyperbox for each class to contain the maximum number of correctly classified patterns. In this stage, multiple boxes are additionally constructed for each class to include any misclassified patterns during previous iterations as well as the number and position of these boxes are optimized during this phase. In the prediction phase, if a new data is positioned inside the area of the existing hyperbox, it is assigned to the class label of that hyperbox. Otherwise, it is assigned to the class label of its nearest hyperbox by using the Euclidean distance in the solution space. The training algorithm of this classifier continues until there is no way to gain a better solution by adding another new hyperbox. 

In each iteration of the mathematical programming-based classifier, a new multiclass prediction model needs to be solved regardless of the boundaries of hyperboxes achieved in the previous iteration. As a result, numerous training instances, which are correctly classified, are repeatedly considered to formulate boundaries of hyperboxes and generate constraints in every iteration. In addition, a weak point of MILP models is that the computational time is relatively long, which results in low efficiency when the method is applied for large size data sets. Hence, \citet{Maskooki13} proposed a modified version of the training algorithm to significantly reduce the training time. The idea of their algorithm is to take advantage of the boundaries gained from previous iterations and remove the correctly classified samples from the training set in each iteration rather than utilizing whole data set for finding new boundaries. Hence, several constraints corresponding to accurately classified patterns are not created and thus the number of binary variables needed to be considered is decreased through iterations. In this training algorithm of \citeauthor{Maskooki13}, only the misclassified instances from different classes are passed to the next iteration. By this method, several constraints are reduced and therefore the number of binary variables $ (E_h) $ is decreased through each iteration. Experimental results showed that the modified model requires only from 1/3 to 1/2 computational time in comparison with the original model of \citeauthor{Xu09}. Another type of improvement for the MILP model was proposed by \citet{Yang15}, where they aimed to refine the quality of the formed hyperboxes. They introduced two new proposals to enhance the performance of the hyperbox based classifier. They firstly extended \citeauthor{Xu09}'s work by incorporating a sample reweighting scheme, where higher weights are assigned to misclassified patterns included in other hyperboxes to adjust the model aiming to consider those difficult patterns in the next iteration. Moreover, to reduce high computational expense of the original model, they proposed a data space splitting technique to partition the training instances into two disjoint areas. The model is then resolved in this new space. The outcomes of experiments \citep{Yang15} indicated that the sample re-weighting technique gives consistently higher prediction accuracy compared with the traditional model. Furthermore, the sample partitioning technique lowered the computational cost by one or two orders of magnitudes but still keep the desirable prediction rates 

Most classical statistical models make assumptions on underlying data, their distributions, independence, stationarity, etc. However, such assumptions are usually unrealistic in the real-life applications \citep{Grzegorzewski13}. Furthermore, the traditional statistical techniques cannot generate ambiguous outputs if its inputs exhibit uncertainty, vagueness, imprecision, or fuzziness. Motivated by the practical demands for simplification, low cost, approximation, and the tolerance of uncertainty in information processing, the granular computing \citep{Bargiela03} was proposed with the aim of the construction of intelligent systems with more human-centric view. \citet{Peters11} introduced a simple and intuitive method, called granular box regression analysis, to establish a fuzzy granulation generalization of a function between several independent variables and one dependent variable in the context of granular computing by using hyperboxes. Formally, the principal idea underlying the granular box regression is to approximate a set of $ N $ samples $ s_l = (y, x_1,..., x_n) $, where $ l = 1,..., N $, by a predefined value of $ K $ hyperboxes. To minimize sum of the volumes of the hyperboxes, \citeauthor{Peters11} proposed to swap the border samples for their closest right and left neighbors in the non-overlapping y-dimension. The granular box regression faces several limitations such as  the need of monotonicity, its sensitivity to outliers, and the necessity of semantic interpretation of the results. \citet{Reyes-Galaviz15} introduced a new technique to form a granular model for clustering in which information granules are represented as hyperboxes.

Most studies on the construction of hyperboxes are based on the original fuzzy min-max neural network proposed by \citet{Simpson92}. Hence, hyperboxes are limited by an allowable maximum size and non-overlapping hyperboxes conditions. While the maximum size can lead to a complex model with a large number of small hyperboxes, the contraction process to remove overlaps causes loss of included data. Therefore, \citet{Reyes-Galaviz15} proposed a novel technique for the construction of hyperboxes, where they deployed conditional Fuzzy C-Means (FCM) \citep{Pedrycz98} to establish direct associations among information granules, and then utilize them to construct hyperboxes. The data sample is collected in the form of input and output pairs. First of all, the output space is divided into intervals to represent information granules. After that, the input data are clustered by employing the conditional FCM, the cluster centers (called prototypes) resulting from this stage are employed as hyperbox cores to build hyperboxes. The authors introduced two different approaches to form hyperboxes from these cores. In the first one, a numerical constraint based on the average distance between the cluster centers is employed to build the hyperboxes \citep{Reyes-Galaviz15}. The second one optimizes the expansion and overlap reduction of hyperboxes by using a differential evolution (DE) \citep{Storn97} associated with considering the maximum number of data covered. If a data point is located in two hyperboxes belonging to two different intervals, it is not obvious to which context this point belongs. Therefore, overlapping regions between hyperboxes of different intervals need to be eliminated. This operation is the same as the contraction phase of the original fuzzy min-max neural network. In the case that an input datum is included in two or more hyperboxes of only one interval, we need to compute the distance between that datum and the cluster center (hyperbox core) or the membership function to allocate this input datum to only one hyperbox. Experimental results shown in \citep{Reyes-Galaviz15} indicated that the clustering performance of this approach increases when data have the large number of features and instances compared to the original fuzzy min-max clustering neural network. 

\citet{Palmer-Brown11} proposed a hyperbox neural network (HNN) algorithm for classification, in which each class is associated with only one hyperbox and a single neuron. Each hyperbox covers and presents the ranges of attribute values of samples belonging to the same class. If a new sample falls in only one hyperbox, it will be classified immediately to the class label of that hyperbox. In contrast, if the pattern is located in the overlapping region among hyperboxes, the neurons of these hyperboxes are used to classify the sample. \citet{Eastwood14} extended the hyperbox neural network by proposing a piece-wise sigmoid adaptive activation function to substitute the piece-wise linear function in \citep{Palmer-Brown11}. They also introduced a combination of the supervised hyperbox and neurons with unsupervised clustering.  In the prediction process, if a new sample is covered by only one hyperbox, it would be assigned the class label of that hyperbox. If a new sample is contained in more than one hyperbox, it would get the class label of the hyperbox with the highest activation. If a new sample does not belong to any hyperbox, then the label of the hyperbox with highest number of dimensions covering the range of the sample is assigned to the sample. When there are many hyperboxes which meet this condition, the activation value is taken into account. In case that all of the sample's dimensions are out of the range of the dimensions of all hyperboxes, the sample cannot be classified \citep{Palmer-Brown11}. The use of only one neuron per each class has one potential weakness that it might only express class distribution within a hyperbox varying in only one single direction \citep{Eastwood14}. Therefore, \citet{Eastwood14} proposed a change to the above method, in which they performed clustering on the data belonging to each hyperbox to form data clusters within each hyperbox. A single sigmoid neuron is then associated with each cluster and is trained on only the data in that cluster. Building more than one neuron for each hyperbox based on clusters allows to handle data sets where the different classes have many overlapping areas, and can model the class distribution within a hyperbox in more than one direction.

\citet{Park14} suggested the development of a hyperbox classifier with hierarchical two-level structure on domino extension from seed hyperboxes. This classifier comprises a core structure built on a basis of a set of hyperboxes and secondary structure constructed on a basis of fuzzy sets. The core structure is formed by the domino extension of seed hyperboxes. As for the domino method, the feature space is discretized, and then a hyperbox chosen as seed in the new space grows up to formulating a cluster. The secondary structure is built by implementing a Hausdorff distance \citep{Olson98} between a group of hyperboxes and a sample.

The topological information is usually ignored in conventional clustering methods. This shortcoming is effectively tackled by using clustering techniques based on hyperboxes to capture  the distribution of data on the feature space. \citet{Ramos09} proposed a hyperbox clustering with ant colony algorithm (HACO) for clustering the unlabelled data by near-optimally placing hyperboxes in the feature space. Hyperboxes, which are frequently smaller than the number of patterns, are sought for by using the ACO algorithm \citep{Dorigo04} and then clustered utilizing the nearest-neighbor approach. The HACO technique comprises two stages. The ACO is first deployed to optimally scatter the hyperboxes on the feature space \citep{Ramos09}. The second stage is to group the hyperboxes to formulate clusters. If the number of clusters is given beforehand, the hyperboxes are grouped using a nearest-neighbor approach; otherwise the output of ACO are the final clustering result. After ACO finished, the second step is to classify the hyperboxes. Overlapping hyperboxes are combined into one cluster, while non-overlapping ones are assigned to their own clusters. This operation yields the number of clusters automatically. If the number of clusters $ K $ is given beforehand and the number of clusters found is greater than the value of $ K $, the clusters closest to each other are categorized together employing the nearest-neighbor technique; otherwise, HACO generates novel hyperboxes to contain the samples with the greatest distances from the existing hyperboxes. The obtained result from HACO is a vector of hyperboxes for clustering data which can be used as machine learning models. If a new sample is included in the regions of an existing hyperbox, it would be assigned to the cluster of that hyperbox, else it is associated with the hyperbox nearest to it. To refine this model, \citet{Ramos08} proposed a hyperbox-based machine learning algorithm with ACO - type 2 (HACO2) by deploying the ant colony algorithm to evolve the geometry of hyperboxes in the feature space to better cover the data in the class. The output of HACO can be used as input of HACO2 to reshape the hyperboxes. Therefore, HACO can be considered as an appropriate initialization step instead of random initialization of hyperboxes so that HACO2 refines the structure. Each solution of HACO2 is a vector of edges of hyperboxes able to be expanded or reduced to better fit the data. 

\section{Real-world applications of hyperbox-based learning algorithms} \label{sec7}
\subsection{General purpose applications} \label{generalpurpose}
\subsubsection{Image processing} \label{imageprocessing}
Image processing is a fundamental component for the construction of computer vision systems. Researchers have proposed a large number of methods to tackle different aspects of these problems such as color image or video sequence segmentation, shadow detection and removal from color images, etc. These issues can be effectively solved by employing various versions of fuzzy min-max neural networks.

Image segmentation aims to partition an image into homogeneous areas to determine differences between specially interesting and uninteresting objects or discriminate between foreground and background in the image. For valuable information acquisition from color images, color image segmentation issue has to be first handled. \citet{Nandedkar09} introduced a way of dealing with this problem by using the granular reflex fuzzy min-max neural network to deal with a group of color pixels instead of handling individual color pixels. This solution significantly decreases computational cost compared to individual pixels processing. Min–max values of the pixels in each grid are then computed to build up a granule. After that, such granules are deployed to train the GRFMN. To segment a new color image, it would be separated into grids of small size to construct granules, and these granules are then put through the GRFMN trained by above method for object segmentation. \citet{Deshmukh06} used Simpson's fuzzy min-max clustering neural network to automatically seek clusters and their labels for the multilevel segmentation system for color images. The obtained information of labels is deployed to form an adaptive multithresholding mechanism used by a multilayer neural network. By the use of the FMM clustering neural network, the system does not require prior information of the number of objects in the image. 

Image segmentation method can be extended for video sequences as proposed by \citet{Nandedkar09}. They used labeled seed images from first frame to train GrRFMN. The acquired information is then deployed to classify subsequent color images. From the experimental results, the authors claimed that the GRFMN is suitable for the situation where online learning is expected.

One of the reasons reducing the dependability of the computer vision system is the existence of shadows. Furthermore, shadows result in degrading the quality of images. Hence, in order to form effective algorithms for image improvement and computer vision system, first crucial step must perform shadow detection and removal. \citet{Nandedkar13} created an interactive tool to detect and eliminate shadows from the color images by employing the learning ability of GrRFMN from data granules extracted from the images. The proposed system consists of the following units. The system first needs the interaction of users in form of strokes to obtain information of shadow and non-shadow areas as well as the region of interest from the color images in RGB color space. From image shadow and non-shadow patterns provided by users, the system conducts feature extraction by splitting these patterns into grids of small size to form granules in form of hyperboxes. Hyperboxes contain the minimum, maximum, and mean values of pixels in each grid. These hyperboxes are then utilized to train GrRFMN with aim of identifying shadow and non-shadow regions from the color image. After that, each pixel in the region of interest is calculated the membership value to shadow and non-shadow classes by building its neighborhood granulates and feeding them to the trained GrRFMN. Pixels in non-shadow regions are used to rectify pixels in the shadow areas. Finally, output image is presented to users for further interaction until the shadow regions are completely deleted. Although the authors described the suitability of using the hyperboxes in the tool to detect shadows, they did not compare their method to other shadow detectors.

\subsubsection{Image recognition} \label{imagerecog}
After the image processing process, recognition of objects from images or videos is the next essential stage to construct the computer vision systems toward a specific purpose such as surveillance systems, image searching tools, or automatic biometric systems.

Object recognition comprises two basic steps, i.e., feature extraction and sample classification. A raised demand for object recognition systems is that they should be invariant to rotation, translation, and scale operations. Therefore, features extracted from the images should also be invariant to translation, rotation, and scale. \citet{Nandedkar06b} introduced an object recognition system from a set of rotation, translation, and scale invariant (RTSI) features. Recognition phase is conducted by the reflex fuzzy min-max neural network with floating neurons using semi-supervised approaches. The authors proposed a novel set of RTSI features including normalized inertia moment, maximum distance to average pixel distance ratio, ratio of average pixel distance to difference between min-max pixel distance from centroid, radial coding and angles. Empirical outcomes indicated that Nandedkar's object recognition system using reflex fuzzy min-max classifier offers the efficient learning ability from data sets formed from unlabeled data in combination with very few labeled patterns.

In many practical applications such as robot navigation systems with demands for classifying obstacles in the path, it is better to adopt data intervals as inputs for object recognition due to the fact that the sensors are not absolutely proper. Hence, \citet{Nandedkar09} applied the granular reflex fuzzy min-max neural network to object recognition systems in images. A minimum and maximum values of RTSI features extracted from images are calculated to construct a granule of n-dimensions per image. Such granules are then used to train the granular reflex fuzzy min-max neural network.

Object recognition is one of the initial steps to construct a content-based image retrieval system. \citet{Kshirsagar16} took advantage of capability of learning in a single pass and no retraining  of general reflex fuzzy min-max neural network to construct an image retrieval system relied on content. GRFMN is employed to assess the similarity of texture and colour properties between each image in the database and a query image.

In practical computer vision systems, face detection is a first step in building most of the face recognition and tracking lines. \citet{Kim06} described a real-time face detection technique including two phases, i.e., feature extraction and classification. The feature extraction stage implements a convolutional neural network (CNN) in which a Gabor transform layer is supplemented at the first layer of CNN to seek for successively larger features. Extracted information is adopted as inputs for the weighted fuzzy min-max neural network model for the pattern classification stage. In addition, the weighted fuzzy min-max neural network was deployed to choose the sets of useful color features for the skin-color filter with the aim of determining candidate regions for the feature selector. Experimental results indicated that their system could adaptively select types of skin-color features for a given illumination condition. The accuracy of the face detection module is more than 90\% with 200 training samples used. Nonetheless, the authors did not compare their proposed method to other approaches concerning feature extraction, skin-color filter or pattern classification.

Character recognition is also one of the popular applications for biometric and many other fields like manufacturing, shipping, and banking. \citet{Chiu97} adopted the original fuzzy min-max neural network to recognize handwritten Chinese characters with arbitrary location, scale, and direction. The FMNN was used to classify ring-data features extracted from characters. \citet{Nandedkar04b} formulated an invariant optical character recognition system to scale, translation, and rotation using the original fuzzy min-max neural network. Compared to character, signature recognition is a more difficult task. \citet{Chaudhari09} introduced a signature classification framework by implementing the fuzzy min-max neural network on extracted features of signatures. \citet{Chaudhari10} then improved their work for online signature classification by applying the FMCN to the global shape features of signature sample.

One of the methods for normal people and deaf-mute people to communicate is the use of costly sign language interpreter. To build a cheaper and more easily accessible solution, researchers have developed a sign language recognition system. This system is able to recognize the sign language, and translate it to the normal language using text or speech. \citet{Kim13b} used the weighted fuzzy min-max neural network \citep{Kim04} to extract fuzzy rules for the dynamic hand gesture recognition problem, which is part of the sign language recognition system. In the later study, \citet{Kim13} applied a similar method for constructing the sign language recognition system including three components: preprocessing, feature extraction, and sample classification. In the preprocessing stage, the feature areas are detected by a skin color analysis process. The authors employed the motion history information, the motion energy data, and the hand-shape features to build input data for the extended convolutional neural network for the feature map generation. From these extracted features, the weighted fuzzy min-max neural network model is implemented for sample classification as well as forming feature relevance factors for rule extraction. The authors proposed a formula to compute the relevance factor between a specific feature and pattern classes based on constructed hyperboxes. If the relevance factor value between the feature $ f_i $ and the class $ c_k $ is positive, the feature $ f_i $ is an excitatory signal for the class $ c_k $. In contrast, if the relevance factor is a negative value, it shows an inhibitory relationship between a feature and a given pattern class. By analyzing a set of relevance factors, the authors generated a set of rules for each sign language pattern. However, the authors did not mention the accuracy of their sign language recognition system or its performance in comparison to other techniques.

\subsubsection{Speech recognition} \label{speechrecog}
In addition to characters, speech is one of common means of communication in daily life. Hence, speech signal recognition contributes significantly to human-machine interface applications. \citet{Doye02} introduced a modular general fuzzy min-max neural network to recognition of spoken Marathi digits. Each Marathi digit was recorded from 20 speakers with ten replications per speaker using the sampling frequency of 10000 Hz. Each record was processed with a low-pass filter having a cut-off frequency of 5000 Hz. Their proposed method achieved 90.2\% recognition accuracy, on average, in speaker dependent mode and 67.8\% for the speaker independent mode. In other work, \citet{Jawarkar11} utilized the fuzzy min-max neural network to identify speakers in speech recordings in Marathi language. The empirical results showed that FMNN gave the better performance compared to the Gaussian mixture model. 

\subsection{Medicine} \label{medicine}
Medical diagnosis is one of the important steps to build treatment protocols for patients. In this process, physicians attempt to determine the disease based on the physical indications, clinical diagnosis results, and other related symptoms. Medical diagnosis tools are developed to assist physicians in making a decision quickly. \citet{Quteishat08a} suggested the use of fuzzy min-max neural network and its two variants to diagnose acute coronary syndromes (ACS). A timely and expeditious diagnosis of suspected patients has an important role to play because it could significantly reduce the mortality risk when proper medical treatment is timely provided to patients. To evaluate the performance of the three kinds of fuzzy min-max networks in medical diagnosis, \citet{Quteishat08a} collected a set of 118 practical medical records of suspected acute coronary syndrome. The empirical results have illustrated that the modified versions of the fuzzy min-max network have fortified the effectiveness and efficiency compared to the original version in case of high size of hyperboxes.

In addition to the acute coronary syndrome, stroke is one of the most popular neurological issues and a leading cause of mortality all over the world \citep{Quteishat10}. This disease causes an interruption of the blood provision to any part of the body, leading to damaged tissues. Hence, early stroke diagnosis for patients plays a crucial role to reduce mortality risks. \citet{Quteishat10} conducted a binary classification for acute stroke patient diagnosis using the modified fuzzy min–max neural network along with a rule extractor based on genetic algorithm. They aimed to predict whether the Rankine scale of patients based on discharge belongs to class 1 (no symptoms or no significant disability) or class 2 (disability levels from slight to severe). The authors conducted experiments on a data sets containing 661 patient records, in which 141 records are in class 1 and 520 records are in class 2 \citep{Quteishat10}. Each patient record included 18 typical features such as age, medical history information, physical examination, test results, \textit{etc}. The experimental results indicated that the combination model of FMNN and genetic algorithms reached classification accuracy of 92.1\%. It outperformed the original FMNN, pruned FMNN, and logistic regression with classification accuracy for all of these methods below 81\%. However, the proposed method was not compared to any other popular machine learning algorithms. \citet{Kim04} proposed an approach to select relevant features from the heart disease data set. They used the weighted fuzzy min-max neural network for considering the weight of each feature to choose interesting features for each sample class. By taking the feature weight into account, the model becomes less sensitive to noisy instances in a data set. In addition, they introduced a relevance factor (RF) from the trained hyperbox network to identify the relevance degree for each feature to a class. If the value of RF is positive, there is an interesting relationship between that feature and the given class. In this way, a list of relevant features for each given class may be extracted.

Another medical diagnosis related to cardiovascular diseases is presented by \citet{Bortolan07}, where the authors designed a two-phase hyperbox classifier to analyze and classify electrocardiography (ECG) data. The ECG signal analysis is one of the most prevalent non-invasive methods to diagnose cardiac diseases and detect arrhythmias. The authors first built a feature space comprising 26 features from ECG and premature ventricular contraction signals. They argued that these features are relatively typical, appealing and widely used in ECG signal analysis community for sample classification. After that, they designed the two-phase hyperbox classifier to classify ECG signal samples to be normal or abnormal. In the first phase, the Fuzzy C-Means \citep{Pedrycz98} technique is implemented to cluster training data. From these data clustering results, a set of seed hyperboxes is constructed and is grown in the second phase by applying the genetic algorithm. Their empirical studies uncovered that a small number of hyperboxes are adequate to capture the topology information of the data. Another study which used semi-supervised clustering method to address the liver disease diagnosis problem was presented by \citet{Tran18}. The authors combined the FMM clustering neural network and semi-supervised learning approach to build a predictive model for diagnosing diseases concerning liver enzyme disorder. All input data in this model were unlabeled patterns, so the expense of labelling data was reduced. Experiments conducted on 4156 instances of patients from a hospital in Vietnam indicated that their method outperformed other compared approaches.

Detection and diagnosis of cancer are also one of the meaningful medical applications. \citet{Zhai14} constructed a computer-aided detection system for lung nodules in X-ray images using the fuzzy min-max neural network with compensatory neurons and K-means clustering algorithm. Their computer-aided detection system achieved the accuracy of 84\% on the sensitivity metric for the nodules with 2.6 fault-positives/CT using 19 computed tomography (CT) scans. The authors concluded that their system is at least on the same performance level as four other comparative methods. In another study, \citet{Deshmukh16} used the pruned fuzzy min-max neural network \citep{Quteishat08b} to detect and diagnose lung cancer. The experiments conducted on the benchmark and real-world datasets showed that the pruned FMNN achieved essentially the same predictive accuracy with the lower number of hyperboxes in comparison to the original FMNN. With the small values of the hyperbox size threshold, the reported accuracy of the system was 100\%. Gene data is a precious source of information able to be applicable for disease diagnosis. Accurate classification of tumor types from gene expression data of DNA microarray experiments is extremely challenging. \citet{Juan07} introduced an enhanced fuzzy min-max neural network, which overcomes drawbacks of choosing a suitable allowable maximum size of hyperboxes, to classify the gene expression data.

\subsection{Health-care} \label{healthcare}
In health-care related studies, probable medical risk profiles recognition is extremely crucial for early diagnosis because it helps to prevent the disease from evolving into more dangerous variants, e.g. cancer. However, this is a laborious process for physicians. Hence, it is desired to develop robust techniques to automatically extract high risk medical profiles. \citet{Ramos09} introduced the hyperbox clustering with ant colony optimization (HACO) approach to assist the process of identifying probable infection profiles from a Human Papillomavirus (HPV) data set to construct a basis for medical examinations. The authors carried out their experiments on a data set including 185 patterns with 21 features of oral mucosa and a questionnaire for 42 physical characteristics and habits among the local inhabitants \citep{Ramos09}. The questionnaire supplies the features, while the mucosa points out the health status (infected or not by HPV) of each individual. Experimental outcomes showed that accuracy of HACO are considerably better than fuzzy C-means and ACO (over 30\% more accurate and 7 times faster) \citep{Ramos09}. The result provided by HACO is the probable risk profile for the particular data region. This information assists the doctors to recognize patients that need further check-ups if their profiles fit the identified risk profile without the burden of a full examination for all patients.

Falling behaviors of the elderly people usually lead to hazardous complications such as stroke and heart attack. Therefore, fall detection is one of the essential applications in public healthcare. \citet{Jahanjoo17} used the multi-level fuzzy min-max neural network to detect fall accurately from acceleration sensor data. The obtained results overcame those using traditional machine learning techniques including multilayer perceptron, k-nearest neighbor, and support vector machines. 

\subsection{Pharmaceutical} \label{drug}
In daily life, drugs are used to prevent and treat many diseases regularly threatening our life quality. However, the ability of designing and discovering new and effective drugs has not met the demands over the years because of the lack of systematic methods able to handle various aspects of the drug discovery process such as specificity, toxic effects, and side effects. Virtual screening of chemical repositories has recently become one of the new methods for the structure-based drug discovery. Nonetheless, screening of millions of compounds in chemical libraries is computationally expensive and time consuming. \citet{Tardu16} used a combination of mixed-integer linear programming based hyperbox (MILP-HB) classification technique \citep{Uney06} and partial least squares regression (PLSR) to filter the compounds in the initial libraries unsuitable for screening process against a specific target protein. These compounds usually have high binding free energy (BFE), so the purpose of classification procedure is to classify compounds into low and high BFE sets. The authors illustrated the efficiency of the proposal on a target protein, SIRT6. The empirical results showed that the proposed method outperformed other techniques and obtained 83.55\% accuracy using six common molecular descriptors.

\subsection{Manufacturing} \label{manufacturing}
Fault detection and diagnosis tasks usually occur in manufacturing sectors. \citet{Meneganti98} applied the fuzzy min-max neural network with a new batch learning algorithm for finding anomalies in the cooling system of a blast furnace. The input data were taken automatically from sensors in the system. They contain 21 features corresponding to sensor measurements and three failure classes, i.e., sensor failure, water loss, and pipe stoppage. The experiments showed the same performance of their method in comparison to other fuzzy min-max neural networks at 95\% confidence interval. In addition, their proposed method performed better than an adaptive optimal fuzzy logic system but was worse than the multi-layer perceptron (MLP) with Quasi-Newton algorithm. In the same study, authors applied the MMM-BL to failure diagnosis in a production line. The images of products captured by the optical system are processed, and 27 features of defects are extracted. From these features, the system has to be capable of classifying the failures to one of seven groups. Their method achieved the error rate of 26.6\% on testing sets and 0\% on training sets. Its classification accuracy was better than Simpson's model, the ART neural networks, fuzzy basis function based neural networks, MLP with Quasi-Newton algorithm, and the adaptive optimal fuzzy logic system.

\citet{Chen04} used fuzzy min-max neural network integrated with symbolic rule extraction to identify and diagnose the heat transfer equipment based on real sensor data taken from the circulating water system of the plant. The experiments on a case study with a set of real sensor data collected from a power generation plant in Malaysia showed that the best predictive accuracy was 97.66\%. However, their study only evaluates the influence of parameters on the performance of the model without comparing to other approaches. In the later work, \citet{Quteishat07, Quteishat08b} deployed the combination of modified fuzzy min–max neural network and rule extraction using genetic algorithms to detect faults happening in the cooling devices of a circulating water system in the factory. Their method achieved an accuracy of 94.82\%, with only three hyperboxes in predicting whether heat transfer conditions in the condenser are efficient or not. Their model also achived an accuracy rate of 97.15\% with only six hyperboxes in predicting the existence of blockage in the condenser tubes.

\subsection{Cybersecurity} \label{cybersecurity}
The communication via the internet is in danger of exposing sensitive information due to network attacks. Attacks have become more and more hazardous as attackers have employed numerous advanced techniques to conceal their intentions, which are ultimate targets of attacks. Identifying malicious intentions or intrusion detection can assist administrators to build better security systems. \citet{Ahmed18} suggested a similarity method for the recognition of attack intention in real time by adopting the fuzzy min-max neural network. Their proposal aims to classify attacks in accordance to the properties and employs a similarity measure to determine motivations of attacks and anticipate intentions of these attacks. Their system uses two phases to investigate the similarities in the attack features. In the first one for the intention recognition system, the attack patterns are produced as a baseline by extracting the principal characteristics of the known attacks. In the second stage, the attack intention is detected by calculating the similarity between the signal of the generated samples and the evidence selected from a particular attack \citep{Ahmed18}. The fuzzy min-max neural network was implemented to identify the similarity degrees of evidence in every observed attack in order to match it to the most likely intention of current attack. It predicts the attack intention by building a set of hyperboxes representing the possible attack evidence. Then, the FMNN employs the membership function to compute the degree of fit between the current input attack intention and generated rules. Experimental results in \citep{Ahmed18} demonstrated that their proposed system created useful information and improved the possibility of recognizing attack intentions.

Regarding intrusion detection, \citet{Azad16} constructed a new intrusion detection system by integrating the genetic algorithm to the fuzzy min-max neural network (FMNN-GA). The system was trained using FMNN and then hyperboxes were optimized by the genetic algorithm. The obtained results showed that their system outperformed other intrusion detection frameworks. In the later work, \citet{Azad17} combined the FMNN and particle swarm optimization to form the intrusion detection system. The experimental results indicated that their improved method is superior to other methods such as FMNN-GA, support vector machines, original FMNN, Naive Bayes, and multilayer perceptron.

\subsection{Transportation} \label{transportation}
Rail vehicles play crucial role in the public transport systems. To ensure the safety of train, faults in the systems need to be isolated and repaired. Suspension system is a critical element of the rail system, so real-time operation statuses monitoring and fault prediction on it are extremely important. \citet{Lv15} compared fault diagnosis and isolation approaches using support vector machines and fuzzy min-max neural network for the train suspension system. The empirical results indicated that both methods could attain quite good accuracy, and support vector machine gave slightly better results than FMNN.

Vehicle navigation, which is one of emerging fields of research, is attracting much attention of research community. This field is one of fundamental steps to develop unmanned vehicles. The problem of training for a vehicle navigation requires to identify and avoid obstacles in the path. \citet{Likas96} used the stochastic fuzzy min-max neural network to solve the collision-free autonomous vehicle navigation problem. They trained the fuzzy min-max neural network to select the suitable driving commands based on the current status of the vehicle provided by eight sensors. There are five driving commands corresponding to five classes of the stochastic FMNN. The input data include eight features which are the distances from the vehicle to the obstacles computed by eight sensors. Empirical results in \citet{Likas96} indicated that the performance of the stochastic fuzzy min-max neural network outperformed the typical feed-forward neural network using Bernoulli output units. In another study, \citet{Duan07} integrated the fuzzy min-max neural network and reinforcement learning to construct the detection and avoidance mechanism of obstacles for the reactive robot. The use of hyperboxes has contributed to the enhancement of the reinforcement learning system.

\subsection{Electrical and electronic engineering} \label{engineering}
In technical equipment, motors are usually utilized to transform electrical power to mechanical power to operate devices. During operation of motors, a great deal of problems can occur such as over-heating, harmonic problems, and the decrease in operational life \citep{Seera12}. Hence, it is desired to construct an effective, low-cost fault detection and diagnosis method to reduce the maintenance and downtime expenses as well as to avoid unexpected failures of motors. \citet{Seera12} designed an effective technique to identify and classify comprehensive fault conditions for induction motors. First, they applied motor current signature analysis to extract data features from the power spectral density of real stator current signals \citep{Seera12}. Then, different kinds of fault conditions were classified using the hybrid model combining fuzzy min–max neural network and the classification and regression tree, namely FMM–CART. Experiments showed a promising outcome with 98.25\% accuracy in noise-free conditions. However, this method only works in offline mode. To deal with online motor detection and diagnosis issues, \citet{Seera14} introduced an improved version of FMM-CART. The empirical outcomes pointed out the potential of the improved FMM-CART model in tackling online fault detection and diagnosis tasks for motors.

\subsection{Utilities} \label{powersystem}
Power quality is a crucial criterion of an electrical network because poor power quality may result in financial loss and catastrophic consequences, particularly in the industrial sector \citep{Schipman10}. It is desired to have a well-founded power quality monitoring system to identify the reasons of disturbances in electrical systems and enhance the quality of electrical power. \citet{Seera15} employed a modified fuzzy min–max clustering neural network to deal with power quality monitoring tasks. The authors claimed that the use of data clustering approaches for power quality monitoring is quite rare. Meanwhile, the analysis of fuzzy cluster showed its potential for assessment of power quality \citep{Duan06}. Hence, they conducted experiments to verify the effectiveness of the modified fuzzy min–max clustering neural network in realizing power quality monitoring tasks with real data sets from a hospital in the state of Pahang, Malaysia. The goal of the authors was to find the root cause of power supply disruptions and to identify if the supplied energy to medical equipment can maintain a satisfactory operation. The authors collected 1601 patterns with twelve features to form the input data to the modified fuzzy min–max clustering neural network. The empirical results proved the applicability of the modified fuzzy min–max clustering neural network to the power quality management domain.

\citet{Seera16} applied a hybrid intelligent method between the modified fuzzy min–max clustering neural network \citep{Seera15} and an improved clustering tree (FMM-CT) for monitoring power quality data aiming to produce explanation when encountering anomalies in the data distribution. When data anomalies happen, a remarkable growth in the number of clusters and leafs can be observed as more hyperboxes are generated in the network to capture the underlying data distribution. This enables FMM-CT to determine anomalies in power quality data instances.

Other studies have also applied variants of the fuzzy min-max neural network for monitoring or detecting faults in the energy distribution systems. \citet{Zhang11} used the DCFMN to manage the operating statuses of the liquid pipeline aiming to identify and alert the abnormal status. \citet{Gabrys99, Gabrys00} illustrated the use of general fuzzy min-max neural network to detect and identify leakage in a water distribution system. The water distribution networks provide the input samples in either fuzzy or crisp form, and their states includes both labeled and unlabeled patterns. In addition, the size, consumption-inflow instances, and anomalies of these networks are unforeseen. Therefore, they require a predictive model capable of evolving to satisfy the demands without retraining as well as performing supervised learning and unsupervised learning simultaneously on fuzzy or crisp input data. As a result, authors constructed a two-level recognition system using GFMM neural networks to handle the issues of leakage detection and identification based on the state estimates and the residuals in the water distribution network. Experimental analyses demonstrated and highlighted an analogy between the high-level information processing capability of the GFMM neural network and human operators when analyzing the operational states of the water distribution network. Besides, the system can be adaptable to include new information without retraining the constructed model. It is noted that only machine learning algorithms with the ability to handling heterogeneous data are suitable for this problem because the water distribution network states contain both labeled and unlabeled data types. \citet{Liu17} proposed the use of the new modified fuzzy min-max clustering neural network for analyzing the oil pipeline internal inspection data. They aimed to extract and analyze the leakage data to detect abnormal phenomena in the pipeline. The data set was extracted from their experimental system on oil pipeline consisting of the control station, air compressor and storage tank. The empirical outcomes indicated that their modified fuzzy min-max clustering neural network obtained better performance than other existing clustering methods such as the original fuzzy min-max clustering neural network.

\subsection{Politics and sociology} \label{politics}
In the field of politics and sociology, the opinion polls are performed to explore the individual attitudes or opinions of relevant people. In many cases, there are several non-response answers in the survey. The easiest way to handle non-response in survey reports is to add a ``don't know" category, but this method usually causes issues at the analysis phase \citep{Little02}. A general method to deal with partial non-response is to replace the missing values with the answers of other respondents and the non-missing responses in other answers of the same individual. Therefore, \citet{Castillo12} proposed applying the fuzzy min–max neural networks for categorical data to imputation of missing voting intention in political polls based on the answers to other questions in the same questionnaire. They selected surveys of general elections which took place in Spain in 2004 and 2008 including 16,345 and 13,280 interviews respectively. Surveys contained different variables, i.e., quantitative variables, ordered categorical variables, and categorical variables with non-ordered choices. They used the categorical voting intention variable as the class variable with eleven different values. In addition, the input dataset contained sixteen features which have numerical and categorical variables. Each missing value in the class variable was replaced by the class value with the greatest membership degree computed from its feature values. Ten-fold cross-validation procedure was used to evaluate the empirical results. Authors claimed that their method outperformed the logistic regression.

\section{Discussion and future directions} \label{sec8}
\subsection{Discussion}
Hyperbox fuzzy sets possess many attractive features to build highly efficient predictive models. In this paper, we gave an overview of properties and characteristics of existing hyperbox-based machine learning algorithms. The main characteristics of machine learning models are summarized in Table \ref{table5}. We categorized the existing studies taking advantage of the hyperbox representation for the construction of predictive models into three groups. The first group is the original fuzzy min-max neural network and its variants. The second one is the combination of fuzzy min-max neural networks and tree-based learning algorithms or the formulation of an ensemble model. The final group includes models that only use pure hyperboxes to cover training samples without forming network or tree architectures.

The idea of using hyperbox fuzzy sets for machine learning models was proposed in the ART neural networks \citep{Carpenter91}, and then it was improved to form the fuzzy min-max neural networks in \citeauthor{Simpson92}'s work. The fuzzy min-max neural networks have many strong points such as online learning ability, soft and hard decisions, construction of nonlinear decision boundaries, \textit{etc}. However, this type of neural networks still faces many drawbacks such as expansion and contraction problems, the dependence on the data presentation order, parameter sensitivity problem, and issues related to the membership function.

As a result, several improvements have aimed to form new learning algorithms for hyperbox-based neural networks such as using new rules to select hyperboxes for expansion, adding or modifying hyperbox test rules. Building a unified framework for fuzzy min-max clustering and classification neural networks provided a significant enhancement to the original version. Many researchers have targeted eliminating the hyperbox contraction process by implementing inclusion/exclusion neurons, compensatory neurons, overlapping neurons, using a multi-level network structure or some heuristic rules to choose the suitable hyperbox for samples located in the overlapping region among hyperboxes. To expand the hyperbox-based machine learning models for dealing with categorical data, several studies have taken discrete variables into consideration when building new membership functions and network architectures.

In addition to different types of hypebox-based neural networks, many studies have taken advantage of online learning ability of FMNN to combine with rule extractors such as decision tree, clustering tree, and classification and regression trees. These types of hybrid models are capable of generating explanatory rules for their outputs. Some other works have concentrated on evolving and optimizing the positions of hyperboxes from the initial set of hyperboxes by using nature-inspired algorithms such as differential evolution, genetic algorithms or ant colony optimization. The applications of hyperbox-based classification and clustering algorithms in the real world were presented in detail in this paper as well.

{
\scriptsize
\begin{longtable}[!ht]{|p{0.18\textwidth}|>\centering p{0.06\textwidth}|>\centering p{0.05\textwidth}|>\centering p{0.07\textwidth}|>\centering p{0.05\textwidth}|>\centering p{0.05\textwidth}|>\centering
p{0.05\textwidth}|>\centering
p{0.06\textwidth}|>\centering p{0.07\textwidth}|>\centering p{0.07\textwidth}|C{1.7cm}|}
\caption{Summary of main characteristics of hyperbox-based machine learning models} \label{table5} \\
\hline 
	\multirow{2}{*}{\textbf{Model}} & \textbf{Using contraction?} & \multicolumn{2}{c|}{\textbf{Data}} & \multicolumn{2}{c|}{\textbf{Learning}} & 
	\multicolumn{2}{c|}{\textbf{Type of model}} &
	\textbf{Affected by Noise?} & \textbf{Affected by data presentation order?} & \textbf{Using a special mechanism for overlapping handling}\\
	\cline{3-8}
	& & Labeled & Unlabeled & Online & Batch & Single & Ensemble & & &\\
	\endhead
        \hline
        FMNN for classification \citep{Simpson92} & \checkmark & \checkmark &  & \checkmark &  & \checkmark &  & \checkmark & \checkmark &  \\
        \hline
        FMNN for clustering \citep{Simpson93} & \checkmark &  & \checkmark & \checkmark & & \checkmark &  & \checkmark & \checkmark &  \\
        \hline
        HFC \citep{Abe95} &  & \checkmark &  & \checkmark &  & \checkmark &  & \checkmark & \checkmark & \checkmark \\
        \hline
        SFMN \citep{Likas96, Likas01} & \checkmark & \checkmark &  & \checkmark &  & \checkmark & & \checkmark & \checkmark &  \\
        \hline
        ARC \citep{Rizzi98} &  & \checkmark &  &  & \checkmark & \checkmark & & &  & \checkmark \\
        \hline
        MMM-BL \citep{Meneganti98} &  & \checkmark &  &  & \checkmark & \checkmark & &  &  & \checkmark \\
        \hline
        2lv-CSWCE \citep{Gabrys99} & \checkmark & \checkmark & \checkmark & \checkmark &  & & \checkmark & \checkmark & \checkmark & \\
        \hline
        GFMN \citep{Gabrys00} & \checkmark & \checkmark & \checkmark & \checkmark &  & \checkmark &  & \checkmark & \checkmark & \\
        \hline
        GFMN \citep{Gabrys02a} &  & \checkmark & \checkmark &  & \checkmark & \checkmark & &  &  & \\
        \hline
        TDFMM \citep{Tagliaferri01} &  & \checkmark &  &  & \checkmark & \checkmark & & &  & \checkmark \\
        \hline
        TDFMMR \citep{Tagliaferri01} &  & \checkmark &  &  & \checkmark & \checkmark & & &  & \checkmark \\
        \hline
        Esb-GFMN \citep{Gabrys02b} &  & \checkmark & \checkmark &  & \checkmark &  & \checkmark &  &  & \\
        \hline
        WFMM \citep{Kim04, Kim05} &  & \checkmark &  & \checkmark &  & \checkmark & & \checkmark & \checkmark & \\
        \hline
        MWFMM \citep{Kim06} &  & \checkmark &  & \checkmark &  & \checkmark & & \checkmark & \checkmark & \checkmark \\
        \hline
        IEFMN \citep{Bargiela04} &  & \checkmark &  & \checkmark &  & \checkmark &  & \checkmark & \checkmark & \checkmark \\
        \hline
        FMCN \citep{Nandedkar04, Nandedkar07b} &  & \checkmark &  & \checkmark &  & \checkmark & &  \checkmark & \checkmark & \checkmark \\
        \hline
        2-pHC \citep{Bortolan07} &  & \checkmark &  &  & \checkmark & \checkmark &  &  &  & \\
        \hline
        MFMNN \citep{Quteishat08b} & \checkmark & \checkmark &  & \checkmark &  & \checkmark & & \checkmark &  \checkmark & \\
        \hline
        HACO2 \citep{Ramos08} &  & \checkmark &  &  & \checkmark & \checkmark & &  & & \\
        \hline
        HACO \citep{Ramos09} &  &  & \checkmark &  & \checkmark & \checkmark &  &  & & \\
        \hline
        GRFMN \citep{Nandedkar09} &  & \checkmark & \checkmark & \checkmark &  & \checkmark &  & \checkmark & \checkmark & \checkmark\\
        \hline
        MCP/MILP \citep{Xu09} &  & \checkmark &  &  & \checkmark & \checkmark &  &  & & \checkmark \\
        \hline
         MFMNN-GA \citep{Quteishat10} & \checkmark & \checkmark &  &  & \checkmark & \checkmark &  &  & \checkmark &\\
        \hline
         DCFMN \citep{Zhang11} &  & \checkmark &  & \checkmark &  & \checkmark & &  &  & \checkmark\\
        \hline
        TEH-GFMN \citep{Eastwood11} &  & \checkmark &  &  & \checkmark &  & \checkmark &  &  & \\
        \hline
        HNN \citep{Palmer-Brown11} &  & \checkmark &  & \checkmark &  & \checkmark & & \checkmark & \checkmark & \\
        \hline
        MDCFMN \citep{Ma12} &  & \checkmark &  & \checkmark &  & \checkmark &  &  & \checkmark & \\
        \hline
        GFMN-CD1 \citep{Castillo12} & \checkmark & \checkmark &  &  \checkmark &  & \checkmark &  & \checkmark & \checkmark & \\
        \hline
        FMM-CART \citep{Seera12, Seera14} & \checkmark & \checkmark &  & \checkmark &  & \checkmark &  & \checkmark & \checkmark &\\
        \hline
        MLF \citep{Davtalab14} &  & \checkmark &  & \checkmark &  & \checkmark &  &  & & \checkmark\\
        \hline
        FMNWSM \citep{Forghani15} &  & \checkmark &  & \checkmark &  & \checkmark &  &  & \checkmark &\\
        \hline
        EFMNN \citep{Mohammed15} & \checkmark & \checkmark &  & \checkmark &  & \checkmark & & \checkmark & \checkmark &\\
        \hline
        MFMMC \citep{Seera15} & \checkmark &  & \checkmark & \checkmark &  & \checkmark &  & \checkmark & \checkmark &\\
        \hline
        GFMN-CD2 \citep{Shinde16} & \checkmark & \checkmark &  & \checkmark  &  & \checkmark & & \checkmark & \checkmark &\\
        \hline
        FMM-CT \citep{Seera16} & \checkmark &  & \checkmark & \checkmark &  & \checkmark & &  & \checkmark &\\
        \hline
        FMMDT for static data \citep{Mirzamomen16} &  & \checkmark &  &  & \checkmark & \checkmark &  &  & & \checkmark\\
        \hline
        FMMDT for streaming data \citep{Mirzamomen17} &  & \checkmark &  & \checkmark &  & \checkmark &  &  & \checkmark & \checkmark\\
        \hline
        FMM-GA \citep{Azad16} & \checkmark & \checkmark & & \checkmark & \checkmark & \checkmark & & & \checkmark & \\
        \hline
        KN-EFMNN \citep{Mohammed17a} & \checkmark & \checkmark &  & \checkmark &  & \checkmark & & \checkmark & \checkmark & \\
        \hline
        EFMNN-II \citep{Mohammed17b} & \checkmark & \checkmark &  & \checkmark &  & \checkmark & & \checkmark & \checkmark & \\
        \hline
        EFMNN-ACO \citep{Sonule17} & \checkmark & \checkmark &  & \checkmark &  & \checkmark & &  & \checkmark & \\
        \hline
        MFMC \citep{Liu17} & \checkmark &  & \checkmark & \checkmark &  & \checkmark & &  & \checkmark &\\
        \hline
        FMM-PSO \citep{Azad17} & \checkmark & \checkmark & & \checkmark & \checkmark & \checkmark & & & \checkmark & \\
        \hline
        FMM-ECT \citep{Seera18} & \checkmark &  & \checkmark & \checkmark &  & & \checkmark &  & \checkmark &\\
        \hline
\end{longtable}
}

\subsection{Future directions}
In spite of many extensive enhancements in existing literature, there are still a number of open problems having not been fully tackled in current studies on hyperbox-based machine learning algorithms, and ones which require more efforts on resolving them in the future work. Some potential research directions can be shown as follows:

\begin{itemize}
    \item \textit{Real-time big data analytics}. In the era of digital information, a critical growing trend of machine learning research based on hyperbox representations is to discover and extract valuable knowledge from real-time big data. However, it is extremely challenging to analyze the real-world big data sets because of their exceedingly high volume and velocity. To deal with these problems effectively, more research efforts should focus on developing parallel or distributed hyperbox-based models to make use of the strengths of grid computing and high performance computing systems. \citet{Ilager17} have been the first ones who proposed the use of MapReduce mechanism to scale the original FMNN for large data sets. Although the solution in that work is simple, the obtained results are promising. However, this research direction needs many enhancements and expansions to realize its full potential. In recent years, reconfigurable hardware, for instance Field-Programmable Gate Arrays (FPGAs), has attracted much attention due to the programmable property, massive parallelism, and energy efficiency \citep{Ghasemi17}. The FPGAs have recently been applied to accelerate large-scale tasks and improve significant performance with considerable power savings, which suggests their potential for large-scale high-performance computations \citep{Putnam14}. Hence, implementation of hyperbox-based machine learning algorithms on FPGAs is also interesting research direction in near future.
    
    \item \textit{Streaming data mining}. In the practical applications, the data may change constantly over time, so hyperboxes need an adaptive ability. It is essential to find a way of re-sizing or evolving the previously learned hyperboxes or ignoring these hyperboxes if they are no longer appropriate for the new case \citep{Mirzamomen17}. Another problem for streaming data is the case where the target concepts to be predicted are likely to change over time in unanticipated ways, which is called the ``concept drift" problem. In general, there is still no formal hyperbox-based techniques for dealing with all kinds of issues, especially in the situations that target concepts change constantly in arbitrary ways.
    
    \item \textit{Automatic adaptation of models}. In order to discover information from real-time data, machine learning models need to adapt with the changes of data. Therefore, significant research efforts should be put into this problem to formulate adaptive predictors.
    
    \item \textit{Generation of ensemble models.} It is really difficult to achieve acceptable classification results for big data sets with massive dimensions. Therefore, one may use hyperbox representation to select valuable features and train classifiers on different training subsets. The results of base classifiers or the hyperbox components in base predictors are then able to be combined following a certain method to output the final result or model. An effective way of generating ensemble models of hyperbox-based classifiers is highly desired.
    
    \item \textit{Automatic optimization of hyperparameters and parameters.} It has been frequently observed that the performance of current hyperbox-based models depends on various user definable hyperparameters and parameters. Therefore, it is necessary to investigate the importance of each parameter to the performance of algorithms. Studies also need to find the optimal configurations for parameters or adapt the values of parameters according to the features of data sets in an automated manner.
    
    \item \textit{Learning from multiple data sources.} All current hyperbox-based machine learning algorithms have learned from only single source-structured data. In the real applications, data can come from many sources in many different formats (structured, unstructured, or semi-structured data). Therefore, future research should handle these challenges by building multi-source hyperboxed-based learning algorithms.
    
    \item \textit{Improvement of membership functions.} The classification performance of current hyperbox-based classifiers significantly depends on the membership function. Therefore, one can propose a novel membership function based on the features of data and the influence of noise for both numerical and nominal data. Current membership function for both numerical and categorical data proposed by \citet{Castillo12} and \citet{Shinde16} did not take the features of data and noise into account. Another problem of current membership functions is that they do not yet classify data in the overlapping regions among hyperboxes belonging to different classes effectively. Certainly, current studies have not yet fully handled all of these problems.
    
    \item \textit{Data editing.} The data quality is one of factors influencing the accuracy of final classification results. In terms of real-time data analysis, the input data may contain noise or missing values or incomplete data. The use of preprocessing methods might not be sufficient to enhance the quality of input data. The ways of dealing with missing values using hyperbox fuzzy sets proposed by \citet{Gabrys02c} for numerical data and \citet{Castillo12} for categorical data are efficient techniques to handle missing or incomplete data. Therefore, many research efforts should concentrate on enhancing the robustness of hyperbox-based learning algorithms when resolving the poor quality of real data.
\end{itemize}

\section{Conclusions} \label{sec9}
This paper presented a comprehensive survey of existing studies on hyperbox-based machine learning algorithms. One of the interesting features of hyperbox-based models is their online learning capability, which meets the urgent demand for constructing effective machine learning approaches for real big data analytics. The hyperbox representation is especially suitable for big data applications such as image processing because the use of hyperboxes assists the processing and inference in the form of granules instead of handling individual points. This solution can significantly decrease computational cost in large systems. The real-world applications usually exhibit uncertain behaviors, so the input data of the machine learning algorithms is likely to contain both granular and crisp points or missing values. Hence, effective algorithms must be able to cope with uncertain and ambiguous data. The hyperbox representation method in the general fuzzy min-max neural networks allows handling both fuzzy and crisp input samples as well as inputs with missing values. An interesting characteristic of the general fuzzy min-max neural networks is that they unify both classification and clustering problems in the same framework. This feature endows the model with the capability of learning from unlabeled, semi-labeled, and labeled data. Another attribute of the practical systems is the continuous changing over time, so they require machine learning models capable of evolving to fulfill the operations without retraining. The hyperbox creation, extension, and contraction mechanisms using fast Boolean algebra help hyperbox-based machine learning algorithms evolve and adjust to adapt rapidly to the changes in the data. Another advantage of hyperboxes is that they can be used to build the rule sets for a given problem. This ability contributes to the increase in the reliability of the predictive results as well as making the algorithms friendly and transparent to end-users.

Since the fuzzy min-max neural network was first introduced by \citet{Simpson92} with a number of interesting features and characteristics, many studies have been focused on enhancing this kind of neural network or inventing novel learning algorithms based on the concept of hyperbox fuzzy sets. Although this survey made an attempt to cover as much research as possible on machine learning algorithms based on hyperboxes, it is very challenging to collect related studies adequately as learning through hyperbox representation has developed rapidly in recent years. We also gain insight that although hyperbox-based machine learning algorithms have achieved remarkable results, there are still outstanding issues and room for the improvements.






\bibliographystyle{spbasic}      
\bibliography{MyRef}

\begin{appendices}
The key steps to seek for literature are presented as follows:

\section{Formulating search terms and selecting research databases} \label{appendix1}
The main terms used to construct the search strings are \textit{hyperbox}, \textit{fuzzy min max}, and \textit{classifier}. We used Boolean operators to build a search expression as follows:

(hyperbox or ``fuzzy min max") and (classifier or classification or clustering or algorithm)

This search string was employed to seek for research articles, conference papers or book chapters in five popular databases including \citet{ScienceDirect18}, \citet{IEEEXplore18}, \citet{SpringerLink18}, \citet{ACM18}, \citet{IOSPress18}. In addition, literature which was cited in the selected ones and was satisfied the inclusion criteria was also considered.

\section{Publication selection criteria} \label{appendix2}
\subsection{Inclusion criteria}
We applied the following inclusion criteria to choose relevant studies including journal articles and magazines, book chapters, and conference papers:

\begin{itemize}
    \item (I1) Studies that propose a new machine learning model using hyperbox representations or a significant improvement of the existing hyperbox-based model.
    \item (I2) Studies that apply the hyperbox-based machine learning models or their variants to deal with the real world problems. These models must illustrate their effectiveness on practical data sets.
    \item (I3) Papers are peer-reviewed and published either in a specialized proceedings or in a reputable journal.
    \item (I4) Studies that were not published from year 1992 to year 2018
\end{itemize}

\subsection{Exclusion criteria}
Exclusion criteria are used to exclude the studies which are irrelevant to this research. We designed these conditions as follows:

\begin{itemize}
    \item Papers which are not relevant to the research question.
    \item Studies that mention to hyperbox-based machine learning algorithms but not focus on enhancing the existing methods or solving a new real world issues.
    \item Papers that their contents are simple, and the authors do not describe or analyze their novel contribution to the research topic.
    \item Studies that do not resolve the classification or clustering problems.
\end{itemize}

\subsection{Literature selection strategy}
Literature was evaluated through three stages: automatic search, screening, and eligible selection. In the first step, above search strings were put into search engines of five databases to seek for studies of interest. In the next phase, the title and the abstract of the studies obtained in the previous phase were verified whether it satisfies our field of interest in this paper. To select the satisfied papers, we read through all papers in the second steps and assessed the studies based on the satisfactions of inclusion criteria either I1 or I2, and I3 and I4. If the paper meets any exclusion criteria, it would be discarded. The quality of the proposed methods in the literature and their efficiency in comparison with other similar approaches were considered in the choice of final papers as well. In the evaluation process, other publications cited in the papers of the third stages were reviewed, and qualified works closely related to this study were also selected. 

\end{appendices}
\end{document}